\RequirePackage{fix-cm}
\RequirePackage{amsmath}
\documentclass{svjour3}                     
\smartqed  
\usepackage{graphicx}
\usepackage{subfigure}
\usepackage{mathtools}
\usepackage{amsmath}
\usepackage{wasysym}
\usepackage{natbib}
\usepackage[bookmarks,final]{hyperref}
\usepackage{multirow}
\usepackage{amssymb}
\usepackage{textcomp}
\usepackage{gensymb}

\hypersetup{
	colorlinks=true,       		
	linkcolor=black,          	
	citecolor=blue,        		
	urlcolor=blue,              
}

 \journalname{Data Mining and Knowledge Discovery}
\begin{document}

\title{Challenges in Benchmarking Stream Learning Algorithms with Real-world Data
}

\titlerunning{Challenges in Benchmarking Stream Learning}        

\author{Vinicius M. A. Souza         \and
        Denis M. dos Reis \and
        Andre G. Maletzke \and 
        Gustavo E. A. P. A. Batista
}

\authorrunning{V. M. A. Souza et al.} 

\institute{V. M. A. Souza\at
                Universidade de S\~{a}o Paulo, S\~{a}o Carlos, SP, Brazil \\              
                University of New Mexico, Albuquerque, NM, USA\\
                \email{vinicius@unm.edu}           
           \and
           D. M. dos Reis\at
                Universidade de S\~{a}o Paulo, S\~{a}o Carlos, SP, Brazil \\           
                \email{denismr@usp.br}
            \and
            A. G. Maletzke\at
                Universidade de S\~{a}o Paulo, S\~{a}o Carlos, SP, Brazil \\                       
                \email{andregustavo@usp.br}
           \and
           G. E. A. P. A. Batista \at
                University of New South Wales, Sydney, Australia \\
                Universidade de S\~{a}o Paulo, S\~{a}o Carlos, SP, Brazil \\                     
                 \email{g.batista@unsw.edu.au} 
}

\date{Received: date / Accepted: date}

\maketitle

\begin{abstract}
Streaming data are increasingly present in real-world applications such as sensor measurements, satellite data feed, stock market, and financial data.  The main characteristics of these applications are the online arrival of data observations at high speed and the susceptibility to changes in the data distributions due to the dynamic nature of real environments. The data stream mining community still faces some primary challenges and difficulties related to the comparison and evaluation of new proposals, mainly due to the lack of publicly available non-stationary real-world datasets. The comparison of stream algorithms proposed in the literature is not an easy task, as authors do not always follow the same recommendations, experimental evaluation procedures, datasets, and assumptions. In this paper, we mitigate problems related to the choice of datasets in the experimental evaluation of stream classifiers and drift detectors. To that end, we propose a new public data repository for benchmarking stream algorithms with real-world data. This repository contains the most popular datasets from literature and new datasets related to a highly relevant public health problem that involves the recognition of disease vector insects using optical sensors. The main advantage of these new datasets is the prior knowledge of their characteristics and patterns of changes to evaluate new adaptive algorithm proposals adequately. We also present an in-depth discussion about the characteristics, reasons, and issues that lead to different types of changes in data distribution, as well as a critical review of common problems concerning the current benchmark datasets available in the literature.

\keywords{Data stream \and Concept drift \and Classification \and Drift detection \and Benchmark data}
\end{abstract}

\section{Introduction}

In the last 20 years, we have witnessed the emergence and notable increase in the interest of algorithms that learn from streaming data. This new generation of machine learning methods is designed to deal with continuous flows of data. Frequently, such streams comprise changes in the distribution of data, which are governed by the dynamics of real-world problems and application domains that evolve. In the context of machine learning, these changes in data distribution are named \textit{concept drifts} \citep{widmer1996learning} and typically occur in data that are observed continuously at a fast rate, which in turn impose time and memory constraints on the algorithms that process them.

Batch learning is the standard machine learning approach that assumes the whole dataset is available at training time. Batch learning is a mature field with clear procedures to evaluate and compare different methods using a vastitude of data shared by researchers for benchmarking. However, in the online scenario of data stream mining, we still face some primary challenges and difficulties related to the comparison and evaluation of new proposals, mainly due to the lack of publicly available non-stationary real-world datasets. For example, we found more than 300 stationary datasets for classification problems from different domains in the UCI Machine Learning Repository \citep{Dua:2017}. In particular, for time-series classification, the UEA \& UCR Time Series Classification Repository \citep{bagnallUEA} stores more than 100 datasets and 20 algorithms from the literature. For data stream mining, although there is the popular open-source framework MOA \citep{bifet2010moa} with a collection of algorithms and synthetic data generators, we do not have any public repository with a reasonably-sized collection of real-world stream datasets accompanied of their detailed description. More alarming, different data stream algorithms often run on specific assumptions about the data (for instance, methods may assume changes to be either incremental or abrupt). However, frequently it is not clear whether employed datasets fulfill such assumptions or not.

As recently noted by \cite{krawczyk2017ensemble}, the comparison of stream learning algorithms is not an easy task, as authors do not always follow the same recommendations, experimental evaluation procedures, datasets, and assumptions. In this paper, we want to mitigate the problems related to the choice of datasets in the experimental evaluation of stream classifiers and drift detectors.

In summary, the main contributions of this article are the following:

\begin{itemize}

 \item Presentation of basic concepts of data stream mining accompanied by an in-depth discussion about the characteristics, reasons, and issues, which lead to different types of changes in data distribution;

 \item A review of the main real-world datasets adopted in the evaluation of stream learning methods accompanied by a critical discussion of issues concerning the data and the challenges imposed due to the lack of a benchmark standard;

 \item Presentation of a relevant public health problem that involves the recognition of disease vector insects by an optical sensor, which is responsible for generating evolving data over time;
 
 \item Building (data collection, preprocessing, and features extraction) of 11 new real-world datasets with controlled concept changes where it is possible to identify the types/patterns of changes and when they occur for each dataset. Such data are accompanied by an experimental evaluation that includes state-of-the-art classifiers and drift detectors;

 \item Development and availability of a repository\footnote{USP Data Stream  Repository -- Available online at \url{https://sites.google.com/view/uspdsrepository}.}
 with 27 real-world datasets for benchmarking the evaluation of stream classifiers and change detectors. 
\end{itemize}

This paper is organized as follows. In Section~\ref{sec:background}, we provide a background on data stream and concept drift, as well as a discussion about characteristics, reasons, and issues that lead to changes in the data distribution. In Section~\ref{sec:datasets_literature}, we present an overview of the most common datasets used in the evaluation of stream mining approaches. In Section~\ref{sec:datasets_criticisms}, we discuss the challenges faced by the stream mining community when the currently most popular real-world datasets are evaluated. In Section~\ref{sec:sensor}, we introduce a benchmark dataset for stream learning with controlled concept drifts, which are generated by an optical sensor that measures characteristics of insect flights. In Section~\ref{sec:repository}, we present to the stream learning community a new public repository with an initial amount of 27 datasets. In Section~\ref{sec:evaluation}, we perform an evaluation followed by a discussion concerning the 11 datasets introduced in this paper. Finally, we conclude our work in Section~\ref{sec:conclusions}.

\section{Background}\label{sec:background}
In this section, we present the main concepts and definitions regarding data streams, concept drift and the tasks of classification and drift detection under non-stationary environments.

\subsection{Data Stream}\label{subsec:ds}

A data stream is an ordered sequence of instances continuously observed over time. Streaming data are increasingly prevalent in real-world applications. Representative examples include network traffic, database transactions, sensor measurements, satellite data feed, stock market and financial data, georeferenced data from mobile devices, among others.

Formally, a data stream is a sequence of instances $\mathcal{DS} = \{\vec{X}_1, \vec{X}_2, \ldots, \vec{X}_t, \ldots\}$, where $\vec{X}_t \in \mathcal{X}$ is a $d$-dimensional vector in the feature space observed at time $t$. In practice, $\vec{X}_t$ is an ordered list of descriptive attributes that represent the observation being made. The attributes can be qualitative (nominal, ordinal, or binary) or quantitative (discrete or continuous).

Among possible tasks such as clustering, regression, graph-mining, outlier detection, and recommender systems, classification is probably the most prominent task on data streams and the focus of this work. In classification problems, each instance $\vec{X}_t \in \mathcal{X}$ is associated to a class label $y_t \in \mathcal{Y}$, where $\mathcal{Y}$ contains $l$ possible labels, $\mathcal{Y} = \{C_1, C_2, \ldots, C_l\}$. Therefore, a classification data stream is a sequence of pairs $\mathcal{DS} = \{(\vec{X}_1,y_1), (\vec{X}_2,y_2), \ldots, (\vec{X}_t,y_t), \ldots\}$.

Due to a data stream's potentially infinite length, traditional batch methods are often not applicable \citep{Gama:2007}. These methods typically fail to comply with at least one of the three most prominent restrictions of the stream setting \citep{Bifet:2009}:

\begin{enumerate}
\item In general, it is impractical to store all events of a stream due to its potentially infinite length. Only a small portion can be retained in memory;

\item Fast-paced streams require each event to be processed in real-time and, afterwards, discarded; and

\item The underlying distribution of the data may change over time. Hence, old data can become irrelevant or even detrimental to model the current concept. Contrary to batch learning, in data stream, we expect the characteristics of the newly observed data to change when compared to past data.
\end{enumerate}

The first constraint limits the amount of memory the algorithms can use, and the second constraint limits the time that is available for processing each event. Therefore, the first two restrictions lead to the development of techniques that reduce the information in a stream of data, such as sampling \citep{Chaudhuri:1999}, \textit{sketching} \citep{Alon:1999}, histograms \citep{Gilbert:2002}, \textit{wavelets} \citep{Matias:2000}, and sliding windows \citep{Datar:2002}. The third constraint guides the development of algorithms that are capable of detecting changes in the data and reacting by updating the existing models.

The non-stationarity of many real environments may lead to changes in the underlying distribution of the observed data, a phenomenon that goes by many names in literature, where the most common is \textit{concept drift}~\citep{widmer1996learning}. According to the same terminology, the data distribution in a given moment is called \textit{concept}. Additionally, a change in its parameters is a \textit{drift}.

Drifts constitute a central issue since they can decrease the performance of machine learning models induced with historical data \citep{pan2009survey,quionero2009dataset,saenko2010adapting,ben2007analysis}. A closely related problem in batch learning is \textit{concept shift}, which occurs when a model is trained with data from one distribution and is later applied on data that follow a different distribution~\citep{moreno2012unifying}. Similarly, \textit{transfer learning} aims to extract the knowledge from a source domain where abundant labeled data are available and applies this knowledge to a related target domain in which insufficient labeled data are available~\citep{pan2009survey}.

\subsection{Concept Drift}\label{subsec:cd}

\begin{figure}[htb]
    \centering
    \includegraphics[scale=0.4]{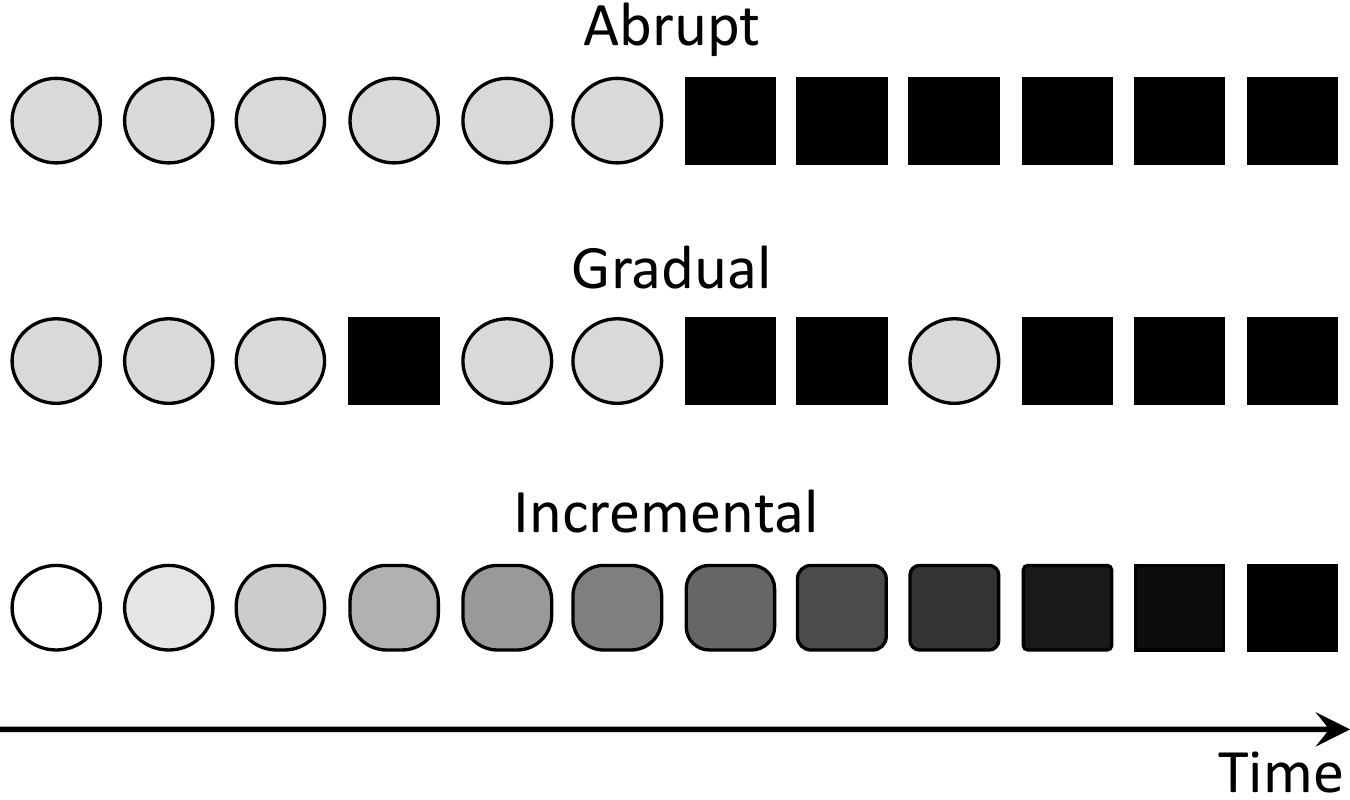}
    \caption{Representation of three types of concept drift over time.}
    \label{fig:drift_types}
\end{figure}

Concept drifts may manifest with different velocity, severity, and patterns. To illustrate different patterns, consider a concept represented by the color and shape of a geometric figure. Fig.~\ref{fig:drift_types} illustrates three types of drifts discussed in this work: abrupt, gradual, and incremental.

An abrupt drift occurs when the underlying distribution of the data suddenly changes into a different distribution. In other words, after an abrupt transition between two observations, all new data points belong to a concept different from the previous one. In the incremental change, there are several intermediary concepts between one initial concept and a final concept. Consecutive concepts within this transition period may be indistinguishable. In the case of gradual concept drift, the transition between two concepts occurs smoothly. However, differently from incremental drift, in a gradual drift, the probability of observing instances that belong to the previous concept decreases over time while, simultaneously, the probability of observing instances that belong to the new concept increases, even though both concepts are remarkably distinct and stationary during the transition period. Concepts that were seen in the past and are later observed again are called recurring concepts. We note that one-off random deviations in the data, such as outliers or noise, are not considered to be concept drifts. 

A practical way to identify such patterns of change is to analyze the data distribution over sliding windows in the stream. A window represents a sample of examples that are observed in sequence within a period. When we move the boundaries that define the first and last data points of this window over the stream to comprise different intervals, we have a sliding window. We note that the use of windows imposes the choice of essential parameters. The most common is the number of data points that will be comprised by the window, and how much is the overlap between consecutive windows. 

We can only indirectly observe the underlying concept of the data by analyzing samples of instances in the sliding window. Therefore, the observation of concept drift is also indirect. As the number of instances is finite, there is a discrete and finite number of observable distributions that can be analyzed. Note that which examples are included in the window change the perception of the distribution: a bigger window can hide inner distributions that would be perceived as distinct with smaller windows. On the other hand, it may be infeasible to recognize certain concepts if we can only observe too few data points for each occurring concept.

\subsection{Independent Distribution}\label{sec:id}

In batch learning, a common assumption is that examples are independent and identically distributed (i.i.d.). In data stream applications, such an assumption usually does not conform to reality. Identically distributed means that the joint distribution of an example and its class label is the same at any time, that is, $P(\vec{X}_t,y_t) = P(\vec{X}_{t'},y_{t'})$, when $t \neq t'$. Meanwhile, independently distributed means that the probability of the current label does not depend on what was observed before; that is, $P(y_t) = P(y_t|y_{t-k})$.

Since data streams are generated in dynamic environments, examples are not identically distributed due to the occurrence of concept drifts. Additionally, while most of the literature assumes independence between examples, some recent studies have found a significant \textit{temporal dependence} in many real-world data \citep{bifet2013pitfalls, vzliobaite2015evaluation}. This dependence on historical class labels has a direct impact on the design and evaluation of stream approaches.

The existence of temporal dependence reinforces that the line separating data stream from time-series is blurred. One possible view for streams is that the instances are independent of each other in the sense that the occurrence or absence of one \textit{particular} instance implies neither the presence or absence of other \textit{particular} instances \citep{reis2017:sac}. However, the observation of any instance is under the influence of a shared background concept, which is not directly observable. Even though the instances are not particularly dependent on each other, the occurrence of one instance \textit{may} be indicative of the likelihood of observing instances of a particular class or in a specific region of the feature space, due to this common background concept. One example is motion recognition. In this problem, sensors attached to a particular participant may indicate that this person is performing the \textit{eating} activity. Since this assessment can be an indication that it is lunchtime, the likelihood of observing other people performing the same activity may increase, albeit all the analyzed people being unrelated.

A different view, closer to time-series, is that one attribute value is the result of an auto-regressive transformation applied to previous instances. Examples are the variation in the price of commodities such as electricity \citep{zliobaite2013good} and the evolution of weather \citep{ditzler2013incremental}.

Datasets can mix the two views mentioned above by combining features from different sources. In this case, it is imperative to make the distinction between both views, since change detection in time-series and drift detection in data streams with independent examples are distinct research topics that require remarkably different approaches. To elucidate, consider a problem where one of the descriptive features is a time-series defined by a strictly crescent monotonic function. Any two non-overlapping sliding windows over this series have statistically different probability distributions. Although this difference in the distribution exists for the time-series, it may not be indicative of a change in other aspects of data. One example is the analysis of the behavior of fish species. Any consistent change in water temperature is statistically detected. However, the magnitude of this change may not be enough to affect fish behavior.

\subsection{Types of Concept Drifts}\label{subsec:typcd}

In the classification task, a predictive model learns a function that maps the input variables $\mathcal{X}$ representing the feature space to discrete output variables $\mathcal{Y}$ of class labels. \cite{fawcett2005response} state that there are two types of problems based on the causal direction of such a relationship between the feature space and the class labels. Additionally, only some types of drifts can occur for each type of problem. The types of problems are:

\begin{description}
\item[$\mathcal{X} \rightarrow \mathcal{Y}$:] the class label is derived from the behavior of the instance. One example is recognizing specific body movements of a person with sensors. The joint distribution is often written as $P(X,Y) = P(Y|X)P(X)$;
\item[$\mathcal{Y} \rightarrow \mathcal{X}$:] the class label determines the values of the features. One example is a disease diagnosis in which the disease causes symptoms. The joint distribution is often written as $P(X,Y) = P(X|Y)P(Y)$.
\end{description}

Furthermore, drift incurs a difference between the two concepts. To simplify our discussion, we call $P_A$ the probability before the drift and $P_B$ the probability after the drift has occurred so that we can compare both distributions.

Based on the types of problems listed before, \cite{moreno2012unifying} review and compile the nomenclatures and definitions from literature into a single reference list of types of drifts. Although general changes are commonly referred to as \emph{concept drift} in the data stream literature, \cite{moreno2012unifying} provide a normalization in which all types of change go by \emph{dataset shift}. Additionally, a change belongs to one among three more specific types: \textit{covariate shift}, \textit{prior probability shift}, and \textit{concept shift}.

\textit{Covariate shift} refers to changes in the feature space alone and, according to \cite{moreno2012unifying}, only happens in $\mathcal{X} \rightarrow \mathcal{Y}$ problems. It is defined as follows:

\begin{definition}
Covariate shift is the case where $P_A(Y|X) = P_B(Y|X)$ and $P_A(X) \neq P_B(X)$.
\end{definition}

\textit{Prior probability shift} refers to changes in the class proportions alone. It is the main subject of study in a new subfield of Machine Learning called class prior estimation or quantification \citep{gonzalez2017review}. According to \cite{moreno2012unifying}, this type of change only happens in $\mathcal{Y} \rightarrow \mathcal{X}$ problems, and is defined as follows:

\begin{definition}
Prior probability shift is the case where $P_A(X|Y) = P_B(X|Y)$ and $P_A(Y) \neq P_B(Y)$.
\end{definition}

We note that, in prior probability shift, although $P_A(X|Y) = P_B(X|Y)$, it is not necessarily true that $P_A(Y|X) = P_B(Y|X)$. This can be easily observed by changing $P(Y)$ in datasets with classes that highly overlap. To illustrate, recall that classifiers typically learn to classify instances in a region of the feature space as the most common class in the region. However, which is the most common class is subject to change according to alterations in $P(Y)$. Yet, the behavior of each class, individually, remains the same.

\textit{Concept shift} is a change in the relationship between the feature space and the class labels, and is, according to \cite{moreno2012unifying}, the hardest type of shift. It is defined as follows:

\begin{definition}
Concept shift is the case where one of the following happens:
\begin{enumerate}
    \item $P_A(Y|X) \neq P_B(Y|X)$ and $P_A(X) = P_B(X)$ in $\mathcal{X} \rightarrow \mathcal{Y}$ problems;
    \item $P_A(X|Y) \neq P_B(X|Y)$ and $P_A(Y) = P_B(Y)$ in $\mathcal{Y} \rightarrow \mathcal{X}$ problems;
    \item $P_A(Y|X) \neq P_B(Y|X)$ and $P_A(X) \neq P_B(X)$ in $\mathcal{X} \rightarrow \mathcal{Y}$ problems;
    \item $P_A(X|Y) \neq P_B(X|Y)$ and $P_A(Y) \neq P_B(Y)$ in $\mathcal{Y} \rightarrow \mathcal{X}$ problems.
\end{enumerate}
\end{definition}

Condition 1 states that the proportions of the classes, given the characteristics of the data points, change, while the distribution of these characteristics remains the same. In practical terms, we have two effects. First, if we ignore the label information and compare the data before and after the drift, they have the same probability distribution. Second, the proportion of classes in regions of the feature space changes. The only difference between conditions 1 and 3 is that the latter is free of the restriction of preserving $P(X)$.

Condition 2 states that the characteristics that define each class change. However, $P(Y)$ must be preserved, while $P(X)$ can change. $P(Y)$, on its own, dictates the proportion of the classes considering all data points. The only difference between conditions 3 and 4 is that the latter is free of the restriction of preserving $P(Y)$. For that reason, condition 4 is similar to the prior probability shift with the aggravating factor that the characteristics of each class have changed.

\cite{moreno2012unifying} state that the concept shifts 3 and 4 are rarer and possibly impossible to tackle. On the other hand, the two first shifts are easier to deal with. However, we find no reason to believe that concept shifts 3 and 4 are rare. We emphasize that such conditions are not mutually exclusive except for the pairs $(2,  4)$ and $(1, 3)$.

\begin{figure}[htbp]
    \centering
    \includegraphics[scale=0.35]{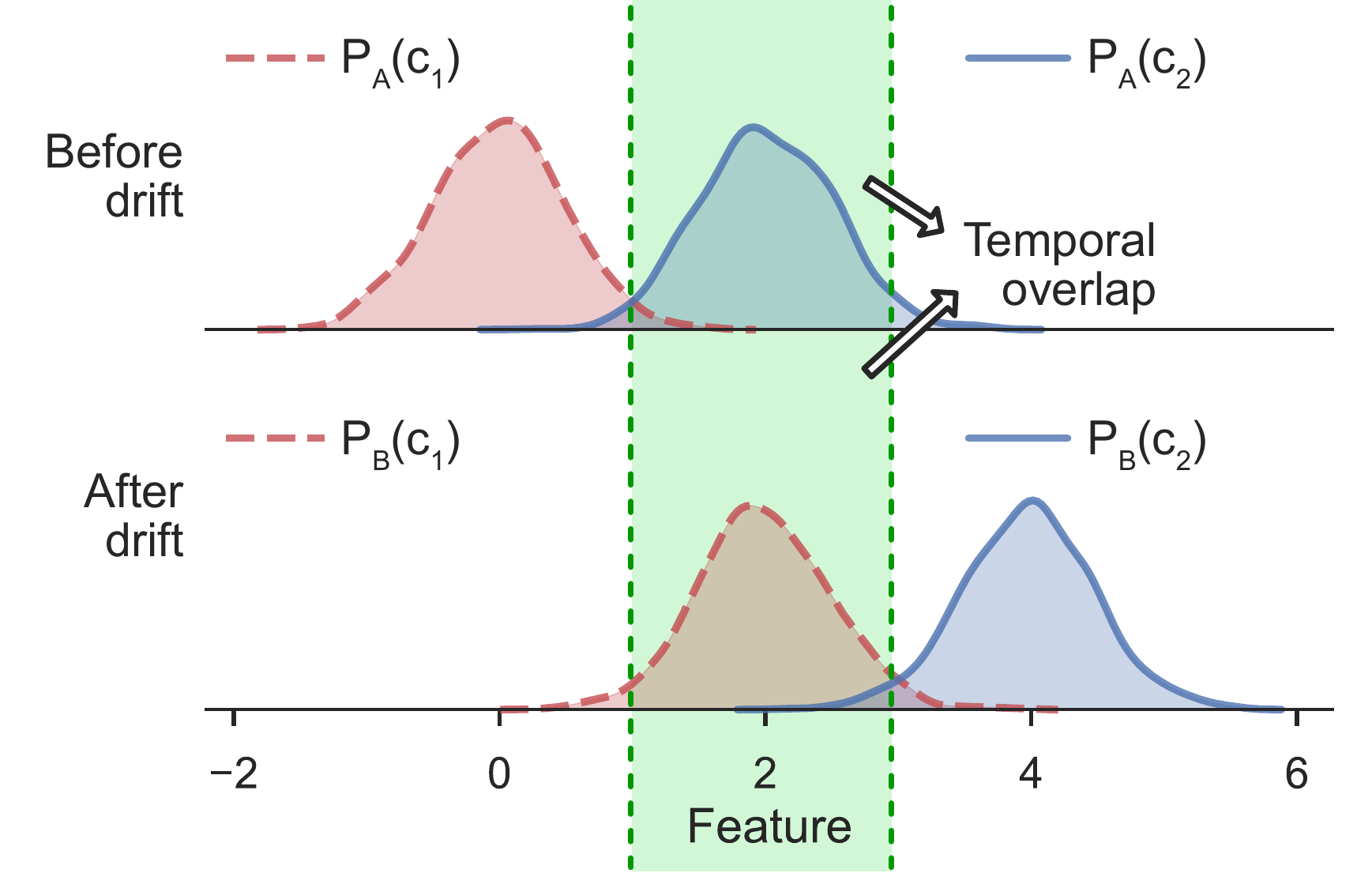}
     \caption[Case of concept shift]{Illustration of a case where $P_A(Y|X) \neq P_B(Y|X)$ and $P_A(X) \neq P_B(X)$. There are two classes ($c_1$ and $c_2$). Their distributions are shown before and after the drift happened. The drift is a global linear transformation that moved the average feature value two units up. The green shade illustrates a temporal overlap: instances that belong to class $c_1$ would seem to belong to class $c_2$ according to the outdated distributions. In this particular example, since there is no change in the proportion of classes, it is also true that $P_A(X|Y) \neq P_B(X|Y)$ and $P_A(Y) = P_B(Y)$. Therefore, this figure represents concepts shifts 2 and 3.}
    \label{fig:concept:shift}
\end{figure}

A simple global linear transformation that moves all instances towards some direction can cause concept shift 3. In this situation, if the proportion of classes is kept the same after the drift, we would also simultaneously fulfill condition 2. Otherwise, we would simultaneously fulfill condition 4. A linear transformation without changes in the proportion of classes is illustrated in Fig. \ref{fig:concept:shift}. We analyze a case of temporal overlap on real-world data in Section~\ref{subsec:insects:temporal:overlap}.

Fig. \ref{fig:concept:shift} has a \textit{temporal overlap}. The temporal overlap is a superposition of instances that belong to different classes in the feature space, and that only occurs if we ignore the temporal aspect of the data. For example, if we process the whole dataset at once. If we fail to temporally split the data so that we can separately analyze the concepts before and after the drift, we identify a greater class overlap than the one that exists in each concept individually. In this particular case, our view of the data would suggest that around 50\% of $c_1$ overlaps with around 50\% of $c_2$ in the feature space, which is a strikingly harder classification scenario than the one found in each isolated concept. The existence of temporal overlap reinforces the importance of adequately choosing the parameters of observation windows.

While concept shifts 3 and 4 may be hard to tackle in typical batch learning problems, some assumptions can make them identifiable in streams. For instance, changes in a stream can be incremental and, therefore, traceable over time \citep{dyer2014compose,souza2015classification,souza2015data}, or can always lead to a previously seen distribution of the data \citep{moreira2018classifying}. 

\begin{figure}[thbp]
\centering
\subfigure[Before concept drift]{
    \includegraphics[width=0.3\linewidth, trim={0 0.3cm 0 0.4cm},clip]{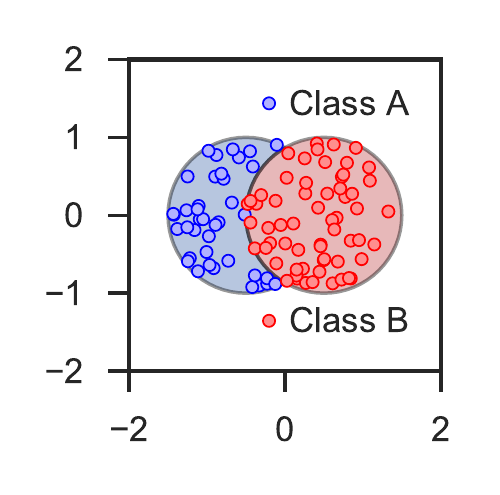}
}
\subfigure[After undetectable drift]{
    \includegraphics[width=0.3\linewidth, trim={0 0.3cm 0 0.4cm},clip]{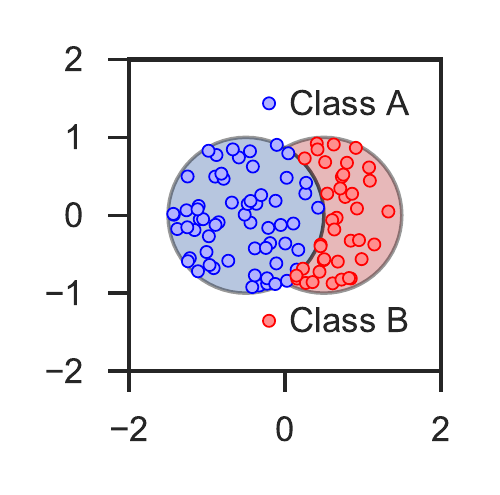}
}
\subfigure[Unlabeled data]{
    \includegraphics[width=0.3\linewidth, trim={0 0.3cm 0 0.4cm},clip]{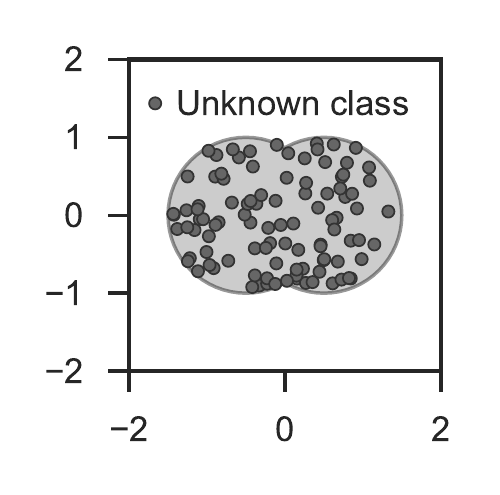}
}
\caption[Undetectable concept drift]{Illustration of a concept drift that is undetectable without true labels in a two-dimensional feature space with two classes. Red dots represent events belonging to class $A$, that are generated with the red-shaded area. Blue dots represent events belonging to class $B$, that are generated withing the blue-shaded area. In general, undetectable changes without true labels are those in which $P_A(X) = P_B(X)$ while $P_A(Y|X) \neq P_B(Y|X)$ \citep{reis2017:sac}.}
\label{fig:undetectable:drift}
\end{figure}

Finally, we contest the easiness of concept shift 1: in fact, this type of drift is impossible to detect in unsupervised settings, since we can only observe $P(X)$ and it does not change \citep{zliobaite2010change}. This fact is visually illustrated in Fig.~\ref{fig:undetectable:drift}. Since the proportion of the classes is changed non-uniformly to preserve $P(X)$, this figure also illustrates a concept shift 4.

\begin{figure}[htbp]
    \centering
    \includegraphics[width=\linewidth]{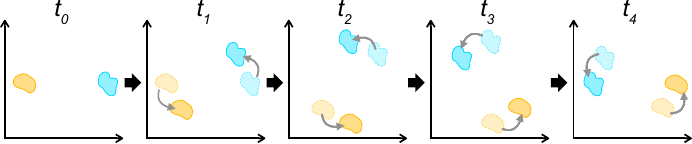}
     \caption[Case of traceable drift]{Illustration of a case of traceable incremental drift. There are two classes, distinguishable by their unique color. This illustration presents five snapshots that represent the evolution of the data over time. The thick arrow represents a passage from one snapshot to the following. The curved arrow represents the movement of the class in the feature space. A brighter class with a dashed border represents the previous position. Notice that if we only compare $t_0$ with $t_4$, we have a case that approximates $P_A(Y|X) \neq P_B(Y|X)$ and $P_A(X) = P_B(X)$, which is undetectable without true labels. However, with the support of the intermediate distributions, we can trace the geometric evolution of the data and therefore distinguish both distributions.}
    \label{fig:traceability}
\end{figure}

We note that, in specific settings, a drift may be undetectable when windows are far apart from each other in the stream. However, if there are intermediate changes of concept between the distributions estimated upon the first and last windows, and with proper setting of the observation windows, it may be possible to trace the evolution of the drift over time and detect it without true labels \citep{dyer2014compose,souza2015classification,souza2015data}. Fig. \ref{fig:traceability} illustrates such a case. 

\cite{kullpatterns} extend the work of \cite{moreno2012unifying} by introducing graphical notations of the dataset shifts types mentioned above, and 12 new additional sub-types of shifts. We point the interested reader to this paper for further information on this topic. Oppositely, \cite{Kelly:1999,tsymbal2004problem} offer a simplified view that is often enough to specify a concept drift problem. According to them, \emph{concept drift} occurs when $P(X), P(X|Y)$ or $P(Y|X)$ change.

The cases where $P(X)$ changes, while $P(Y|X)$ does not change, are referred to as \textit{virtual drift}. Opposite cases, where $P(Y|X)$ changes while $P(X)$ does not, occur due to alterations in the \textit{hidden context}. Hidden context is the information that is not included in the observable predictive features but is relevant to determine the class label \citep{harries1998extracting}. Furthermore, severe changes in $P(Y)$ can lead to class imbalance. Such changes in the class distribution can make a majority class to become a minority class and vice-versa in the course of a stream~\citep{Maletzke:2018-LIDTA}.

\begin{figure}[htb]
    \centering
    \includegraphics[scale=0.52]{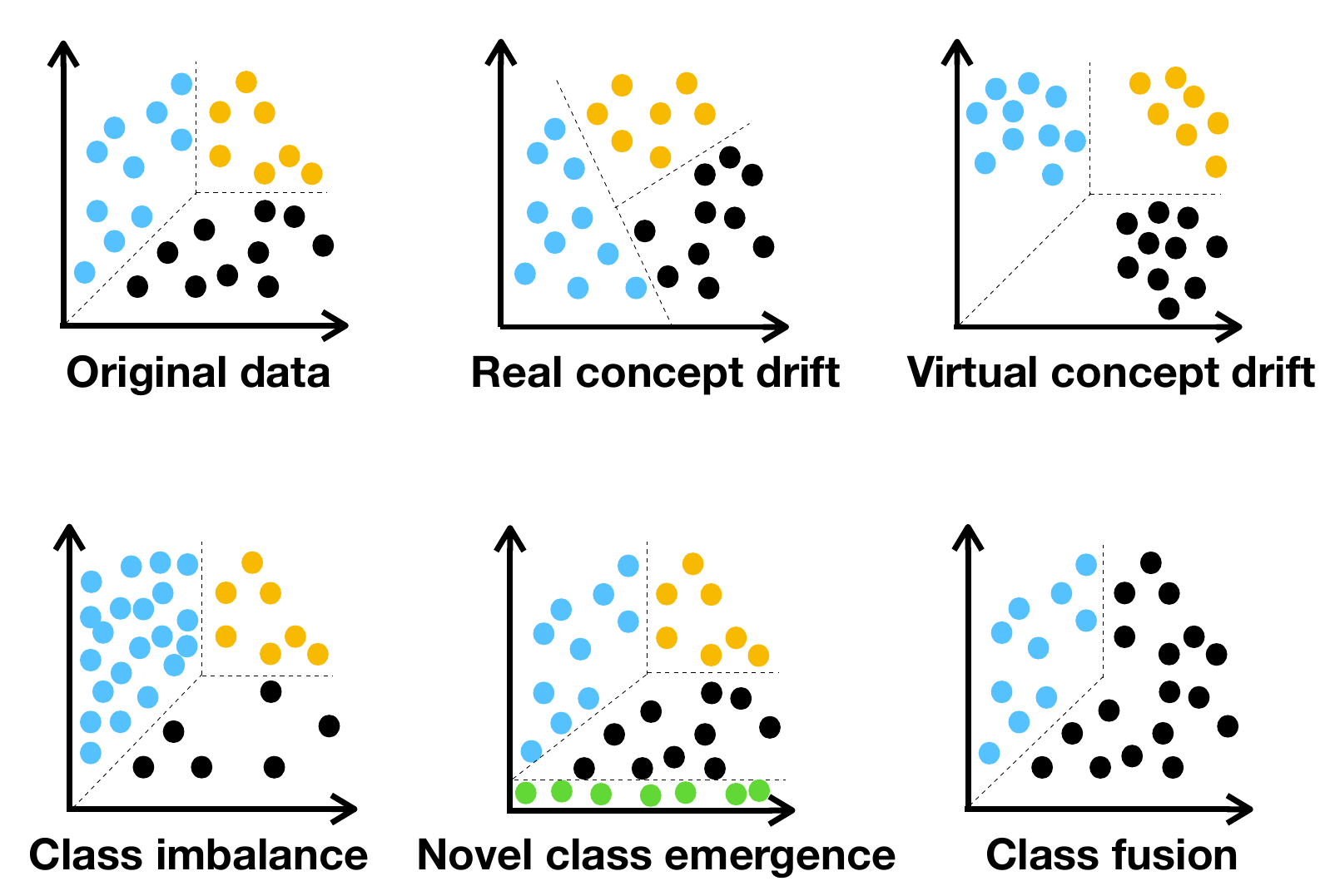}
    \caption{Some types of concept drifts in data stream frequently found in literature. Dashed lines indicate the separation margin between the classes.}
    \label{fig:drift_type}
\end{figure}

Perceiving the occurrence of a concept drift can carry different meanings and consequences depending on the application. For instance, in an application where there is interest in detecting new classes of data, a change where a new cluster of data points appears may represent the emergence of a novelty. Fig.~\ref{fig:drift_type} illustrates different practical types of drift found in data stream literature. In this example, circles represent instances, and colors represent classes. The figure also shows the decision boundary that discriminates the classes.

\subsection{Data Stream Realization}
One point of concern is the nature of the sequence and how it translates to the divergences in how consecutive instances are observed.

In certain sequences, the last observed instance is a transformation of previous instances and therefore \textit{could} not have existed before their materialization \citep{harries1999splice,zhu2010stream}. This is the case of most data from time-series problems. In a considerable amount of them, the observed data are complete, \emph{i.e.}, we observe all instances of the problem, and the practical objective is to predict future readings according to the data trend. In that sense, each individual observation can be considered of little importance. We highlight that in those cases, the feature-values registered for each observation are highly dependent on previous observations. For that reason, these data are the most affected by temporal dependence. We name sequences under this scenario \textit{materialization sequences}.

We recall that instances are distributed according to a background concept that evolves over time. However, in certain cases, the order in which instances were actually observed does not imply the necessary order of their materialization, that is, the order in which the instances started to exist \citep{vzliobaite2011combining,ikonomovska2011learning,katakis2009adaptive,souza2015data,Denis-KDD-2016}. One example is the data collected by a mosquito trap that measures flight characteristics of insects using sensors. While multiple insects may coexist in the trap vicinity at the same time, the ones that fly into the trap do so in a somewhat randomized order. However, all those insects have their behaviors affected by shared environmental factors, which change over time. In most cases that follow this structure, the observed data are only a sample of a more extensive set of instances that may not ever be observed, and the practical objective is to determine the class of each instance. In that sense, each observation is considered of high importance. We name sequences under the described setting \textit{observational sequences}.

The arrangement of data points in a sequence is commonly tied to time, be it the order in which data points were observed or materialized in the world. We call sequences that have their arrangement tied to time \textit{temporal sequences}. However, not all streaming data are tied to the chronological order of events.

Therefore, another relevant aspect of data stream is the physical nature of the sequence. Frequently, a stream is not chronological even though there is a logical sequence \citep{blackard1999comparative}. One example is the analysis of the pavement quality of a road \citep{souza2018asphaltCID, souza2018asfault}. The extent of the road can be split into sections that are analyzed individually, and the order of such sections can follow their spatial positions. Therefore, the resulting sequence follows a logical sequence, yet the actual time when data for each section were collected is irrelevant and interchangeable. When the order of the instances is related to their spatial disposition, we call the resulting sequence \textit{spatial sequence}. Finally, sequences not tied to time nor space are called \textit{logical sequences}.

\begin{figure}[htb]
\centering
   \includegraphics[scale=0.3]{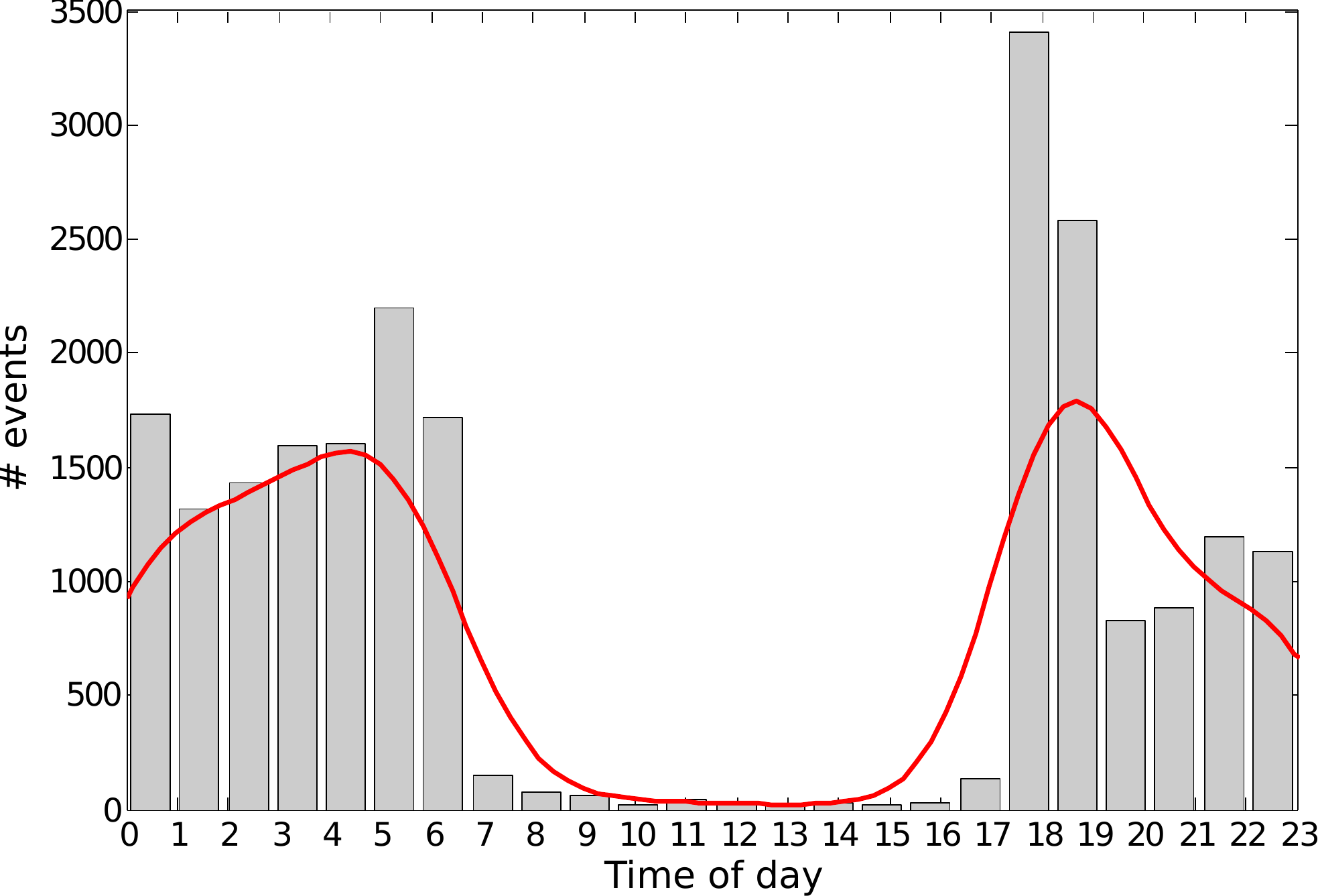}
   
         \caption{Circadian rythm of \textit{Culex quinquefasciatus}~\citep{souza2016classificaccao}. Each bar represent the amount of insect passages over the trap's sensor given a time of day.}
   \label{fig:circadian_rythm}
\end{figure}

When the concept behind the instances in a sequence is tied to either time or space, a relevant aspect is the spacing between instances. In typical materialization and spatial sequences, instances can generally be observed at regular time/space intervals. However, there are cases where the time/space between consecutive observations vary, and this setting poses particular complications for observational sequences. For example, in the mosquito trap application mentioned above, we know that different species of flying insects show more or less activity according to their circadian rhythm \citep{shinkawa1994variability}. Fig.~\ref{fig:circadian_rythm} illustrates the circadian rhythm of \textit{Culex quinquefasciatus} mosquitoes, measured by the trap's sensor over a week.

We can see that mosquitoes of this species abruptly become inactive in the dawn and resume their activity in the dusk. Observations done by the trap rely on the activity of insects so that it is improbable to collect data while specimens are inactive. However, even if the trap is not making observations, time passes and environmental condition changes, and therefore the flight characteristics of insects also changes. If the sequence does not have timestamps for each observation, we could probably observe an abrupt and inexplicable change of behavior for \textit{Culex quinquefasciatus}. On the other hand, with timestamps, we can notice that we lacked data for a prolonged period in which the behavior might have changed.

The example mentioned above illustrates the importance of temporal or spatial marks to understand concept drifts. In similar cases where the order of observations defines the sequence of the stream according to when they were made, if the observations are not evenly distributed and are tied to the temporal or spatial progression of the background environment, it is essential to include timestamps or longitudinal data to understand changes. 

For the sake of completeness, there are cases where data are considered to be a stream only due to its long length, although it lacks logical ordering \citep{cattral2002evolutionary,vergara2012chemical}. If an instance is as likely to be observed at time $t$ as at time $t'$, $t \neq t'$, $\forall t,t'$, then there is no concept drift. In fact, any two windows of data are going to be equally distributed, since both are uniform samples of the data. In such cases, forgetting mechanisms to discard old data is not beneficial, but should not be unfavorable to the performance of the classifier either. We call sequences under this setting \textit{unordered sequences} \citep{cattral2002evolutionary}.

\subsection{Stream Classification}
Classification is probably the most common task in data mining and a topic of active research in data stream~\citep{street2001streaming, bifet2013pitfalls, souza2015data, gomes2017adaptive}. Classification is the process of inducing a general model from previously known data (training data), and then using such a model to predict class labels (discrete values) for previously unknown data objects (test data). Differently from batch learning, in data stream or online learning, the test examples arrive continuously in an orderly fashion over time, and a classifier should generally predict the label of each instance in real-time or at least before the arrival of the next example.

In classification, the objective is to build a model that approximates the true relation between instances ($\mathcal{X}$) and their respective class-labels ($\mathcal{Y}$) and, therefore, is capable of predicting the class-labels of unlabeled instances. In other words, we want to build $h: \mathcal{X} \rightarrow \mathcal{Y}$ such that $h(\vec{X}_t) = y_t$. The task of inducing $h$ from labeled instances is referred to as \textit{training}.

In batch learning, the classification model $h$ is typically induced beforehand using a training set of labeled instances. However, in data stream classification, several approaches differ significantly regarding which data are used for training.

A recent approach assumes that labeled data for all possible concepts are available before the stream begins to be processed~\citep{reis2017:sac, moreira2018classifying}. In that case, individual classifiers are trained for each concept without regard for the data stream at first. Such models are only later \textit{deployed} to classify examples in the stream. However, there is not a single attribute that can easily identify, which is the concept of current data. Therefore, we need means to detect, which is the adequate classifier to be used for recent data.

Nevertheless, the most common setting in the data stream community is that there is no separate training set to induce a definitive model $h$. Thus, the model needs to be constructed and updated on the fly as new data are observed \citep{vzliobaite2015evaluation}. Among approaches in this setting, some methods incrementally evolve the classification model but do not adapt to changes in data. Such methods are capable of dealing with the fast pace of streaming data and keep low usage of memory. One example is the Very Fast Decision Tree -- VFDT~\citep{domingos2000mining}, where new incoming examples update the statistics of the leaves of a tree-based model. As more instances are observed, the statistics are used to decide which and when leaves are split into new leaves, resulting in the growth of the tree. Recently, VFDT was adapted as the newly introduced Extremely Fast Decision Tree to become faster \citep{manapragada2018extremely}.

However, due to changes in data distribution, the learner must incrementally adapt its model $h$ over time or perform updates when necessary to maintain stable predictive performance. Therefore, we also have a sequence of models $h_1, \ldots, h_i$, which can be discarded or reused in recurring situations. There are two main approaches to deal with concept drifts in classification problems \citep{khamassi2018discussion}: $i)$ evolving models, and $ii)$ adaptive models. \citet{gama2014survey} names such approaches of $i)$ blind, and $ii)$ informed.

The evolving models update the learner at regular intervals without considering whether changes have occurred. To do so, the model uses mechanisms for learning new concepts and forgetting old ones. A common approach is a sliding window with a fixed or variable length to store training examples or by weighting the data by age/utility~\citep{klinkenberg2004learning}. Such strategies consider that the most recent data are more representative of the current concept, so we can discard old data or assign them less weight. The main weakness of this approach is that the forgetting of old concepts is carried out at a constant speed for the whole time. Therefore, old data are discarded even when changes are not happening.

The evolving methods are naturally able to handle gradual and incremental drifts. Examples of work that implement evolving approaches are the algorithms from the FLORA family \citep{widmer1996learning}. In FLORA2, incoming examples are added to the window, and the oldest ones are deleted. A naive approach that falls into the same category is periodically retraining a new classification model with the last observed instances. Besides, some algorithms recently proposed, such as COMPOSE \citep{dyer2014compose} and SCARGC \citep{souza2015data}, use the sliding window strategy to deal with incremental changes in scenarios in which the actual labels of test instances are never available to the learner.

Adaptive models explicitly detect concept changes using drift detectors, updating the model only when changes are flagged. One of the advantages of explicit detection is the production of information about the dynamics of the data generation process and the reduced amount of updates in conditions without concept changes. An example of relevant work that uses drift detection is the extension of VFDT called Concept-adapting Very Fast Decision Trees -- CVFDT \citep{hulten2001mining}. CVFDT works by keeping its model consistent with a sliding window of examples. However, it does not need to learn a new model from scratch every time a new example arrives; instead, it updates the sufficient statistics at its nodes by incrementing the counts corresponding to the new examples and decrementing the counts corresponding to the oldest example in the window (which now needs to be forgotten). If the concept is changing, some splits that previously passed the Hoeffding test will no longer do so, because an alternative attribute now has higher gain. In this case, CVFDT begins to grow an alternative subtree with the new best attribute at its root. When this alternate subtree becomes more accurate on new data than the old one, the old subtree is replaced by the new one.

According to \citet{khamassi2018discussion}, the main issues of the approaches that use drift detectors are related to knowing $i)$ how to track concept drift, $ii)$ which data to keep and which data to forget, and $iii)$ how to adapt the learner parameters and structure to react according to the requirements of the new environment.

\subsection{Drift Detection}

Drift or change detection is the task of identifying significant data distribution changes in a stream. Although drift detection is a common mechanism of adaptive stream classifiers as a trigger for model updates, it is also a separated task from the classification process that contributes to other real applications as those related to signal analysis or time-series. 

For example, change detection can be used to provide alerts when the value of a stock is falling in the market \citep{oh2002analyzing} or identifying a fault in the monitoring of industrial processes \citep{venkatasubramanian2003review}. An important application of change detection methods is in burst detection \citep{gama2010knowledge}. Burst regions are time intervals in which some feature values are unexpected. For example, gamma-ray burst in astronomical data might be associated with the death of massive stars; bursts in document streams might be valid indicators of emerging topics, and so on.

In classification problems, drift detection methods are categorized into two major groups according to the availability of labeled data in the stream \citep{faithfull2019combining}: $i)$ supervised, and $ii)$ unsupervised. Supervised drift detection methods assume the immediate availability of class labels of each instance. These methods surveil indicators of classification performance, such as accuracy to detect drifts. On the other hand, when the class labels are delayed or unavailable, unsupervised methods detect drifts by comparing data distributions at different time intervals.

Based on the taxonomy proposed by \citet{gama2014survey} of dimensions which characterize drift detection methods, we consider three main categories:

 \textbf{I) Methods based on differences between two distributions.} In this approach, the methods monitor the distributions of two data windows. We can consider a reference window with old data and a detection window composed of recent data. These windows are compared using statistical tests, with the null hypothesis that the data of both windows are drawn from the same distribution. Thus, a concept drift is flagged when the test rejects the null hypothesis. The windows can contain unsupervised information as the raw data, learner parameters, classifier's outputs such as probabilities estimate or classification scores, as well as supervised information such as the error rate of the classifier or even the class labels.

Some parameters are fundamental to the success of these methods, such as how to measure the change and how to determine the size of the windows. To measure the change, different types of statistical tests as univariate or multivariate and parametric or non-parametric can be employed. Examples of tests are the Kullback-Leibler divergence \citep{dasu2006information}, Hotelling's $t^2$ \citep{hotelling1992generalization}, semi-parametric log-likelihood -- SPLL \citep{kuncheva2013change}, Kolmogorov-Smirnov \citep{Denis-KDD-2016}. Regarding the window size, it is important to note that a window smaller than the changing rate may lead to a higher number of false negative detections, and a window larger than the changing rate may delay the detections. Most of the work done is based on fixed-size windows, where delay in detections is frequent \citep{ganti1999framework, kifer2004detecting, dasu2006information}. Other pieces of work consider windows with dynamic size. For example, ADWIN \citep{bifet2007learning} finds two windows of different sizes through multiple tests between consecutive sub-windows within a window with fixed and large enough size. To detect drifts, ADWIN uses the Hoeffding bound to compare the sub-windows. Similarly to ADWIN, SEED \citep{huang2014detecting} uses two sub-windows with dynamic sizes but also performs block compressions to reduce the number of window comparisons. It also computes the volatility shift to describe the relationship of proximity between consecutive drift points in the stream.

\textbf{II) Methods based on sequential analysis.} The method Sequential Probability Ratio Test -- SPRT \citep{wald1947sequential} is the foundation of detection methods such as CUSUM and Page-Hinkley \citep{page1954continuous}. To better understand SPRT, consider a subsequence of $N$ examples from the stream  $\mathcal{DS} = \{\vec{X}_1, \vec{X}_2, \ldots, \vec{X}_N, \ldots\}$ where the subset of instances $\mathcal{D}_1 = \{\vec{X}_1, \ldots, \vec{X}_w \}$ with $1 < w < N$ is generated from an unknown distribution $P_A$ and the subset $\mathcal{D}_2 = \{\vec{X}_{w+1}, \ldots, \vec{X}_N \}$  is generated from another unknown distribution $P_B$. A change is declared at time $w$ if the probability of observing examples under the distribution $P_B$ is significantly higher than the $P_A$. For this verification, SPRT tests the logarithm of the likelihood ratio considering the two distributions. The main difference compared with the approaches previously discussed is that the test of SPRT is made sequentially one by one with different values of $w$, until the decision to accept or refuse the null hypothesis that $P_A$ and $P_B$ are the same distribution. SPRT is a classic method proposed in statistics, and data stream applications still employ it to detect changes with competitive performance~\citep{faithfull2019combining}.

\textbf{III) Methods based on statistical process control.} For decades, the quality control of products in continuous manufacturing is made using standard statistical techniques called control charts. Different methods such as DDM \citep{gama2004learning}, EDDM \citep{baena2006early}, and EWMA \citep{ross2012exponentially} are based on these statistical techniques to detect changes in data stream. These drift detection methods consider the classification problem as a statistical process and monitor the evolution of some performance indicators, such as the error rate, to apply heuristics to find points of change. For example, the method DDM considers three different states for the classification error evolution: $i)$ \textit{in-control}, when the error is stable; $ii)$ \textit{out-of-control}, when the error is increased significantly as compared to the recent past; and $iii)$ \textit{warning}, when the error is increasing but has not reached the out-of-control state. The method stores the data in a short-term memory during the \textit{warning} state and rebuild the classification model with this data when the error state is changed to \textit{out-of-control}.

The method employs a set of rules considering the mean and variance of the Binomial distribution of the classifier's errors to define the threshold of the states. An advantage of this method is that the rate of a change can be measured according to the number of examples or the time between the \textit{warning} and \textit{out-of-control} states. In this case, short times indicate fast changes, while longer times indicate slower changes. Inspired by DDM, the EDDM also takes into account the distance between consecutive errors as opposed to considering only the error magnitude.

The work of \citet{gama2014survey}, \cite{ditzler2015learning}, and \citet{khamassi2018discussion} provide interesting reviews about different drift detection methods from the literature. Also, \citet{gonccalves2014comparative} performs an experimental comparison of drift detection methods.

\section{Stream Datasets from Literature}\label{sec:datasets_literature}

\cite{bifet2009new} note the difficulty of finding large real-world datasets for public benchmarking, especially with substantial concept drift. We would like to quantify how this difficulty impacts the variety of datasets used in data stream research. Therefore, we performed a broad literature review over the last two decades. We analyzed more than 150 papers from high-impact conferences and top-tier journals to find the most used datasets. Unlike batch learning, in which a few hundred static datasets are available for evaluation, the data stream learning community has supported their findings in approximately 15 real-world datasets. In what follows, we describe the most popular datasets.

\begin{description}
    \item \textbf{Electricity} \citep{harries1999splice}. This dataset probably is the most used for the tasks of stream classification and drift detection. The data are from the Australian New South Wales Electricity Market. Prices are affected by demand and supply, which are assessed every five minutes. The learning task is to predict a rise or a fall in electricity prices, given recent consumption and prices in the same and neighboring regions. The dataset contains 45,312 instances, eight attributes, and two class labels (UP and DOWN);
    
    \item \textbf{Forest Covertype} \citep{blackard1999comparative}. This dataset contains information about the forest cover type of 30 $\times$ 30-meter cells obtained from the US Forest Service Region 2 Resource Information System. It contains 581,012 instances, 54 attributes, and seven class labels related to different forest cover types.

    \item \textbf{Poker-hand} \citep{cattral2002evolutionary}. Each record of this dataset is a poker hand consisting of five playing cards drawn from a standard deck of 52. Each card is described by two attributes (suit and rank). The dataset contains 1,025,010 instances, 11 attributes, and 10 class labels related to a possible poker hand such as a \textit{one pair, two pairs, flush, full house}, among others;
    
    \item \textbf{Intrusion Detection} or KDDCUP99 \citep{tavallaee2009detailed}. This dataset is from the KDD Cup 1999 Competition. The MIT Lincoln Labs gathered such data for nine weeks. The data consist of raw TCP dump data from a local area network. The learning task is to build a predictive model capable of distinguishing between normal connections and intrusive connections such as DoS (denial-of-service), R2L (unauthorized access from a remote machine), U2R (unauthorized access to local superuser privileges), and Probing (surveillance and other probing) attacks. The original task comprises 24 training attack types. The full dataset has about five million connection records, but it is usual to consider a subset with only 10\% of the size; 

    \item \textbf{Airlines} \citep{ikonomovska2011learning}. This dataset is from the Data Expo Competition 2009. The dataset consists of flight arrival and departure records of commercial flights within the USA, from October 1987 to April 2008. The learning task is to predict whether a given flight will be delayed, given the information of the scheduled departure. The dataset contains 539,383 examples, seven attributes, and two class labels (Delayed and Not delayed);
    
    \item \textbf{Gas Sensor Array} \citep{vergara2012chemical}. The dataset was gathered from January 2007 to February 2011, totaling 36 months, in a gas delivery platform facility situated at the University of California, San Diego. It comprises recordings from six distinct pure gaseous substances: Ammonia, Acetaldehyde, Acetone, Ethylene, Ethanol, and Toluene.
    The dataset contains 13,910 instances, where each instance consists of the measurements of 16 chemical sensors attached to an array. For each instance, \textit{only one} of the gases is diluted in dry-air at a varying concentration at a time inside of a chamber with the sensor array. The chamber where the gases are measured avoids any interference of the dynamics of the gases to the measurements. Therefore, only the presence of the gases should induce the conductivity of the sensors. An updated version of the dataset includes, for each instance, the concentration of the gas \citep{rodriguez2014calibration}. A discrete number of concentrations was assessed for each gas. The amount and which concentrations were measured according to the gas type.  \textit{Drift} was expected in a class due to the difference in concentrations. However, the original dataset was not intended to be a streaming dataset, and each instance was sampled independently from the other ones. The dataset was originally divided into batches that do not even follow the same logical sampling order. The classification problem is to identify which gas is measured.

    \item \textbf{Luxembourg} \citep{vzliobaite2011combining}. This dataset was constructed using the European Social Survey 2002 -- 2007. The task is to classify a subject concerning the internet usage as \textit{high} or \textit{low}. A possible source of drift is internet usage change over time. The dataset has 20 features (31 after transformation of categorical variables) based on the answers to a survey questionnaire and 1,901 examples collected over five years;
    
    \item \textbf{Chess.com} \citep{vzliobaite2011combining}. This dataset comprises data from \emph{chess.com} portal. The data consist of game records of one player over a period from December 2007 to March 2010 comprising seven attributes such as start date of the game, speed of the move in days, number of moves until the end of the game, type of the game (personal, tournament, and championship), current rating, opponent rating, and piece's color. Each player has a rating, which changes depending on achieved results. A possible source of drift is the fact that a player develops skills over time, besides engaging in different types of tournaments and competitions. The rating and the type of game determine how the system selects an opponent. The task of this data is to predict if the player will win, lose or draw a game; 
    
    \item \textbf{Ozone} \citep{Dua:2017}. This data consists of air measurements collected from 1998 to 2004 at the Houston, Galveston, and Brazoria areas. The learning task is to predict the ozone level eight hours ahead of time. The dataset has 72 attributes, 2,534 examples, and two class labels (Ozone day and Normal day).
    
    \item \textbf{Sensor Stream} \citep{zhu2010stream}. This dataset contains environmental information (temperature, humidity, light, and sensor voltage) collected from 54 sensors deployed in the Intel Berkeley Research Lab. The whole stream contains information recorded consecutively over two months (one reading every 1--3 min). The learning task is to identify the sensor ID based on the sensor data.  This dataset contains 2,219,803 instances, five attributes, and 54 class labels;

    \item \textbf{Powersupply} \citep{zhu2010stream}. This dataset contains hourly power supply data from an Italian electricity company. The data were collected from two sources: power supplied from the main grid and power transformed from other grids. The stream contains 3-year data from 1995 to 1998, and the learning task is to predict which hour of the day (1 out of 24 possibilities) the current power supply belongs. The argument for concept drift is that it is mainly driven by season, weather, time of the day (e.g., morning and evening), and the differences between working days and weekends. This dataset contains 29,928 instances, two attributes, and 24 class labels;
    
    \item \textbf{Spam Assassin Corpus} \citep{katakis2009adaptive}. This dataset consists of email messages chronologically ordered according to their date and time of arrival. The learning task is to identify if an email contains spam or a legitimate message. In this problem, the authors consider the occurrence of abrupt and gradual drifts. For the first case, consider that the user can inform the machine learning system of email filtering about his/her interests by marking messages as ``interesting'' or ``junk''. For example, a user subscribed to a mailing list might suddenly stop to be interested in messages containing smartphone reviews just after the purchase of a device. A situation where both abrupt and gradual concept drifts can occur is the user regaining interest in topics that he has been previously interested in. The dataset has 9,324 examples and 97,851 attributes. There are two classes, legitimate and spam, with the ratio around 25\% of spam;

    \item \textbf{Rialto Bridge Timelapse} \citep{losing2016knn}. This dataset was built using images extracted from time-lapse videos captured by a webcam with a fixed position. The recordings cover 20 consecutive days during May -- June 2016, capturing ten colorful buildings next to the famous Rialto bridge in Venice. Each captured image was segmented to cover each building and generating ten different instances. The classification problem of this dataset is to identify the correct building. Continuously changing weather and lighting conditions affect the data representation over time. Each one of the ten classes of this dataset has 8,225 examples encoded in a normalized 27-dimensional RGB histogram, totaling 82,250 examples; 
    
    \item \textbf{Outdoor Objects} \citep{losing2015interactive}. This dataset was built from images recorded by a smartphone camera in a garden environment. The task is to classify 40 different objects such as balls, shoes, pliers, cans, among others. One hundred pictures were taken of each object under varying lighting conditions (sunny and cloudy), affecting the color-based representation, and from different distances and positions. Altogether 4,000 images were recorded and arranged in temporal order. The examples from this dataset are represented using a normalized 21-dimensional RG-Chromaticity histogram;
    
    \item \textbf{Keystroke} \citep{souza2015data}. It is a subset of the larger CMU dataset~\citep{killourhy2010did}, where 51 users type the password ``.tie5Roanl'' plus the \emph{Enter} key 400 times captured in eight sessions performed in different days. In the Keystroke data, the typing rhythm is used to recognize four different users. In this classification task, ten features are extracted from the flight time for each pressed key. The flight time is the time difference between the instants when a key is released, and the next key is pressed. This dataset contains 1,600 instances that incrementally evolve due to the users' practice;

    \item \textbf{NOAA Weather} \citep{ditzler2013incremental}. The dataset consists of weather measurements collected over 50 years at Bellevue, Nebraska by the National Oceanic and  Atmospheric Administration (NOAA). This dataset contains eight features: temperature, dew point, sea-level pressure, visibility, average wind speed, max sustained wind-speed, minimum temperature, and maximum temperature. The learning task is to determine whether it will rain or not. The dataset contains 18,159 daily readings of which 5,698 are \textit{rain} and the remaining 12,461 are \textit{no rain}.
    
\end{description}

Table~\ref{tab:literature_datasets} presents a summary of the characteristics of the datasets.

\begin{table}[htb]
\scriptsize
    \centering
       \caption{Characteristics of the main stream learning datasets available for evaluation. \textsuperscript{1} Details about ordering are not provided; \textsuperscript{2} Timestamps or spatial marks are not included.}
    \label{tab:literature_datasets}
    \begin{tabular}{lllllll}
         \textbf{Dataset} & \textbf{Instances} & \textbf{Features} & \textbf{Classes} & \textbf{Sequence type} & \textbf{Ordering} \\ \hline
          Electricity & 45,312 & 8 & 2 & materialization & temporal \\ 
          Forest Covertype & 581,012 & 54 & 7 & observational & spatial\textsuperscript{1,2}\\ 
          Poker-hand & 1,025,010 & 11 & 10 & observational & unordered \\ 
          KDDCUP99 & 494,021 & 41 & 23 & observational & temporal\textsuperscript{2} \\ 
          Airlines & 539,383 & 7 & 2 & observational & temporal \\ 
          Gas Sensor Array & 13,910 & 128 & 6 & observational & logical\textsuperscript{1}\\ 
          Luxembourg & 1,901 & 30 & 2 & observational & temporal \\ 
          Chess.com & 534 & 7 & 3 & observational & temporal \\ 
          Ozone & 2,534 & 72 & 2 & materialization & temporal \\ 
          Sensor Stream & 2,219,803 & 5 & 54 & materialization & temporal \\ 
          Powersupply & 29,928 & 2 & 24 & materialization & temporal \\ 
          Spam Assassin & 9,324 & 97,851 & 2 & observational & temporal \\ 
          Outdoor & 4,000 & 21 & 40 & observational & temporal \\
          Rialto & 82,250 & 27 & 10 & observational & temporal \\
          Keystroke & 1,600 & 10 & 4 & observational & temporal\textsuperscript{2} \\
          NOAA Weather & 18,159 & 8 & 2 & materialization & temporal \\ 
          \hline
    \end{tabular}
\end{table}

The small number of real-world datasets publicly available impose restrictions on comparative studies of new proposals~\citep{krawczyk2017ensemble}. Such a lack of benchmark data leads to the use of approaches to simulate changes in static data or the generation of synthetic data with concept drifts.

Some common approaches to simulate changes in real data with static distribution are~\citep{sobolewski2013concept}:

\begin{itemize}
    \item \textit{Switching the features.} To simulate concept drifts, we can switch the values of some features while maintaining the class labels of a set of data samples~\citep{ ramamurthy2007tracking,  vzliobaite2009determining}. For example, given a static dataset, we first split it into two samples. In the second sample, the original feature 1 replaces feature 2, the original feature 2 replaces feature 3, and so on, while the last feature substitutes feature 1. The class labels of the examples remain the same;
    \item \textit{Swapping classes}. In this approach, we randomly pick two classes in the data set and swap their labels~\citep{klinkenberg2000detecting, kuncheva2008nearest}; 
    \item \textit{Joining classes.} We can join two or more classes in a unique class and consider this one as a new concept in the stream~\citep{vreeken2007characterising}.
    \item \textit{Reordering the data according to a hidden feature.} We can hide a feature and use it as a shared concept for the instances. In that case, we reorder the instances to group instances within the same concept together. Within the same concept, instances may be drawn uniformly to remove any other possible source of drift. If the hidden feature is nominal, the concept drifts are usually abrupt. In the case of an ordinal hidden feature, we can simulate incremental drift. Numeric features can be turned into ordinal features so that we can more easily draw instances from the same concept \citep{moreira2018classifying}. This approach suits observational sequences better than materialization sequences.
\end{itemize}

Some examples of synthetic data generators widely used by the community are  STAGGER~\citep{schlimmer1986incremental}, SEA~\citep{street2001streaming}, Rotating Hyperplane~\citep{hulten2001mining}, Random RBF~\citep{bifet2009new}, LED~\citep{breiman1984classification}, and Waveform~\citep{breiman1984classification}. We can also cite the framework proposed by~\cite{narasimhamurthy2007framework}, the Sine, Line, Plane, Circle, and Boolean datasets proposed by~\cite{minku2010impact}, and the synthetic datasets generated by~\cite{dyer2014compose} and \cite{souza2015data} to evaluate incremental changes.

The main problem of simulating drifts in real data or the use of generators is the introduction of data bias in the experimental evaluation. Data bias is the conscious or unconscious use of a particular set of data to confirm the desired finding, and that can lead to incorrect conclusions~\citep{keogh2003need}.

\section{Criticisms to Current Datasets for Stream Learning}\label{sec:datasets_criticisms}

In addition to the reduced number of publicly available real-world stream datasets, the most used datasets have problems such as a limited number of events and a small number of concept drifts. Unfortunately, these issues can lead to biased or incorrect conclusions when assessing the performance of stream algorithms. In this section, we discuss these problems and possible consequences.

\subsection{Uncertainty about Changes}

One of the main problems regarding the existing stream datasets is the uncertainty about the presence of concept drifts. The community frequently assumes that the performance degradation of a classifier over time is evidence of changes in the data distribution. However, concept drift is not the only cause of performance degradation, that might have other origins such as poor generalization (e.g., underfitting or overfitting~\citep{domingos2012few}) and noisy data arriving along the stream.

Even for datasets with known presence of concept drifts, the type of change (covariate, probability, or concept shift), pattern (abrupt, gradual, incremental, or reoccurring) and the exact moment these drifts occurred are frequently unknown. 

The lack of knowledge about change characteristics and when they occur can limit the evaluation of stream algorithms. A straightforward example is the evaluation of change detection methods that use criteria such as the probability of correct change detection, the probability of false alarms, and the lag of detection~\citep{gama2014survey}. Due to the lack of annotation of drift location in real data, the analysis of methods such as EDDM~\citep{baena2006early}, appropriated for slow, gradual changes, and EWMA~\citep{ross2012exponentially}, fit to abrupt changes, is only possible with the aid of artificial data. Finally, the use of inappropriate datasets for the problem tackled can lead to incorrect conclusions. One example is the use of a dataset where changes follow $P_A(Y|X) \neq P_B(Y|X)$ and $P_A(X) = P_B(X)$ to evaluate unsupervised detection algorithms.

Virtually all publications that present real data make informal assumptions regarding the existence of drift. As far as we know, \cite{quantifyingLGCDrift} is the first to make an effort to quantify their assumptions. Although their work is limited to the settings where drift is given by the incremental and spatial displacement of the classes in the feature space, such assumptions are valid for a broader number of existing work \citep{dyer2014compose,souza2015data}. \cite{quantifyingLGCDrift} introduce supervised means to measure the intensity of the displacement of the classes, its direction, and, more importantly, its unsupervised traceability.

\cite{webb2018analyzing} further raise awareness of the importance of measuring drift in datasets, introducing the task of \textit{concept drift mapping}. Particularly, they measure the divergence between consecutive snapshots (built with observation windows) of the data to represent the distributions of concepts over time. The divergence between concepts is called, in this scenario, \textit{drift magnitude}. The magnitude of the drift can be individually measured for \textit{marginals} ($P(X)$ and $P(Y)$) and \textit{conditionals} ($P(X|Y)$ and $P(Y|X)$) to provide different views of the data drift, revealing more information regarding its evolution. The authors also make comparisons between drifts on specific attributes and the total drift magnitude to highlight the contribution of different attributes to the drift. Although the drift magnitude was measured with total variation distance, any dissimilarity function that applies to distributions can be employed. Finally, we note that this work intends to provide tools to describe data, rather than mechanisms to detect drift actively.

\cite{goldenberg2018survey} review many applicable dissimilarity functions to verify which are good options for measuring drift magnitude. The work targets covariate shift (changes in $P(X)$) explicitly, suiting the task of unsupervised drift detection. The authors recommend Hellinger distance to measure the divergence between distributions of univariate and low-dimensional data.

When distributions are approximated by samples of numeric values, the use of Hellinger distance implies the discretization of the data with histograms. Many options have gone untested by \cite{goldenberg2018survey} and we refer the reader to \cite{gonzalez2017review} for more options and to \cite{cha2002measuring} for the particularly interesting ORD, which takes the distance between different bins into account. \cite{andre2019aaai} introduces SORD, a version of ORD that exempts the discretization of numeric values to compare univariate sample distributions, and can be seen as a particular and fast-to-compute case of the Earth Mover's Distance.

Another interesting aspect to know about the data is if they contain temporal overlap, as previously discussed in Section~\ref{subsec:typcd} and illustrated in Fig.~\ref{fig:concept:shift}. When it is absent, one can approach the classification problem with an incremental learner that need not implement a forgetting mechanism to discard old concepts or a system to switch between previous models. This scenario is significantly less challenging than problems that must deal with temporal overlap.

We suggest a naive approach to indirectly measure temporal overlap if the concepts are known, and data from each concept can be isolated. One can build a classifier for each concept and individually test their performance on their respective concept. An additional classifier should be built with data from all concepts and tested with a test set that also contains examples from all concepts. If the performances of the classifiers for individual concepts are, on average, superior to that of a single classifier that single-handles all concepts, we have evidence that there is temporal overlap. Otherwise, we have evidence that we do not need to use forgetting mechanisms and incremental learners.

\subsection{Temporal Dependence}
Nearly almost ten years after the first evaluation on the Electricity data in the stream setting by \cite{gama2004learning} and the use of these data by several studies (e.g., \cite{gama2005learning}, \cite{baena2006early}, \cite{bifet2009new}, \cite{bifet2010leveraging}, \cite{brzezinski2011accuracy}, \cite{chen2011towards},  \cite{ditzler2013incremental}, \cite{demvsar2018detecting}, and \cite{shao2018robust}), ~\cite{zliobaite2013good} pointed out the problems of this dataset related to the temporal dependence of class labels. 

Suppose we employ a naive classifier that predicts the next label to be the same as the current label. This classifier will be our baseline henceforth. For instance, if the price goes UP now, the baseline will predict that the price will go UP for the next time step as well. If the labels were independent, such a predictor would achieve 51\% given the class proportions of this particular dataset. However, if we test such an approach on the Electricity dataset as it is, we obtain a much higher accuracy of 85\%. Therefore, the labels are not independent, since there are long periods of consecutive UP and long periods of consecutive DOWN labels. 

\citet{zliobaite2013good} discusses the problem of temporal dependence for the Electricity dataset; however, another two popular datasets, Forest Covertype, and Poker-hand, also have the same problem. For example, in the Forest Covertype the data are probably organized according to the geographical location of the observations, although the dataset does not include annotations for the position. Thus, there is a high probability that neighboring regions are of the same class. In the Pokerhand, we have a more significant issue. The MOA's website provides a supposedly normalized version of the dataset and a link for the original version at UCI Repository. The issue is that,  besides not really having been normalized, MOA's version has a different ordering for the instances. While the original version does not present temporal dependence and the No-Change baseline achieves 43\% accuracy (which is the expected accuracy if there is no temporal dependence, given the proportion of the classes on this particular dataset), the same baseline achieves staggering 75\% accuracy on MOA's normalized version. We can only wonder whether this temporal dependence was purposefully implanted into the data and why. From now on, we will consider the MOA's version to illustrate the effects of temporal dependence better.

\begin{figure}[htb]
\centering
   \subfigure[Electricity]{
     \includegraphics[scale=0.222]{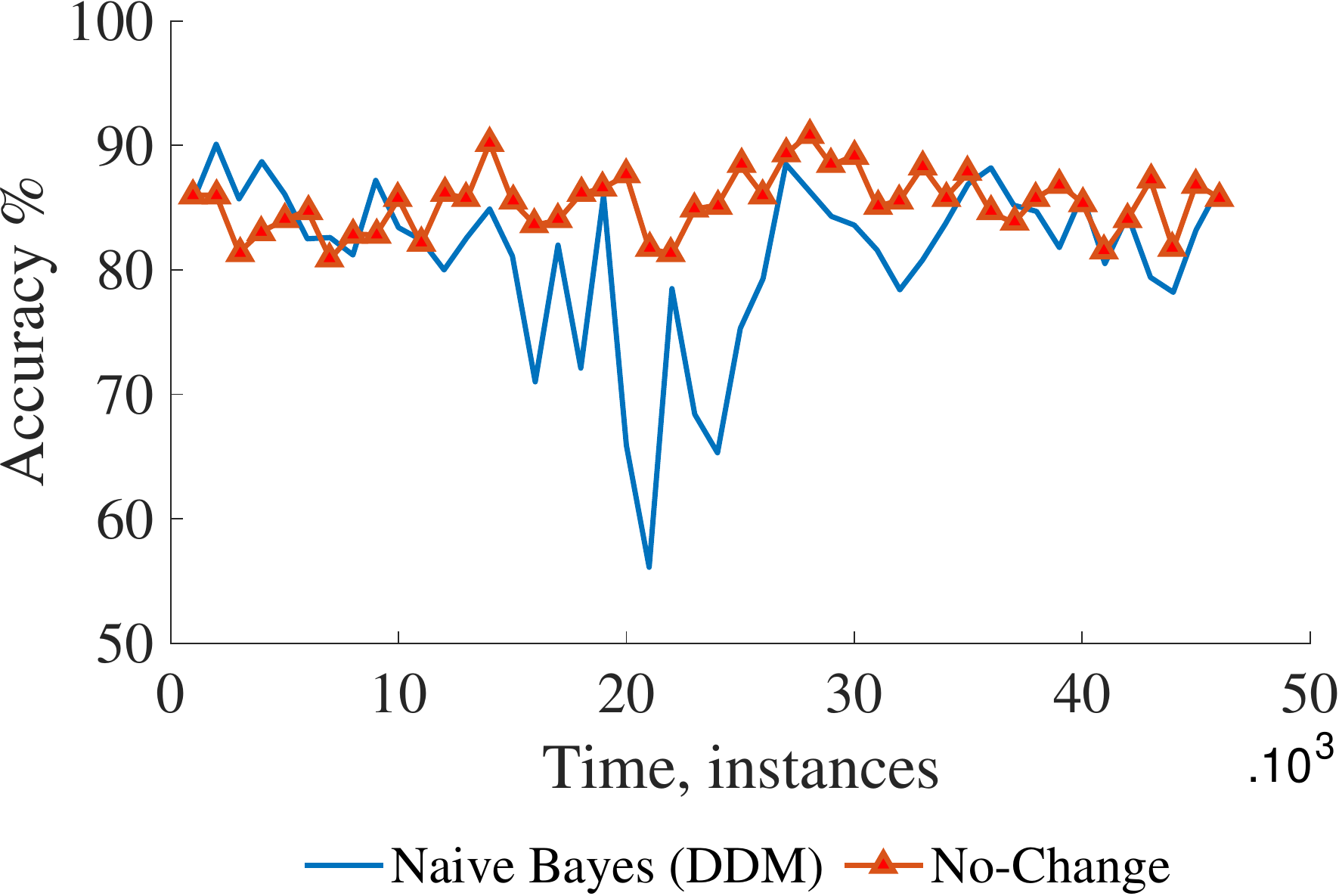}
   }
   \hspace{-0.4cm}
   \subfigure[Forest Covertype]{
     \includegraphics[scale=0.222]{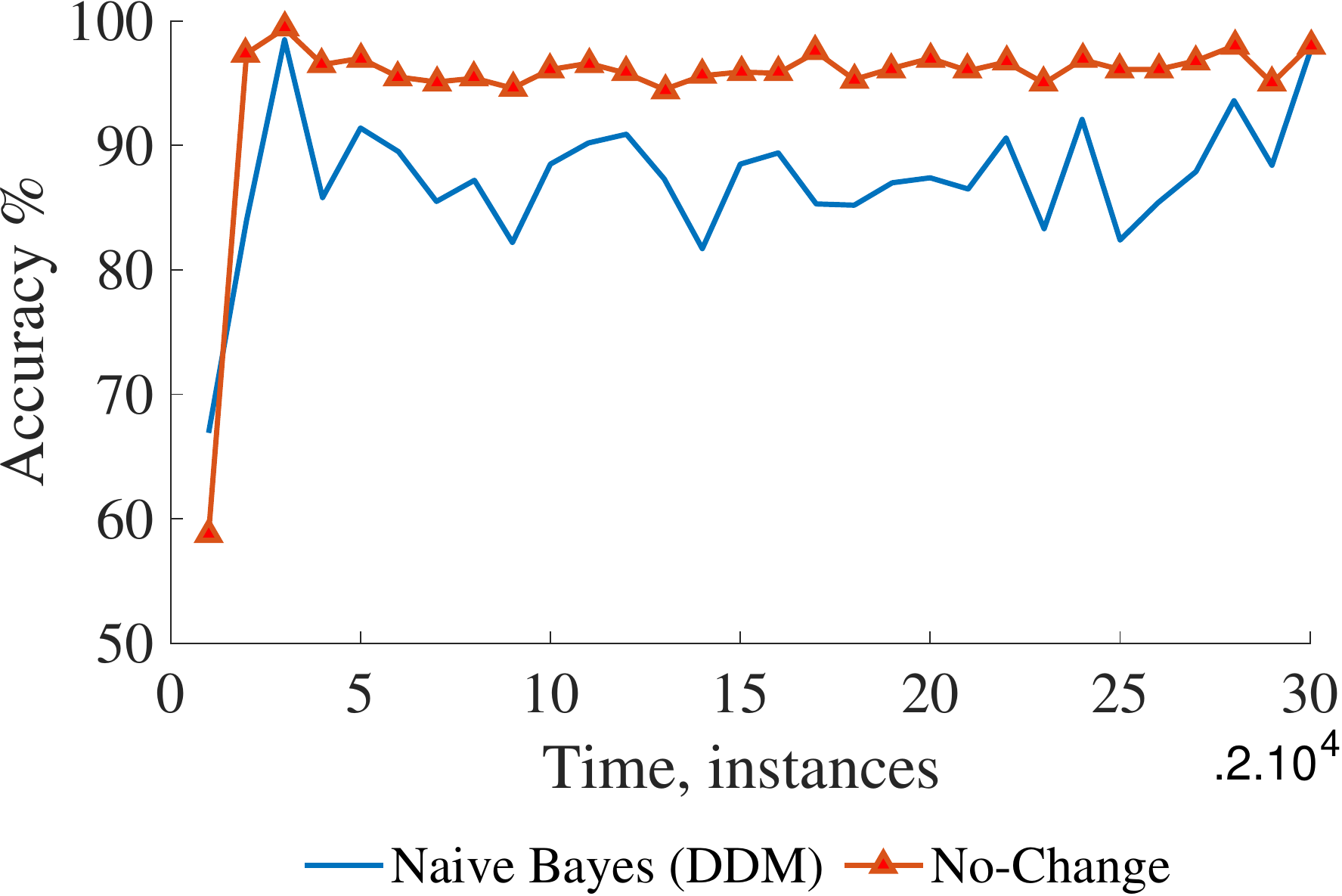}
   }
   \hspace{-0.4cm}
      \subfigure[Poker-hand]{
     \includegraphics[scale=0.222]{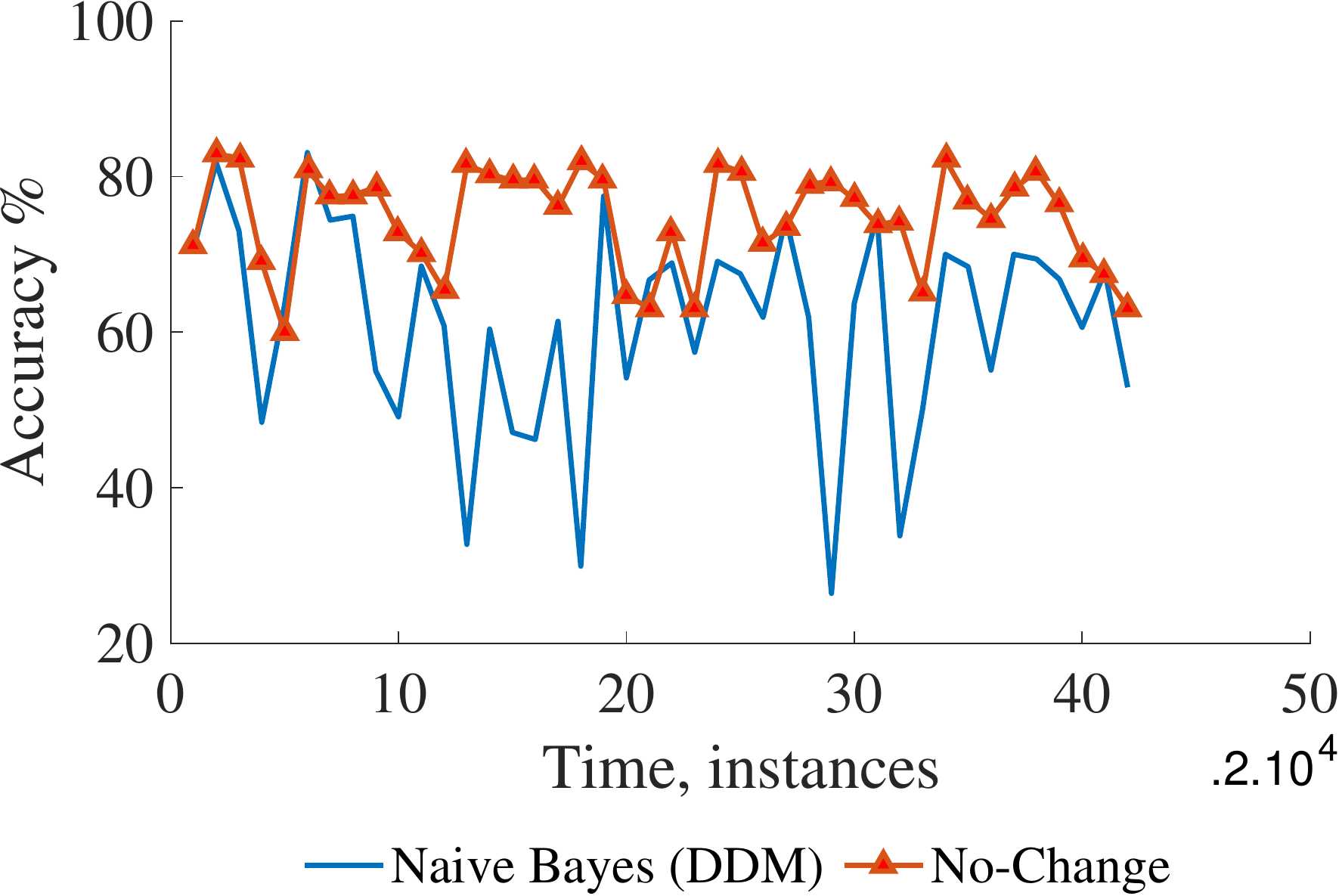}
   }
         \caption{Prequential accuracy of Naive Bayes with Drift Detection Method (DDM) and the baseline classifier No-Change on the Electricity, Forest Covertype, and Poker-hand datasets.}
   \label{fig:temporal_dependence}
\end{figure}

Fig.~\ref{fig:temporal_dependence} presents the prequential accuracy of two classifiers: $i)$ Naive Bayes with Drift Detection Method (DDM)~\citep{gama2004learning}, and $ii)$ the baseline classifier No-Change that predicts the current label to the next event for the three mentioned datasets. Given a sequential dataset, in the prequential procedure (or test-then-train), every example is first used for testing and then for updating the model.

In all cases, the baseline classifier surpasses the results of the classifier that detects changes and periodically updates their model. For Electricity data, the No-Change classifier shows an accuracy of 85.33\%, while the Naive Bayes with DDM achieves only 81.23\%. For Forest Covertype, the baseline presents 95.07\% of accuracy, and Naive Bayes with DDM has 88.04\%. In the Poker-hand dataset, the No-Change presents an accuracy of 74.51\%, and the Naive Bayes with DDM achieves just 61.96\%. 

In this sense, a new proposal that uses a solution based on the temporal dependence of the examples probably will show promising results on these data. However, such a good performance does not necessarily mean that the classifier has a good generalization power or it adapts well to changes.

\cite{bifet2013pitfalls} proposed a new evaluation measure (Kappa-Temporal) to avoid biased conclusions. Kappa-Temporal considers the difference between the prequential accuracy of a given classifier and the accuracy achieved by the naive classifier that ever predicts the last seen class label. However, we argue that this measure is a palliative solution to be used in the evaluation process of stream learning methods to mitigate the consequences generated by a characteristic inherent to some datasets. Further, we add that the baseline No-Change and the measure Kappa-Temporal are not well suited to compare with and evaluate classifiers that do not rely on true labels, like those that make use of unsupervised drift detection methods, since the baseline and the measure depend on such unavailable piece of information.

\subsection{Data Bias}
Due to the reduced number of real datasets, we frequently come across stream evaluations that consider three, two, or even only one real dataset together with a larger number of synthetic data. The main problem of this practice is the possibility of data bias, as previously discussed in Section~\ref{sec:datasets_literature}.

With a reduced number of datasets, we can demonstrate any findings we  wish~\citep{keogh2003need}. For example, consider the use of three datasets to compare the classification performance of the Naive Bayes algorithm with two different drift detectors: DDM and CUSUM~\citep{alippi2008just}. 

\begin{figure}[htb]
\centering
   \subfigure[Scenario 1]{
     \includegraphics[trim={0 0 0 1.2cm},clip, scale=0.565]{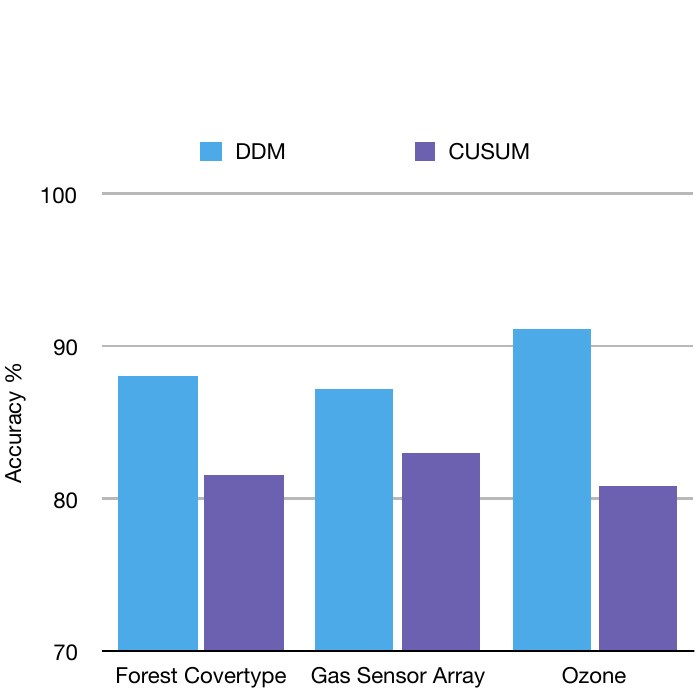}
   }
   \hspace{-0.2cm}
   \subfigure[Scenario 2]{
    \includegraphics[trim={0.3cm 0 0 1.2cm},clip, scale=0.565]{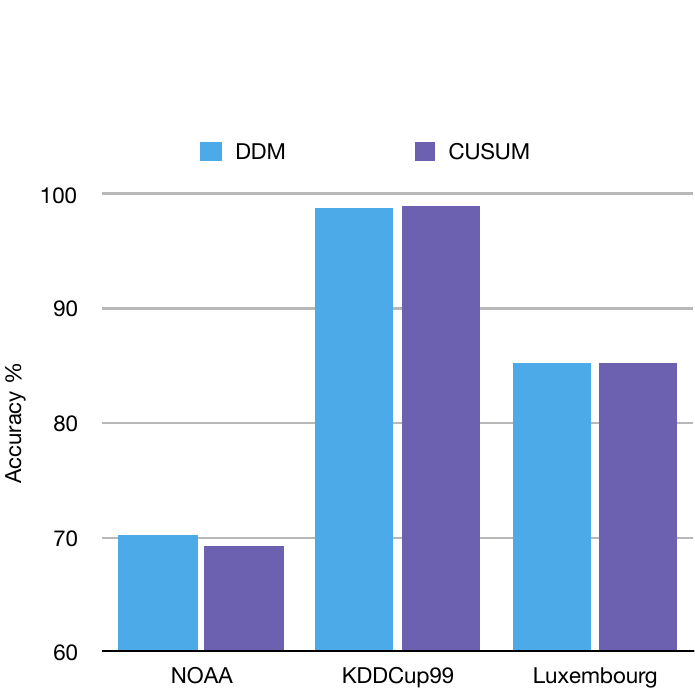}
   }
   \hspace{-0.2cm}
      \subfigure[Scenario 3]{
    \includegraphics[trim={0.3cm 0 0 1.2cm},clip, scale=0.565]{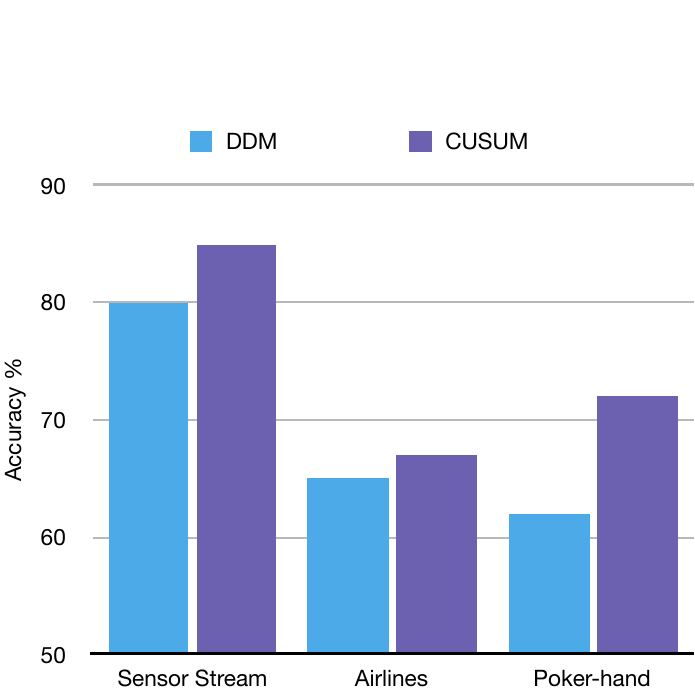}
   }
         \caption{In the first scenario (a), we consider Forest Covertype, Gas Sensor Array, and Ozone. In the second scenario (b), we consider  NOAA Weather, KDDCup99, and Luxembourg. In the third scenario (c), we consider Sensor Stream, Airlines, and Poker-hand. According to the evaluated datasets, our conclusions can may be biased.}
   \label{fig:data_bias}
\end{figure}

In the first scenario (Fig.~\ref{fig:data_bias}-a), if we consider the Forest Covertype, Gas Sensor Array, and Ozone datasets, our obtained results would suggest that DDM outperforms CUSUM. However, if we consider a second scenario with the datasets NOAA, KDDCup99, and Luxembourg, we would conclude that both methods perform very similarly (Fig.~\ref{fig:data_bias}-b). On the other hand, in a third scenario where we chose the datasets Sensor Stream, Airlines, and Poker-hand, we can conclude that the CUSUM outperforms DDM (Fig.~\ref{fig:data_bias}-c).

The results presented in Fig.~\ref{fig:data_bias}, allow us to state that the use of a reduced amount of datasets and the ``right'' choice of them can lead to biased conclusions. To avoid this problem, we claim by the need of a sufficiently large benchmarking data that covers different properties for stream learning, as it is already usual in batch learning.

Besides the reduced number of datasets, the typical procedure employed for evaluating the performance of adaptive learning models could also be responsible for leading to biased conclusions. As noted by \cite{vzliobaite2014controlled}, the standard procedure, named Prequential (or test-then-train)~\citep{gama2013evaluating}, allows processing a dataset only once in the fixed sequential order. The positions where and how changes happen remain fixed; thus, a single test concludes how well a model would adapt to this fixed configuration of changes. While different learning models have different adaptation rates, the results on a fixed test snapshot with a few changes may not be sufficient to generalize how this adaptive model would perform online on a given problem. To make the evaluation more confident, \citet{vzliobaite2014controlled} proposes the employment of multiple tests with variations of the original dataset. The various tests are generated by permuting the data order in a controlled way to preserve local distributions.

\subsection{Insufficient Amount of Instances}

Stream predictive models that operate in changing environments have different requirements from the traditional batch setting. The three main requirements are~\citep{gama2014survey}:
\begin{enumerate}
    \item Detect concept drifts (and adapt if needed) as soon as possible;
    \item Distinguish drifts from noisy data;
    \item Operate faster than the example arrival time and use a fixed amount of memory for any storage.
\end{enumerate}

Here, we call attention to the third requirement. As data stream is frequently defined as an infinite sequence of examples, accommodating such data in the machine's main memory is considered impractical or infeasible. However, this definition is inconsistent with the number of instances present in the commonly used stream datasets.

From the 16 popular stream datasets presented in Table~\ref{tab:literature_datasets}, only two of them have more than one million examples (Poker-hand and Sensor Stream). Further, more than half have less than 50,000 examples, an amount that can be handled by most of batch learning algorithms. In general, these numbers of examples do not represent a challenge to data processing and storage. One possibility is that researchers in the community might feel challenged enough to design memory-efficient algorithms. In reality, we have noticed that very few papers analyze the memory requirements, with some exceptions, such as the system streamDM-C++~\citep{bifet2017extremely}.

\subsection{Lack of Complex Distributions}

For many real-world applications such as financial fraud detection, natural disaster prediction, spam filter, fault monitoring, or disease diagnosis, we have an interest in events that occur with a very low frequency. In these cases, some classes are difficult or expensive to collect. Consequently, the classes are not equally represented in the data, which leads to the problem of \emph{class imbalance} or \emph{skewed class distributions}~\citep{batista2004study}. Class imbalance can cause a bias towards the majority class, and the classifiers may tend to misclassify minority class examples due to the poor generalization~\citep{wang2013learning}.

The machine learning community has widely researched the class imbalance problem for more than 20 years~\citep{chawla2004special}. However, this issue is still challenging and subject of intensive research in the static learning setup~\citep{yang200610}. Although class imbalance and concept drift are intimately related when changes occur in prior probabilities $P(Y)$, learning with class imbalance has still received little attention on stream learning~\citep{hoens2012learning, ghazikhani2013recursive, krawczyk2017ensemble}. As recently noted by~\cite{wang2018systematic}, most existing work in stream learning focuses on the concept drift in posterior probabilities (i.e., real concept drift or changes in $P(Y|X)$, as discussed in Section~\ref{subsec:cd}) and most proposed concept drift detection approaches are designed for and tested on balanced data streams.

Differently from static learning, in data stream setting the class distribution is not fixed. Instead, the class ratio varies, and the relationship between majority and minority classes may change over time. It becomes even more complicated in multi-class problems.

\begin{figure}[htb]
    \centering
    \subfigure[Electricity]{
    \includegraphics[scale=0.32]{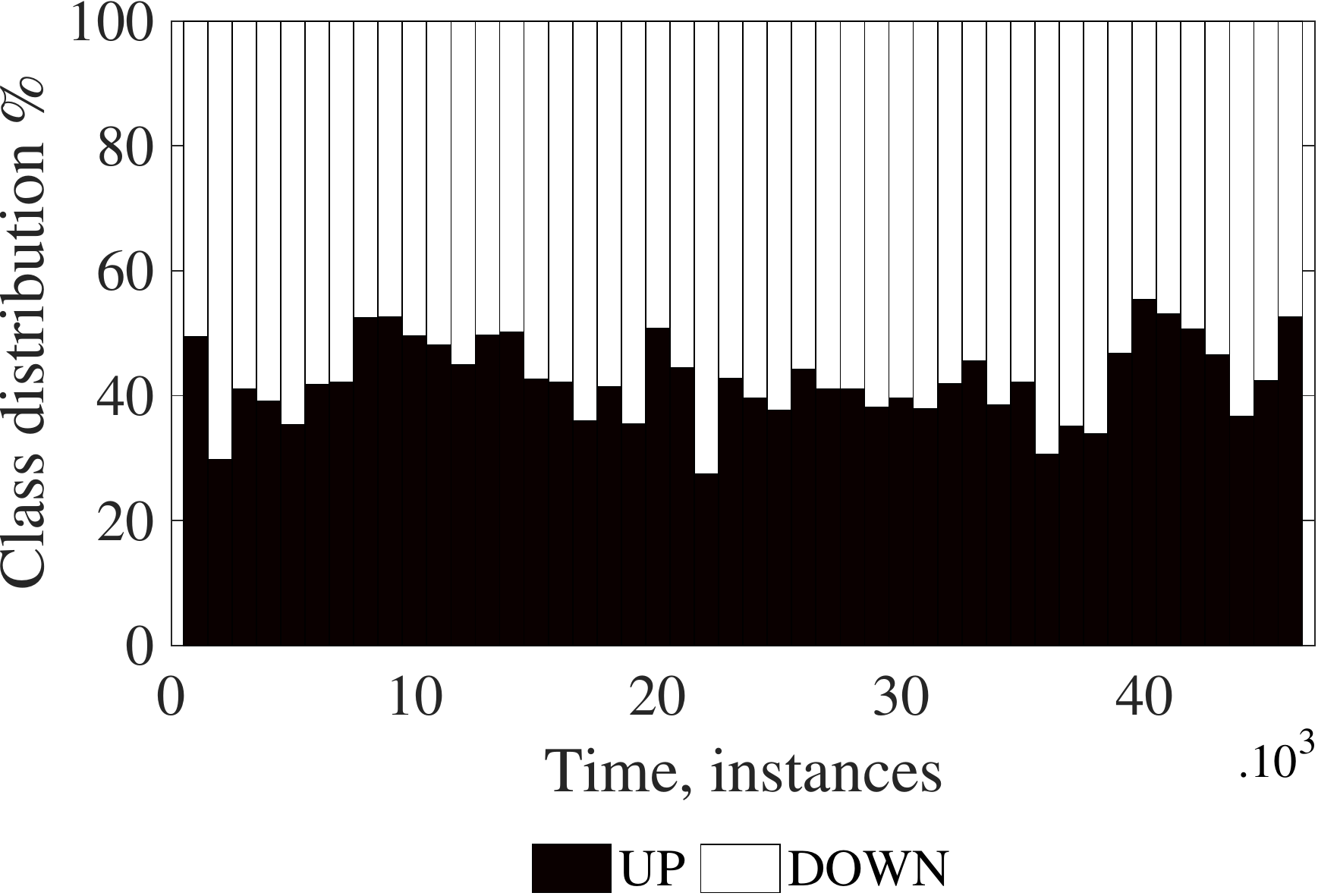}
    }
    \subfigure[NOAA Weather data]{
    \includegraphics[scale=0.32]{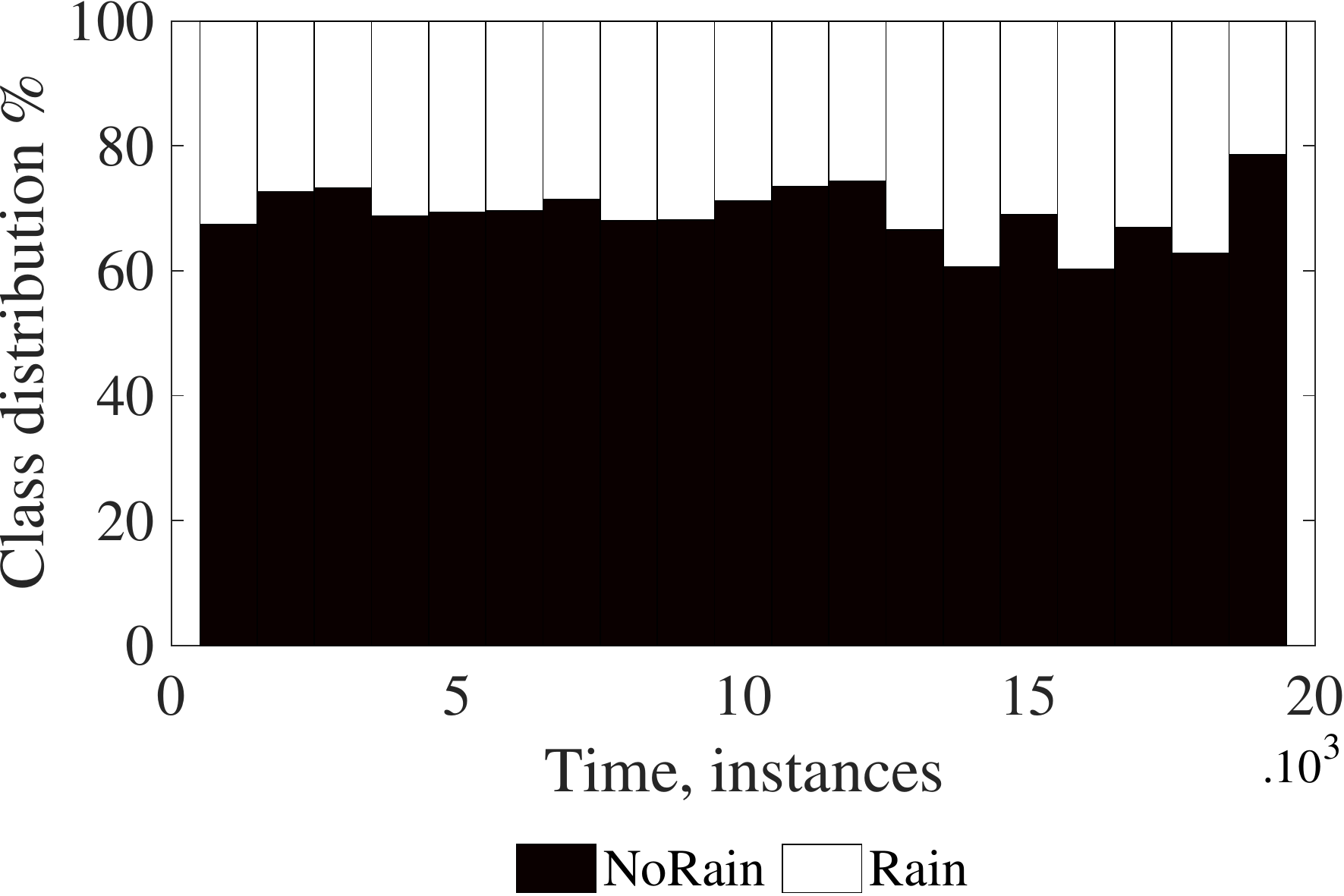}    
    }
    \caption{Changes in data distribution over time for the classes UP/DOWN from Electricity data and Rain/NoRain from NOAA Weather data. Each bar represents the counting of 1,000 consecutive examples from the stream.}
    \label{fig:elecnoaa_dist}
\end{figure}

We believe that the lack of real stream datasets with complex distributions that contain changes in both $P(Y)$ and $P(Y|X)$  (or $P(Y)$ and $P(X)$) limit the research and evaluation of data stream research in realistic scenarios. For example, Fig.~\ref{fig:elecnoaa_dist}-a illustrates the changes in the classes proportion in the Electricity dataset given a window with an arbitrary size of 1,000 instances. The class distribution barely changes over time. As this data does not contain class imbalance, an alternative is to under-sample one of the classes as proposed in the evaluation of Learn$^{++}$.NIE algorithm~\citep{ditzler2013incremental}. However, this practice can be interpreted as a modification in the real problem characteristics.

In recent work, \cite{wang2018systematic} proposes the use of  PAKDD 2009 credit card~\citep{linhart2009pakdd}, UDI Twitter Crawl~\citep{li2012towards}, and NOAA Weather as real-world datasets to evaluate different approaches for imbalanced class distributions on stream learning. 

The PAKDD data were collected from the private label credit card operation of a Brazilian retail chain. The task of this problem is to identify whether the client has a good or bad credit, where the ``bad'' credit is the minority class with 9,868 examples taking 19.75\% of the 49,973 examples. This dataset has gradual changes since a client with bad credit can improve their status by meeting their financial commitments over time. In the same way, a good client can stop paying their debts, changing the status to bad. In the UDI Twitter Crawl, the task is to predict the tweet topic. To build this imbalanced stream dataset was chosen a subset of 8,774 examples from the original data that include 50 million tweets posted from 2008 to 2011. Next, the problem was reduced for two classes by using only two out of seven possible topics. As noted by the authors, the tweet topic change can be much faster and more noticeable when compared to PAKDD 2009 data. Both datasets are important contributions. However, they present some of the previously discussed drawbacks, such as uncertainty about changes and an insufficient amount of examples. For the last data evaluated, NOAA Weather, although the majority class has 12,461 examples (68.62\%), this ratio is almost constant over time, as shown in Fig.~\ref{fig:elecnoaa_dist}-b. Thus, there is a need for representative stream data with more complex distribution changes over time to evaluate the problem of imbalanced classes better.

\subsection{Streaming as an afterthought}
One glaring aspect of the data that is often used to test data stream learners is that such data are not conceptually meant to this task. The most conspicuous example is Poker-Hand dataset. We first note that the size of the dataset is small enough (around 20 megabytes) so that it can be fed to batch learners. Last, and more importantly, the nature of any variant of a Poker game inherently grants an equal chance for every combination of cards to be drawn at any given moment. Hands drawn from a real deck of cards are independently and identically distributed so that the hands in a stream should not be bound to a background hidden, evolving concept, and there should not be temporal dependence. This means that none of the challenges that are defended to be present in streaming data actually happen in this dataset. This is reflected by the original version of Poker-hand, found at UCI Online Repository.

However, for reasons that are beyond our knowledge, the ``normalized'' version distributed at MOA's website has a different ordering for the hands and is biased to present temporal dependence. While this fact is not made clear on the website, we can suspect the reason is to make the dataset more challenging and interesting for benchmarking data stream algorithms, despite the ordering being unnatural. We are not against the use of reordering to repurpose a dataset to benchmark by any means. However, we call attention to two important issues: the unnatural ordering should be explicitly explained since it is the only source of streaming challenges and it is not present in the original data; and the only challenge is temporal dependence, which is still one of the least interesting problems to have in a streaming application.

Another dataset that has been repurposed for a streaming application is Gas Sensor Array. The data collection process reassured that each example is independent, involving the use of precision equipment to set the concentration of each gas before registering the measurements of the sensor array. Only a discrete number of sparsely distributed concentrations were tested, and the different gases (that are the class labels) are never mixed together. Instead, each gas is only diluted in dry-air for the measurement. Similarly to the normalized Poker-hand, the only aspect directly associated with stream data is the unusual ordering of the examples, which is not well explained.

A less blatant example of dataset not well suited for streaming problems is Forest Covertype. The order of the examples in the dataset is likely related to their physical position in the world, which means that consecutive examples are likely to share characteristics and class labels. However, there is an immeasurable number of ways of iterating over square cells in a region, each way with its own implications in the ordering of the examples, and consequently in the temporal dependence of the stream. Yet, how the specific ordering in the dataset was achieved is unspecified, and the data do not contain the geolocation of the examples. It is also debatable if a linear representation of such data is an appropriate approach for learning tasks.

\section{A Real-world Streaming Application with Concept Drifts}\label{sec:sensor}

In this paper, we introduce to the data mining community, a benchmarking dataset with different properties to evaluate stream classifiers and drift detectors. The dataset is based on a real-world streaming application based on the use of optical sensors to recognize flying insect species in real-time.

In the last years, our research group has been working in the next generation of electronic insect traps to selectively capture only certain species~\citep{batista2011sigkdd, de2013classification, chen2014flying, qi2015effective, silva2015exploring}. Such smart traps use Machine Learning techniques to recognize the insects that pass in front of the sensor. The trap selectively captures species of interest such as vectors of mosquito-borne diseases and agricultural pests, freeing all other species and, therefore, reducing the impact of this control device on the environment.

For this application, we cannot assume that a stationary stochastic process generates the data due to the existence of variations in environmental conditions that can influence the behavior of the insects. For example,  temperature influences the metabolism of insects~\citep{Taylor:63, villarreal2017impact} Also, ambient conditions such as air pressure~\citep{Chadwick:1949} and humidity~\citep{Mellanby:1936} can change their flying behavior. For these reasons, the data measured by the sensor suffers from concept drifts over time, requiring adaptive models to perform the classification task of insect recognition.

We present the details of the smart trap in Section~\ref{subsec:smart_trap} and the optical sensor used into the core of the trap in Section~\ref{subsec:sensor}. Section~\ref{subsec:data_collection} details the procedures of data collection using our sensor on changing environments. Section~\ref{subsec:features_extraction} presents the predictive features extracted from the insect signals. Finally, we introduce the proposed stream benchmark data in  Section~\ref{subsec:benchmark_description}.

\subsection{Smart Trap for Insects}\label{subsec:smart_trap}

Controlling insect pests and vector of diseases is an important task and the main focus of the active research in the last decades. Entomologists have proposed dozens of techniques from insecticides to biological control~\citep{medlock2012review}. However, these techniques can be made safer and more cost-effective with the knowledge of the spatial-temporal distributions of the insects in a certain area. 

Traps are the main tool for the surveillance of insect populations. For instance, sticky traps are used in crop fields, where they are installed and collected at regular time intervals. A human expert is required to manually classify each collected individual and count the species of interest. Although sticky traps are usually inexpensive in terms of material cost, the whole procedure is expensive since it involves manual counting and classification.

\begin{figure}[htb]
\centering
     \includegraphics[trim={4cm 0 3cm 0},clip, scale=0.3]{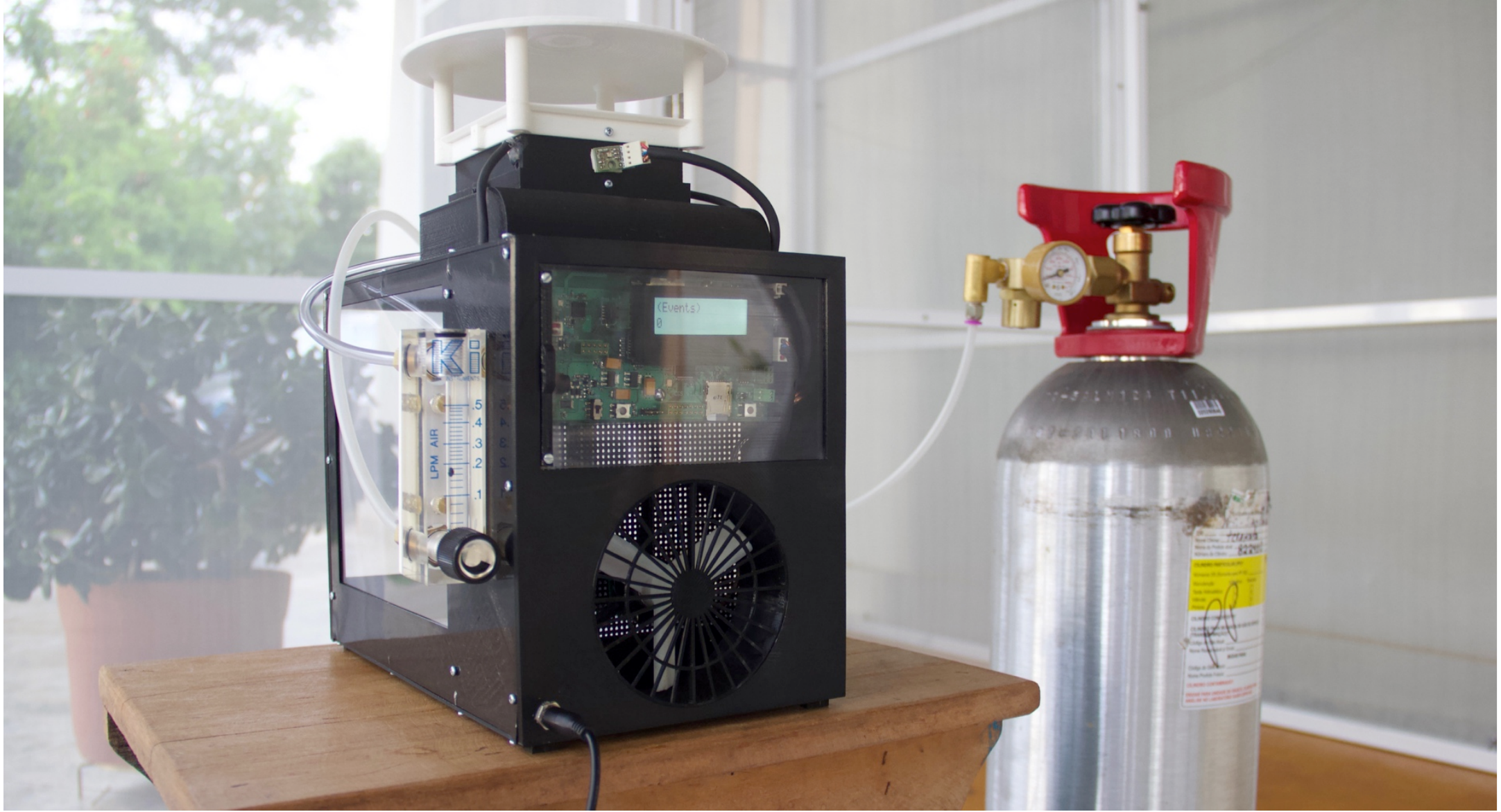}
         \caption{Smart Trap for counting and classifying mosquitoes in real-time using optical sensor.}
   \label{fig:smart_trap}
\end{figure}

The main advantage of smart traps such as proposed in our research is their capability of counting and classifying flying insects in real-time without requesting the time and cost of analysis made by experts. Also, differently from other traps, our device deliberately does not capture non-target species, such as pollinators insects. Fig.~\ref{fig:smart_trap} illustrates a recent prototype of our device. The trap turns a fan on and off at the moment it senses a mosquito near the sensor, significantly reducing the power consumption.

To classify flying insects in real-time, the trap combines the optical sensor to measure the light variation over time and a circuit board to filter and record data, as well as to extract predictive features which are used by a supervised machine learning classifier.

\subsection{Sensor to Measure Insect Flying Data}\label{subsec:sensor}

\begin{figure}[htb]
\centering
   \subfigure[Side-view]{
     \includegraphics[scale=0.33]{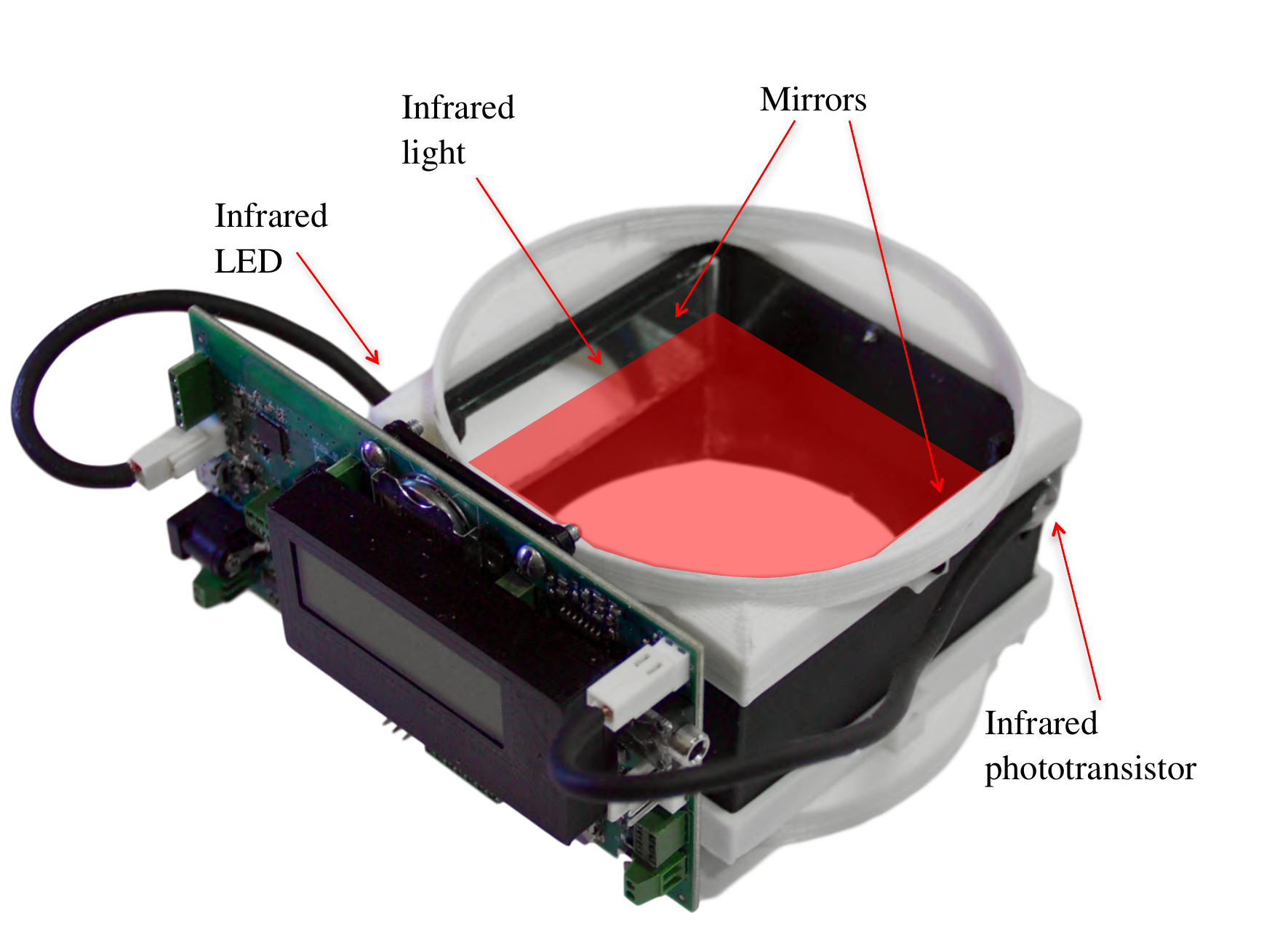}
   }
   \hspace{-0.7cm}
   \subfigure[Top-view]{
     \includegraphics[trim={0.6cm 0 1.2cm 0},clip,scale=0.37]{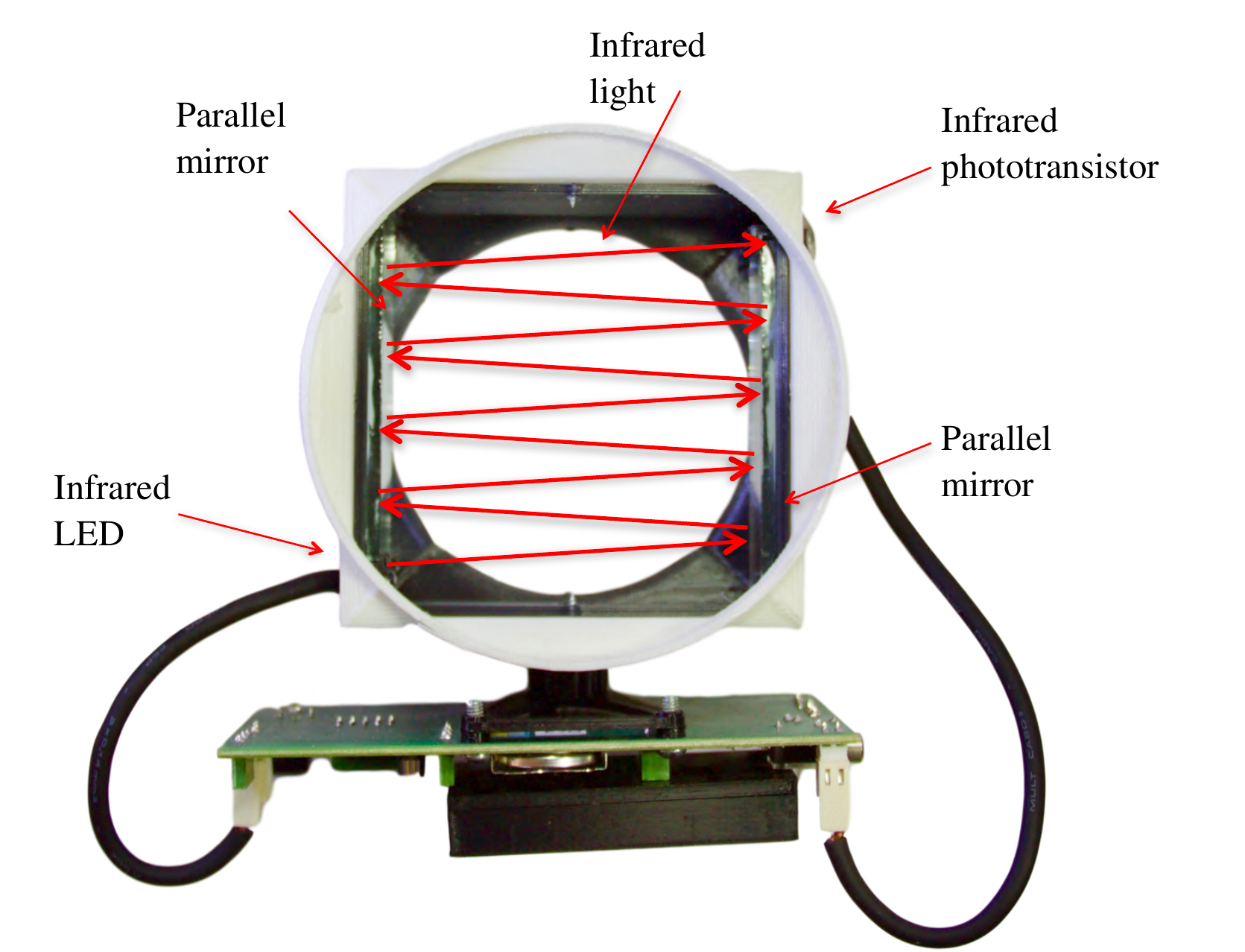}
   }
         \caption{Optical sensor to capture information about flying insects. When an insect flies across the sensor, a light variation is registered by the phototransistor as an audio signal.}
   \label{fig:sensor}
\end{figure}

The proposed data in this paper were obtained from an optical sensor built with low-cost components to capture information about flying insects remotely. This sensor is the core of the electronic smart trap presented in Section~\ref{subsec:smart_trap}. Fig.~\ref{fig:sensor} shows the design of the sensor.

The sensor has two parallel mirrors face-to-face. An infrared LED uses the mirrors to create an infrared light window that is captured by a phototransistor. The infrared light bounces back and forth between the mirrors until it reaches the phototransistor. When a flying insect crosses the light, its wings and body partially occlude the light, causing small variations that are captured by the phototransistor as an audio signal. The optical device is essentially deaf to any agent that does not cross the light. This is an important feature compared to regular microphones which are susceptible to noise.

\begin{figure}[htb]
\centering
   \subfigure[Signal]{
     \includegraphics[scale=0.3]{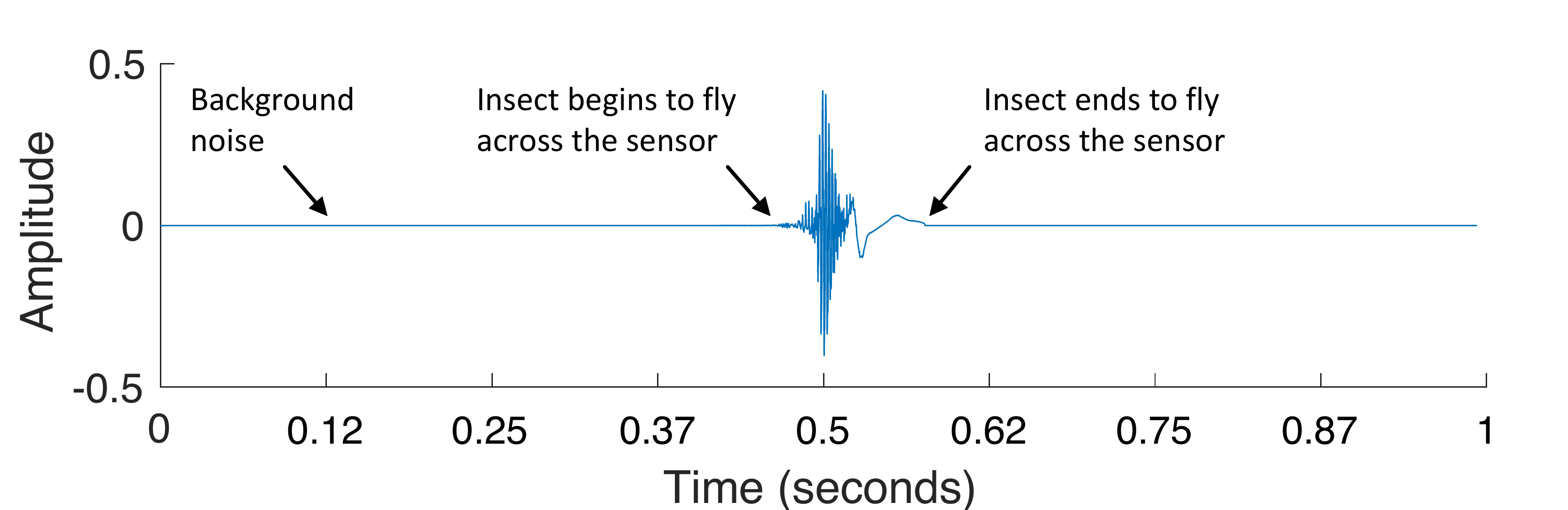}
   }
   \subfigure[Spectrum]{      
    \includegraphics[scale=0.3]{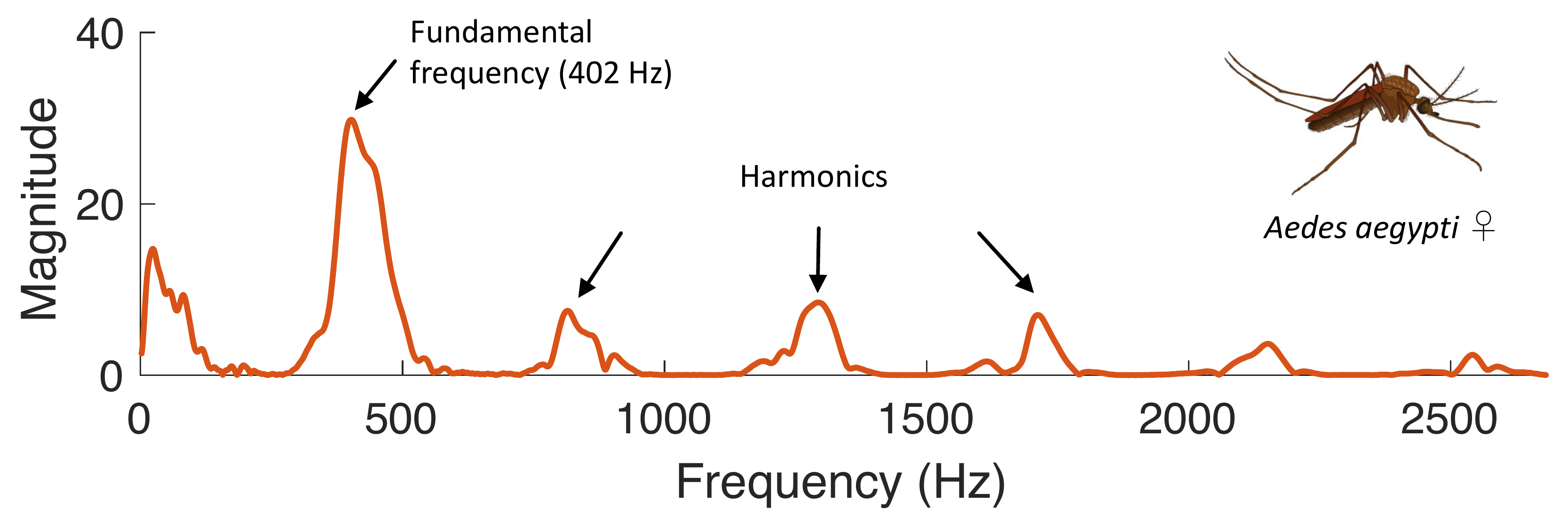}
   }
   
         \caption{A signal generated by the optical sensor given the crossing of an \emph{Aedes aegypti} (female) through the light and the spectrum of frequencies of the same signal. From the spectrum of frequencies, we can see the wing-beat frequency of the insect (402 Hz) according to the fundamental frequency. Also, the location of harmonics in the spectrum is a piece of important information for species discrimination.}
   \label{fig:signal_spectrum}
\end{figure}

Fig.~\ref{fig:signal_spectrum}-(a) shows an example of data collected by the sensor given a mosquito crossing. That signal was collected from a female \textit{Aedes aegypti} mosquito, a vector of diseases such as dengue, chikungunya, yellow, and Zika fever. The data consist of an audio fragment that usually lasts for a few tenths of a second. 

To classify the insect species, the wing-beat frequency is one of the most relevant pieces of information that can be extracted from the signals. When the signal is represented in the frequency domain, certain properties such as the fundamental frequency are made explicit. In the case of insect data, the fundamental frequency is directly related to the wing-beat frequency. Beyond the fundamental frequency, the spectrum of a signal also has harmonic components with (typically) smaller magnitudes multiples of the fundamental frequency. The position and amplitude of these harmonics also constitute important information to distinguish different insect species. Fig.~\ref{fig:signal_spectrum}-(b) shows both wing-beat frequency and harmonics, given the same signal generated by a female \textit{Aedes aegypti} mosquito.

\subsection{Data Collection in Changing Environment}\label{subsec:data_collection}

To build the insect stream datasets with concept drifts, we collect data from different species using our optical sensor in a  non-stationary environment for three months approximately. We collected data in S\~{a}o Carlos, S\~{a}o Paulo, Brazil (University of S\~{a}o Paulo campus).

To know the true class label of each insect passage during data collection, we build different collector devices in which only one insect species (with many specimens) is present inside the collector. Temperature, humidity, luminosity, and air pressure sensors are positioned in the internal part of the collector.

\begin{figure}[htb]
\centering
   \subfigure{
     \includegraphics[trim={0 0 0.4cm 0.44cm},clip, scale=0.47]{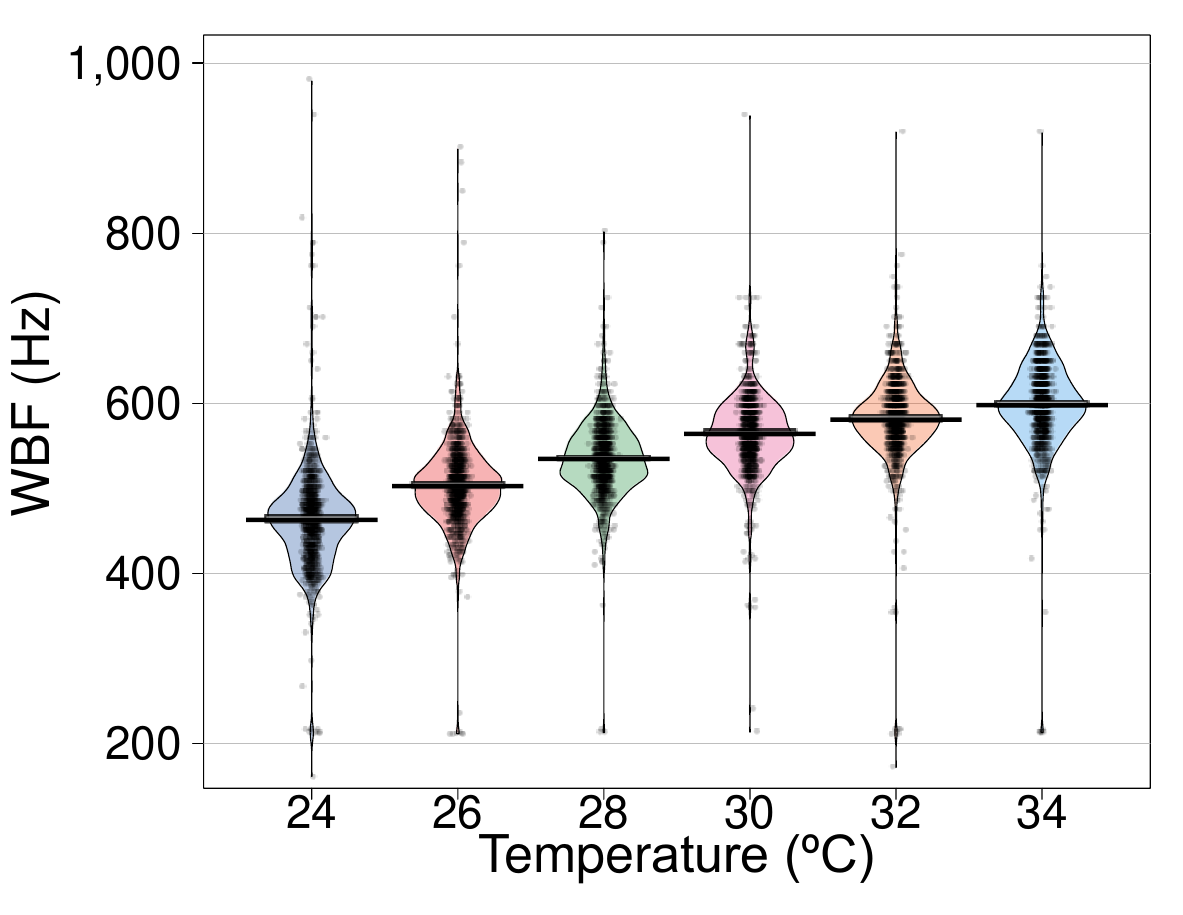}
   }
   \hspace{0cm}
   \subfigure{
     \includegraphics[trim={0 0 0.4cm 0.45cm},clip, scale=0.47]{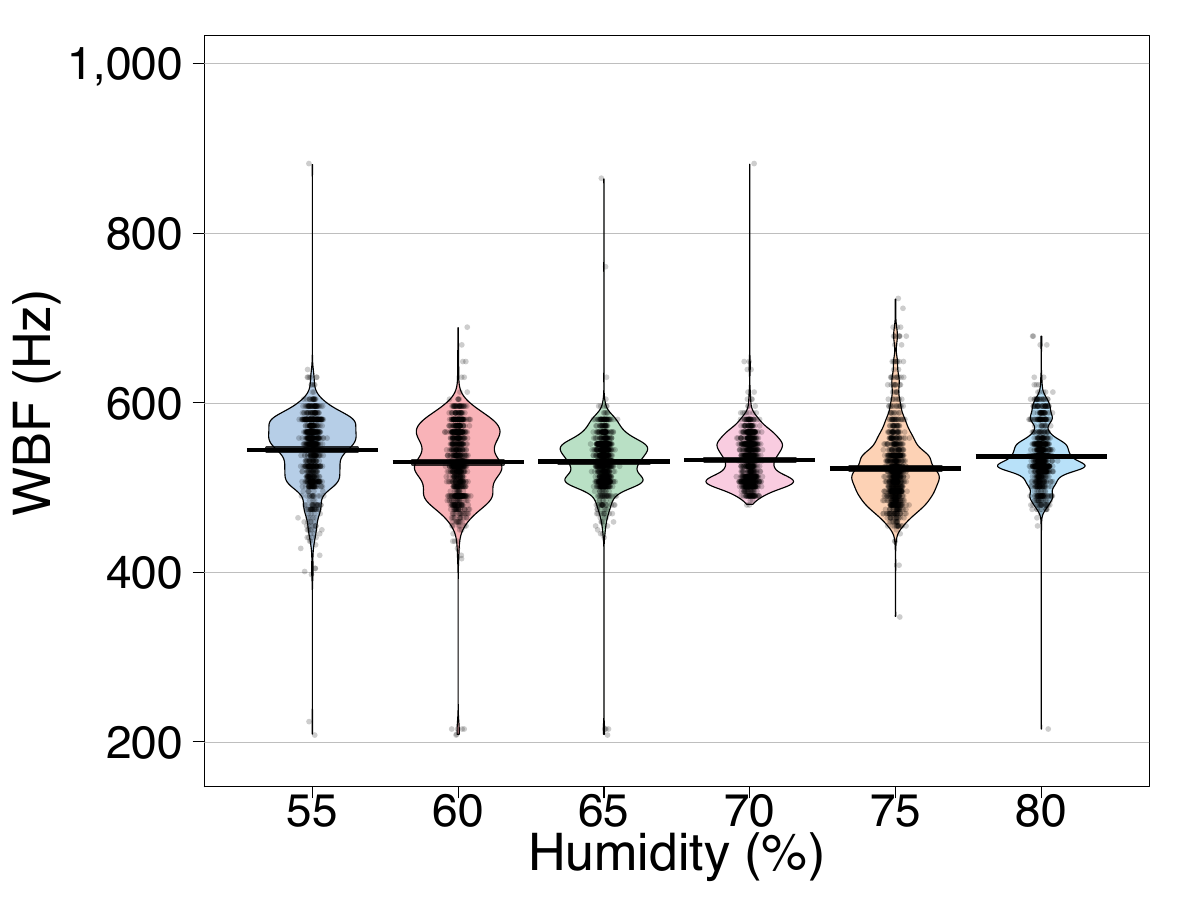}
   }
         \caption{Influence of temperature and humidity on the wing-beat frequency observed for \emph{Aedes aegypti} (female) mosquitoes.}
   \label{fig:temperature_plots}
\end{figure}

The temperature has a direct influence on the measured data by the sensor with impact in the wing-beat frequency \citep{Taylor:63, villarreal2017impact, gebru2018multiband}. However, we do not find clear evidence that humidity has any significant effect. For example, in Fig.~\ref{fig:temperature_plots} we show the WBF versus temperature and humidity for female \emph{Aedes aegypti} mosquitoes. This plot is similar to a typical box plot, but it also shows the kernel probability density of the data at different values. To collect data for this plot, we varied temperature from 24$^{\circ}$C to 34$^{\circ}$C (increments of 2$^{\circ}$C) while keeping relative humidity constant at 70\% and varied humidity from 55\% to 80\% (increments of 5\%) while keeping the temperature constant at 28$^{\circ}$C.

\begin{figure}[htb]
     \centering
    \includegraphics[trim={0.5cm 0 0cm 0},clip,scale=0.3]{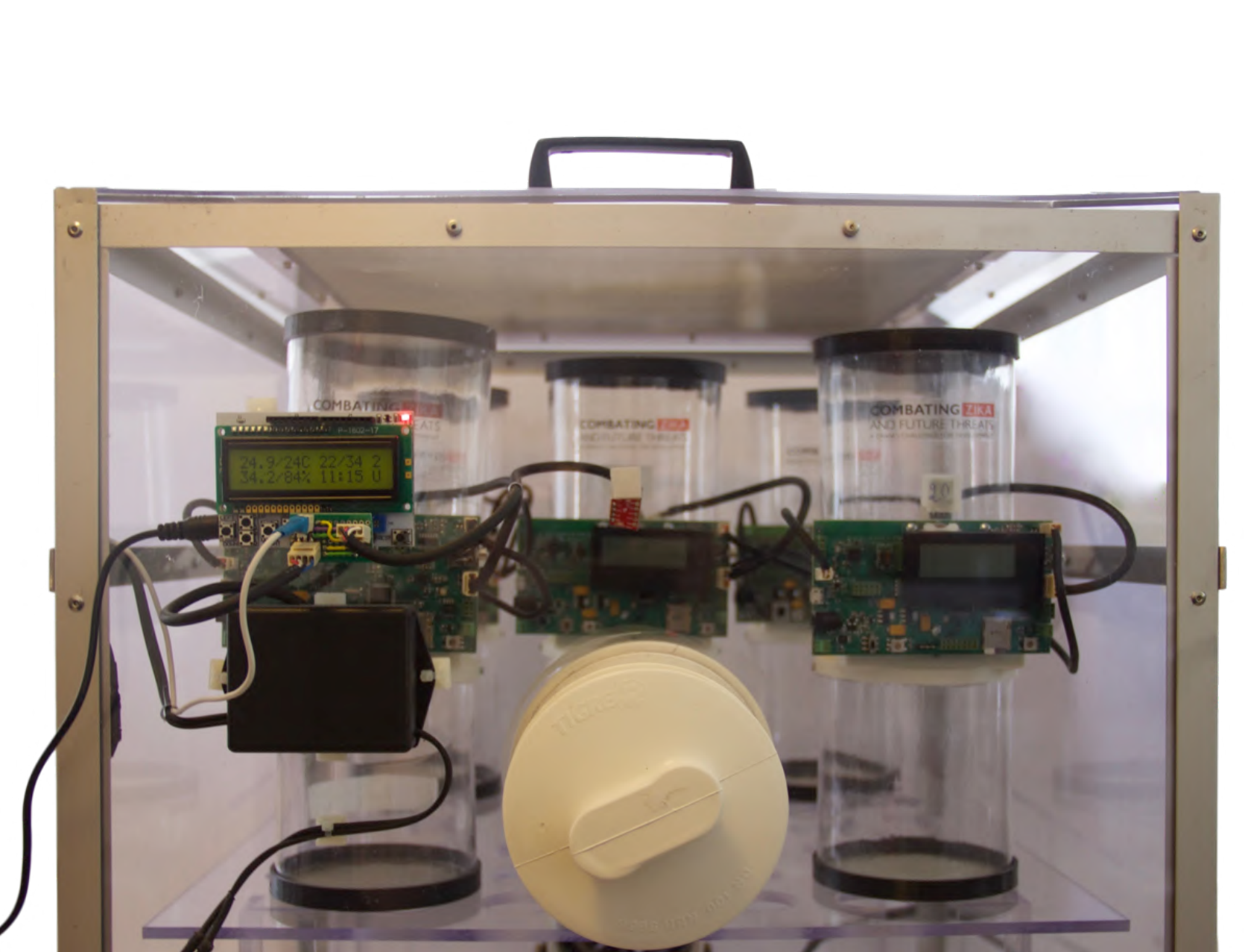}
     \caption{Chamber used to control temperature and humidity conditions in data collection.}
     \label{fig:chamber}
 \end{figure}

To collect data that contemplate a wide range of environmental variation, but in a controlled manner, we built chambers where we can control temperature and humidity manually using a custom circuitry. We put the collectors inside the chambers to gather data of different species in parallel with the same environmental condition. In Fig.~\ref{fig:chamber}, we show a chamber with five data collectors inside. 
 
\begin{figure}[htb]
    \centering
    \includegraphics[scale=0.4]{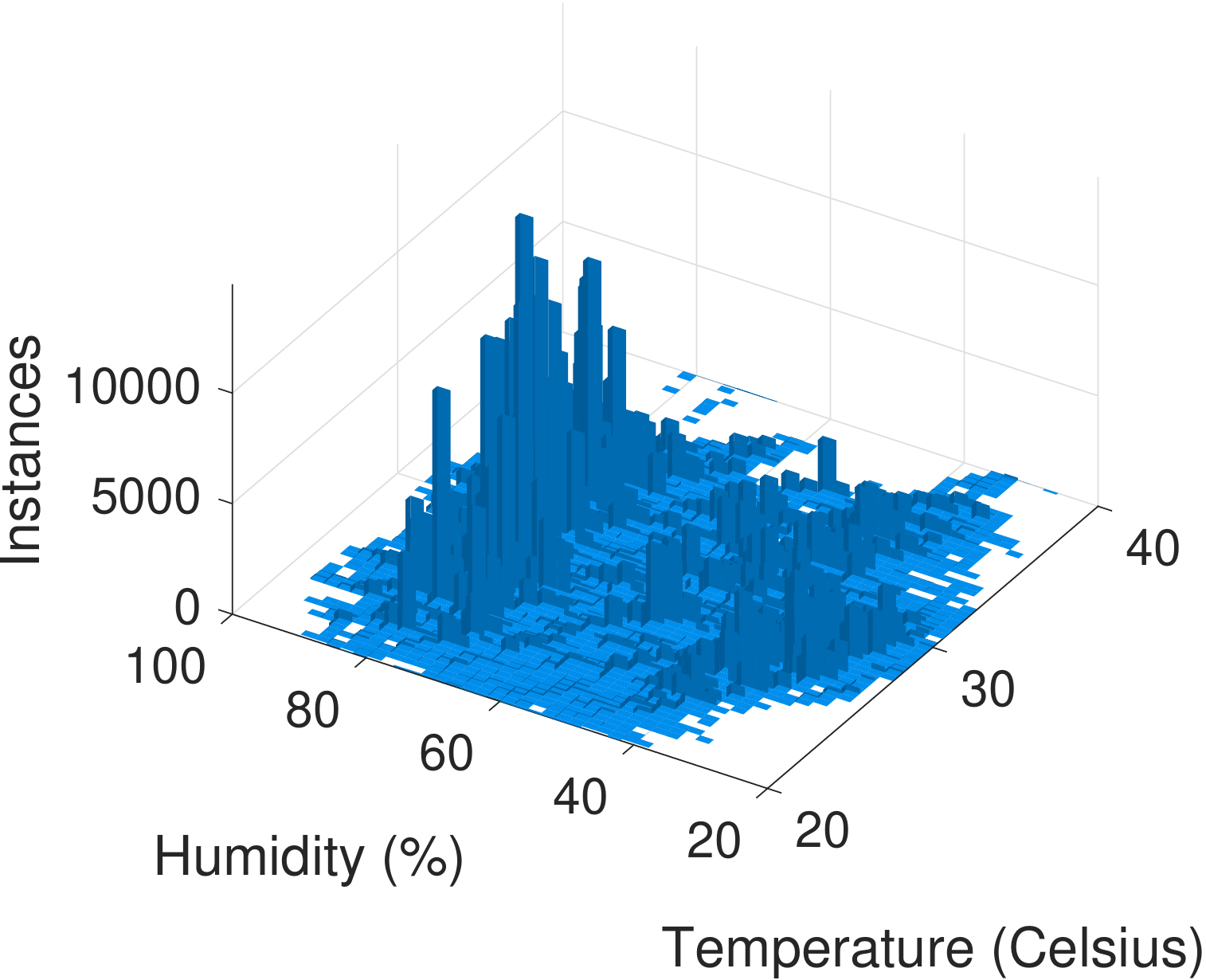}
    \caption{Number of instances observed given different values of temperature and humidity in the data collection for all species.}
    \label{fig:tempXhum}
\end{figure}

We collected around one million instances for 17 different insect species, including mosquitoes, houseflies, bees, and wasps. For 7 of the 17 insect species, it was possible to collect the data separated by sex, totaling 21 class labels. For approximately three months, we varied the temperature from 20$^{\circ}$C to 40$^{\circ}$C and the humidity from 20\% to 90\%, considering different combinations of both variables. In Fig.~\ref{fig:tempXhum}, we show the distribution of the instances from different species over both variables. In this plot, each blue bar represents the number of insect passages given a value for humidity and temperature. As we can see, our data collection has contemplated a wide range of combinations, with more instances when the humidity is around 80\%. We note that the proportions of observations made for different combinations of humidity and temperature do not necessarily translate to how proportionally active the insects are concerning such variables in nature. 

As some species are less active during certain times of the day or present a reduced lifetime, it was not possible to collect observations for all of them covering the entire range of variation in temperature and humidity. For these reasons, we built our datasets considering a subset with three species from both sexes, generating six class labels. We choose the following most active species:

\begin{itemize}
    \item \emph{Aedes aegypti.} Also known as the yellow fever mosquito, is a mosquito that can spread dengue fever, chikungunya, Zika fever, Mayaro and yellow fever viruses, and other disease agents. This mosquito originated in Africa~\citep{mousson2005phylogeography}, but is now found in tropical, subtropical and temperate regions throughout the world~\citep{eisen2013aedes};
    
    \item \emph{Aedes albopictus.} Also known as Asian tiger mosquito or forest mosquito, is a  species that can be currently found in temperate and tropical Asia (its area of origin), Europe, North and South America, Africa and several locations in the Pacific and Indian Oceans~\citep{paupy2009aedes}. It is an epidemiologically important vector for the transmission of many viral pathogens, including yellow fever, dengue fever, and Chikungunya fever, as well as several filarial nematodes such as Dirofilaria immitis~\citep{gratz2004critical};
    
    \item \emph{Culex quinquefasciatus.} Commonly known as the southern house mosquito, is a medium-sized mosquito found in tropical and subtropical regions of the world. It is the vector of Wuchereria bancrofti, avian malaria, and arboviruses including St. Louis encephalitis virus, Western equine encephalitis virus, and West Nile virus~\citep{bartholomay2010pathogenomics}.
    
\end{itemize}

\begin{figure}[htb]
    \centering
    \subfigure[\textit{Aedes aegypti}]{
    \includegraphics[scale=0.3]{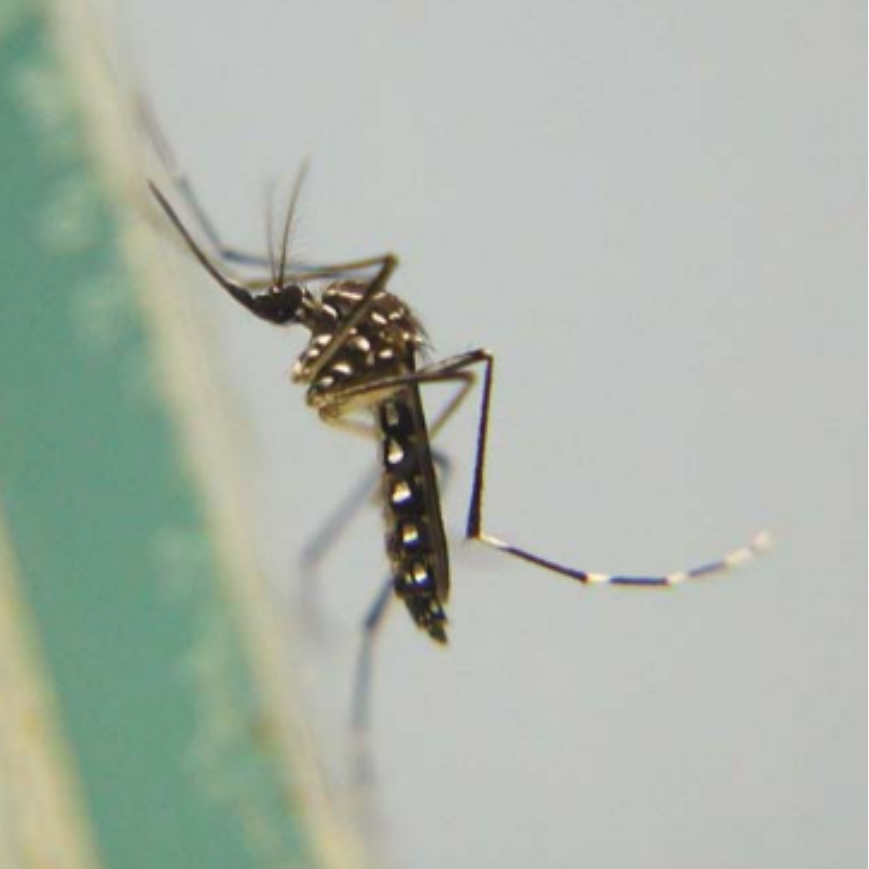}
    }
    \subfigure[\textit{Aedes albopictus}]{
    \includegraphics[scale=0.3]{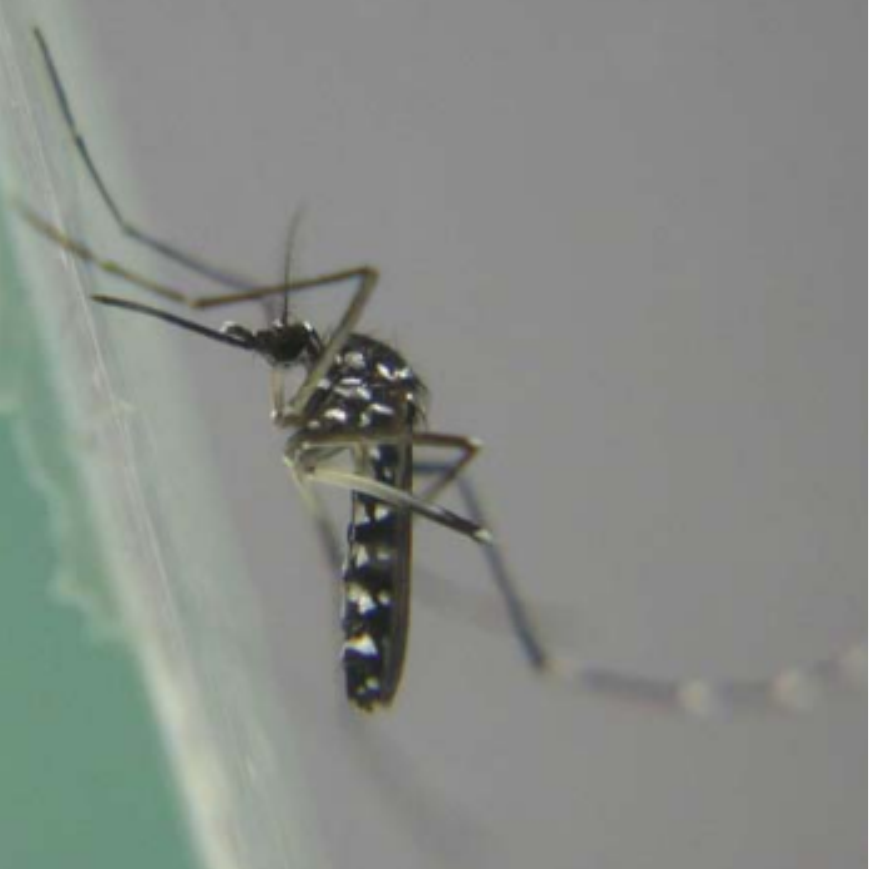}
    }
    \subfigure[\textit{Culex quinquefasciatues}]{
    \includegraphics[scale=0.3]{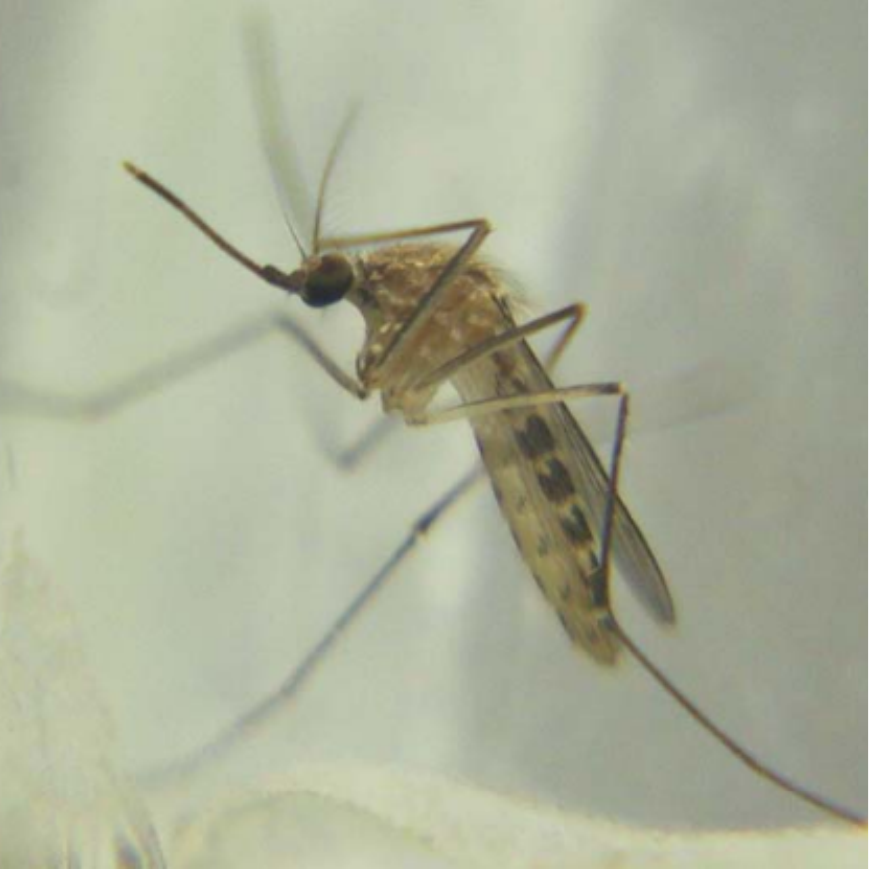}
    }
    \caption{Adult male mosquitoes from the species present in our datasets. All photographs were taken by Michele M. Cutwa~\citep{cutwa2006photographic}.}
    \label{fig:species_photo}
\end{figure}

The anatomy of the three species is very similar, especially when we consider species from the same genus (\textit{Aedes aegypti} and \textit{Aedes albopictus}). This similarity is also observed in the flight of the insects and consequently in the measured data by the sensor. Fig.~\ref{fig:species_photo} illustrates with photos the species present in our datasets.

\subsection{Features Extraction}\label{subsec:features_extraction}
\begin{figure}[htb]
    \centering
    \includegraphics[scale=0.4]{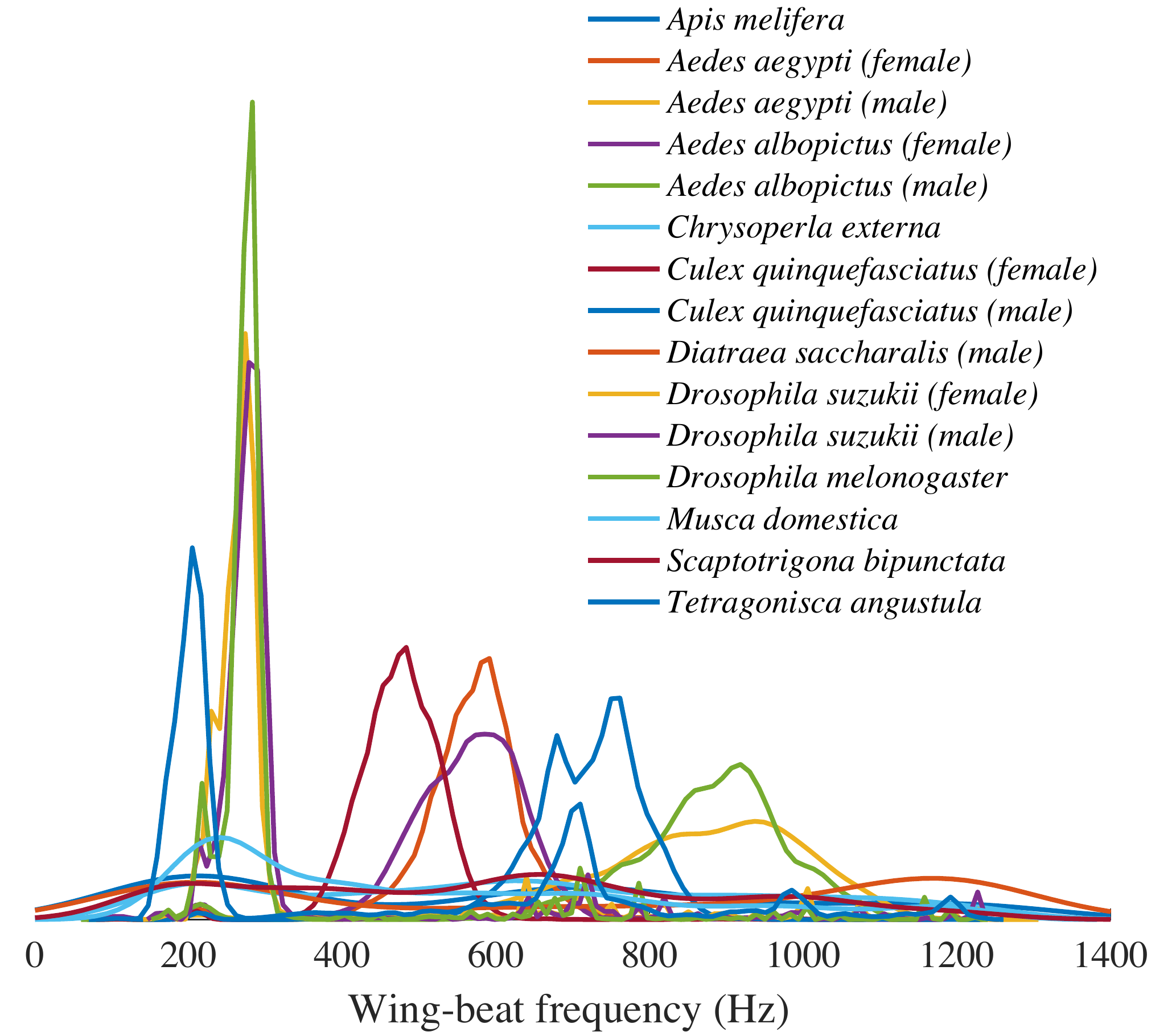}
    \caption{Density functions that fit the histograms of different insect species.}
    \label{fig:species_distribution}
\end{figure}

As we consider a more substantial number of species, it is clear from the pigeonhole principle~\citep{ajtai1988complexity} that to classify those species with high accuracy it is required additional features than only the wing-beat frequency (WBF). For instance, Fig.~\ref{fig:species_distribution} illustrates the distributions of the wing beat frequency for 15 species across all temperatures for which we possess data. The wing-beat frequency is one of the most distinctive attributes available for this application. In this figure, we can see that although some species show a well-defined peak in the mean values of WBF, there is a significant overlap among the species. In this sense, the use of only this feature to classify the insect species can be inaccurate.

Our signal, although optical, is very similar to audio, as previously shown in Fig.~\ref{fig:signal_spectrum}, and consequently high-dimensional. Thus, we employ signal processing techniques to extract additional discriminative features from data. 

Each audio file was pre-processed and transformed into a feature vector. We extracted a series of features such as the wing-beat frequency, complexity measures of the signal spectrum, statistics from temporal representation, among others. For the benchmarking data, we provide 33 features related to the energy sum of frequency peaks and harmonics positions.

\subsection{Insect Stream Data}\label{subsec:benchmark_description}

Given the impact of temperature on the measured data by the optical sensor leading to the occurrence of concept drifts, we built our benchmarking data based on changes in this variable. Each temperature was measured in Celsius degrees and rounded to the nearest integer value. Thereby, we ordered the observations of the examples over time in the stream following different patterns of change in temperature while hiding this variable from the dataset. We reiterate that although we have manipulated the sequence of the examples to control the drifts, all these changes are feasible in the real use of the sensor on dynamic environments. Additionally, for each temperature, we uniformly sampled examples that were collected within that temperature. As a result, we eliminate all other sources of drift beside the changes in temperature. Finally, in addition to sampling from individual temperatures, we also vary the proportion of the classes over time, to mimic natural influences in the activity of insects: circadian rhythm, the presence of predators, among others. We consider the following changes, which also name our datasets:

\begin{itemize}
    \item \textbf{Incremental.} In this pattern, the instances are arranged so that the temperature values are incrementally increased from 20$^{\circ}$C to 40$^{\circ}$C over all the stream;
    
    \item \textbf{Abrupt.} We consider five sudden change points in this pattern. The first instances of the stream were collected at a temperature of 30$^{\circ}$C, and then they abruptly change to 20$^{\circ}$C. After a time, the temperature back to change for values around 35$^{\circ}$C. Similarly, other three abrupt changes occur until the end of the stream;
    
    \item \textbf{Incremental-gradual.} In this pattern, the observed temperature in the first instances is around 37$^{\circ}$C and incrementally decrease until 35$^{\circ}$C. For a period, we have a gradual change where the temperature of the instances intercalates in the values of 35$^{\circ}$C 23$^{\circ}$C until definitively change for 23$^{\circ}$C. In this period, two different concepts are active at the same time. At the end of the stream, the temperature back to incrementally increase until 27$^{\circ}$C;
    
    \item \textbf{Incremental-abrupt-reoccurring.} This pattern provides three recurrent cycles of incremental changes where the temperature increase from 20$^{\circ}$C to 40$^{\circ}$C. Between the end and beginning of a cycle of incremental changes, we have an abrupt change;
    
    \item \textbf{Incremental-reoccurring.} In this pattern, there exist three cycles of incremental changes over time. In the first cycle, the temperature increases from 20$^{\circ}$C to 40$^{\circ}$C. In the second cycle, the temperature decreases from 40$^{\circ}$C to 20$^{\circ}$C. In the end, the temperature turns to increase to 40$^{\circ}$C. Although the stream presents two clear recurrent patterns where the values are increased, we also can consider the cycle of decreasing temperature as recurrent, but in an ``inverse'' arrival order of the instances;
    
    \item \textbf{Out-of-control.} In this case, we have a lack of pattern in the occurrence of changes in the temperature. It means that is expected the arrival over time of instances observed at any temperature. This dataset is composed of all collected data in uniformly random order. As each example is sampled uniformly sampled at each time during the stream, this dataset must be drift-free.
\end{itemize}

\begin{figure}[htb]
\centering
   \subfigure[Incremental]{
     \includegraphics[scale=0.4]{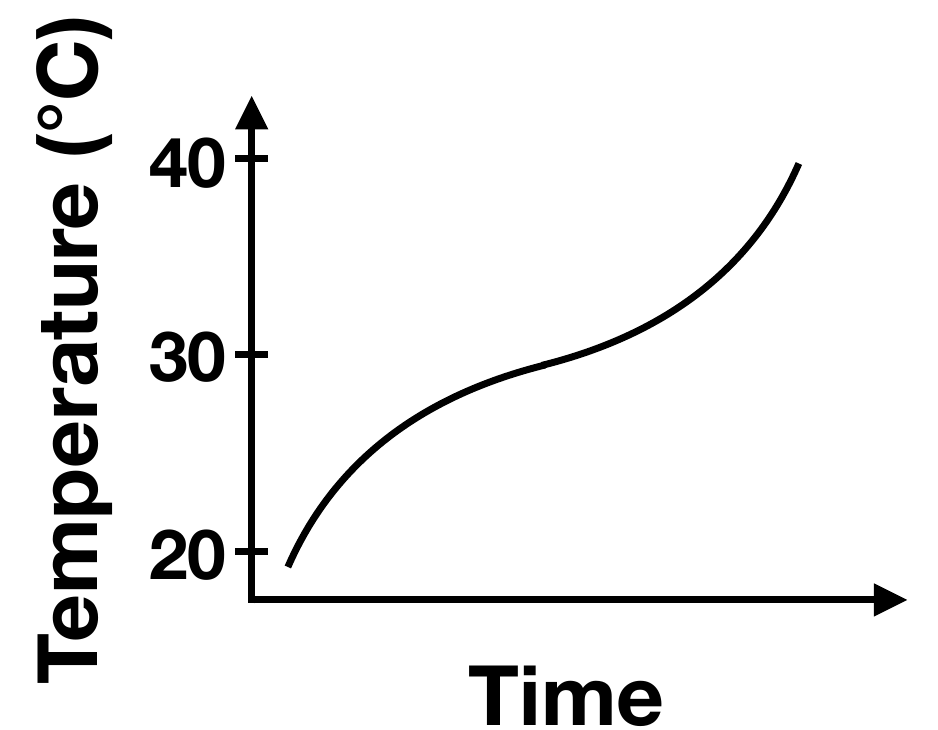}
   }
   \hspace{-0.28cm}
   \subfigure[Abrupt]{
     \includegraphics[scale=0.4]{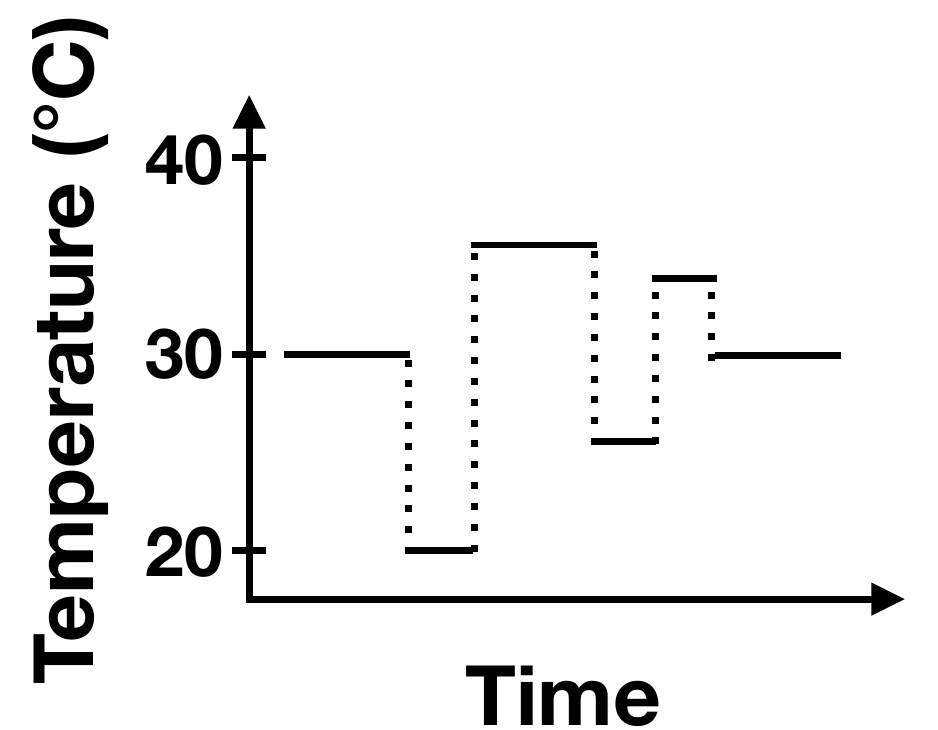}
   }
   \hspace{-0.28cm}
   \subfigure[Incremental-gradual]{
     \includegraphics[scale=0.4]{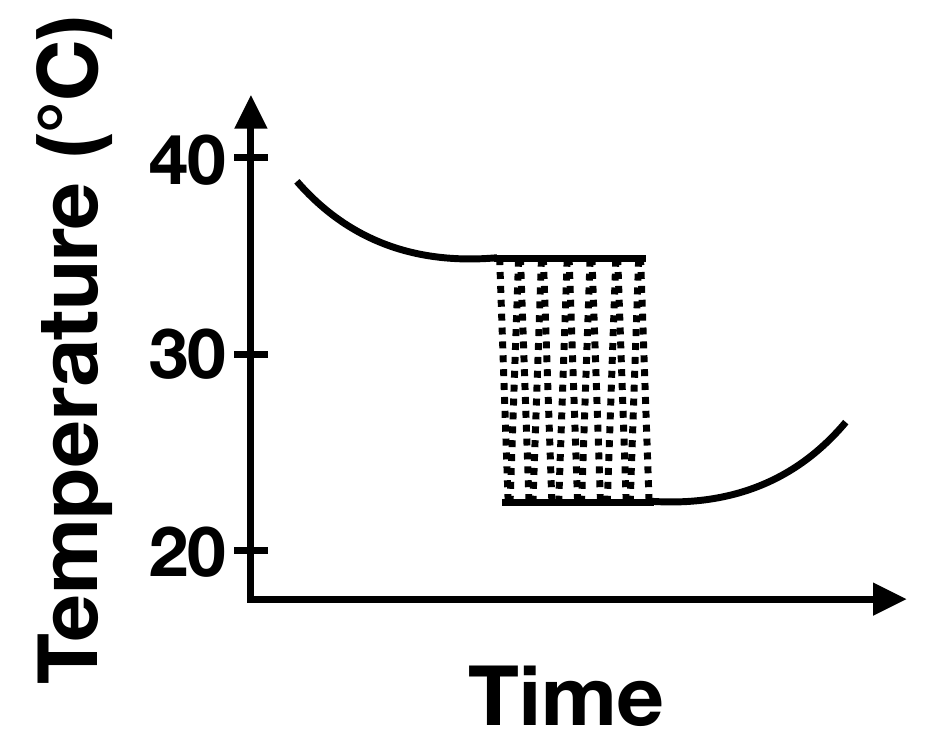}
   }   
   \hspace{-0.28cm}
   \subfigure[Incremental-abrupt-reoc.]{
     \includegraphics[scale=0.4]{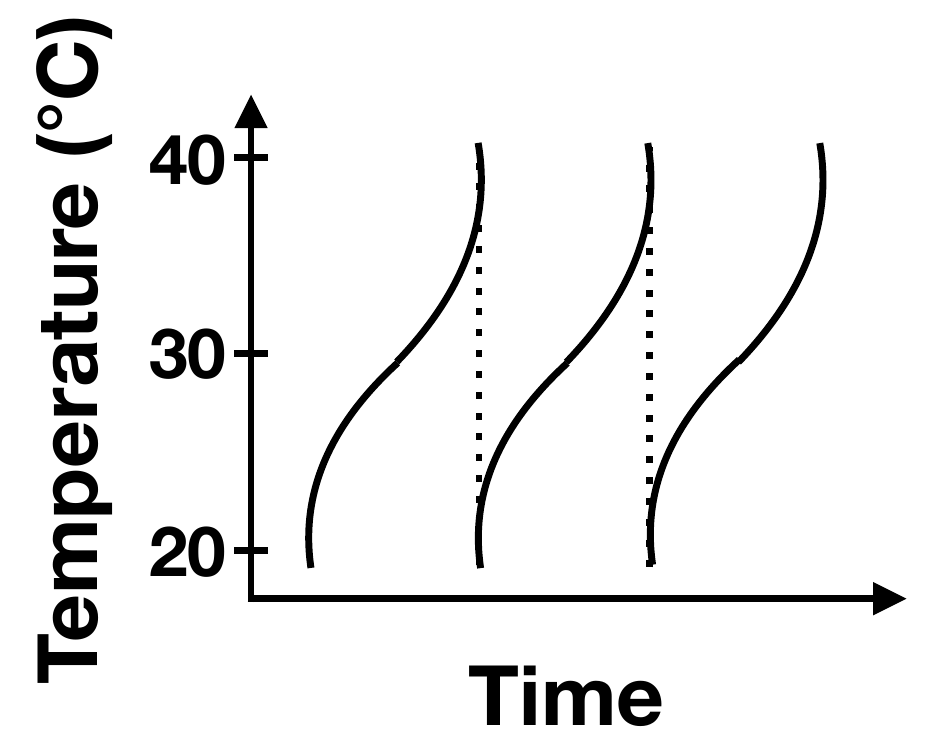}
   }  
   \hspace{-0.28cm}
   \subfigure[Incremental-reoccurring]{
     \includegraphics[scale=0.4]{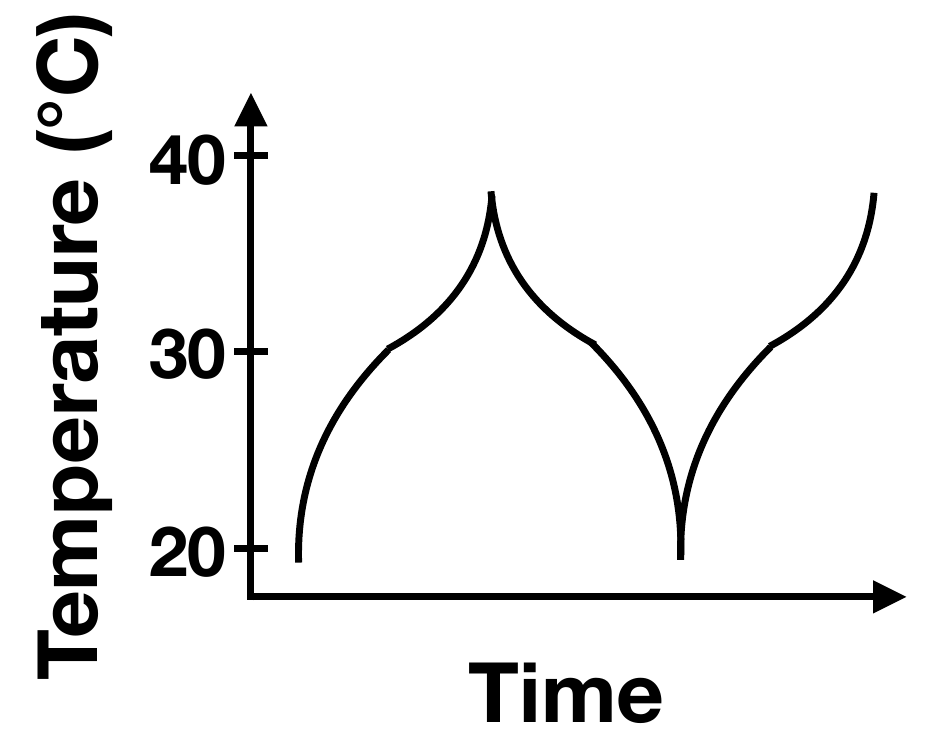}
   }
   \hspace{-0.28cm}
   \subfigure[Out-of-control]{
     \includegraphics[scale=0.4]{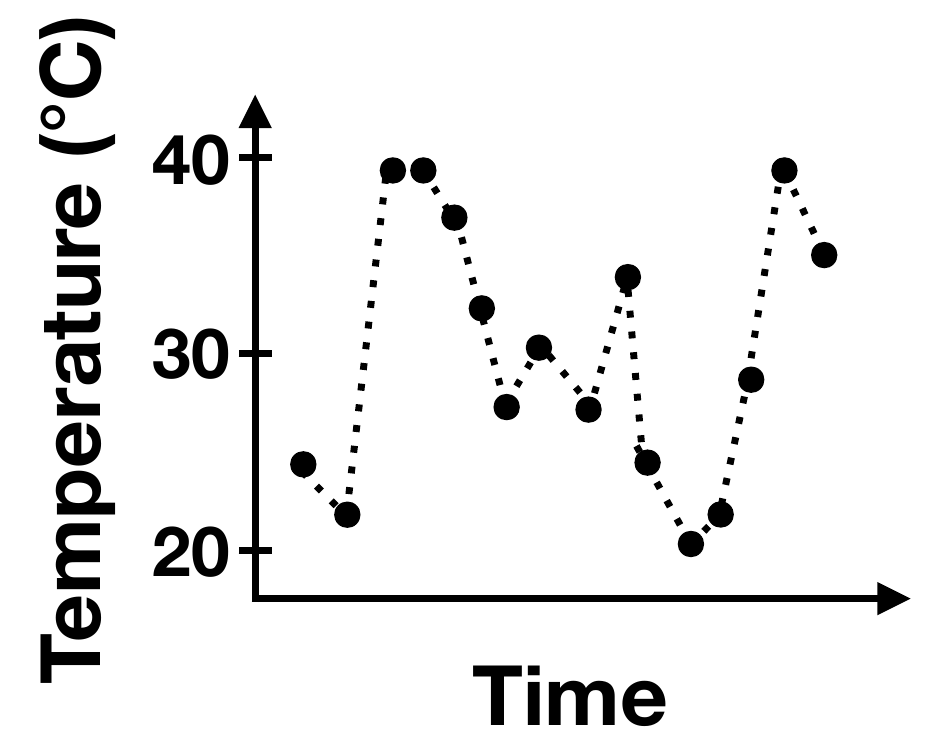}
   }     

         \caption{Patterns of changes given the variable temperature to build the Insect Stream Data.}
   \label{fig:patterns_insects}
\end{figure}

Fig.~\ref{fig:patterns_insects} graphically illustrates the patterns of changes presented in our datasets. 

For the first five patterns showed in Fig.~\ref{fig:patterns_insects}, we built two datasets for each one, being the first with balanced and the second with imbalanced class distribution. For the last dataset (Out-of-control), we have only an imbalanced version. Thus, we have a total of 11 different datasets. 

\begin{figure}[htb]
    \centering
    \includegraphics[scale=0.37]{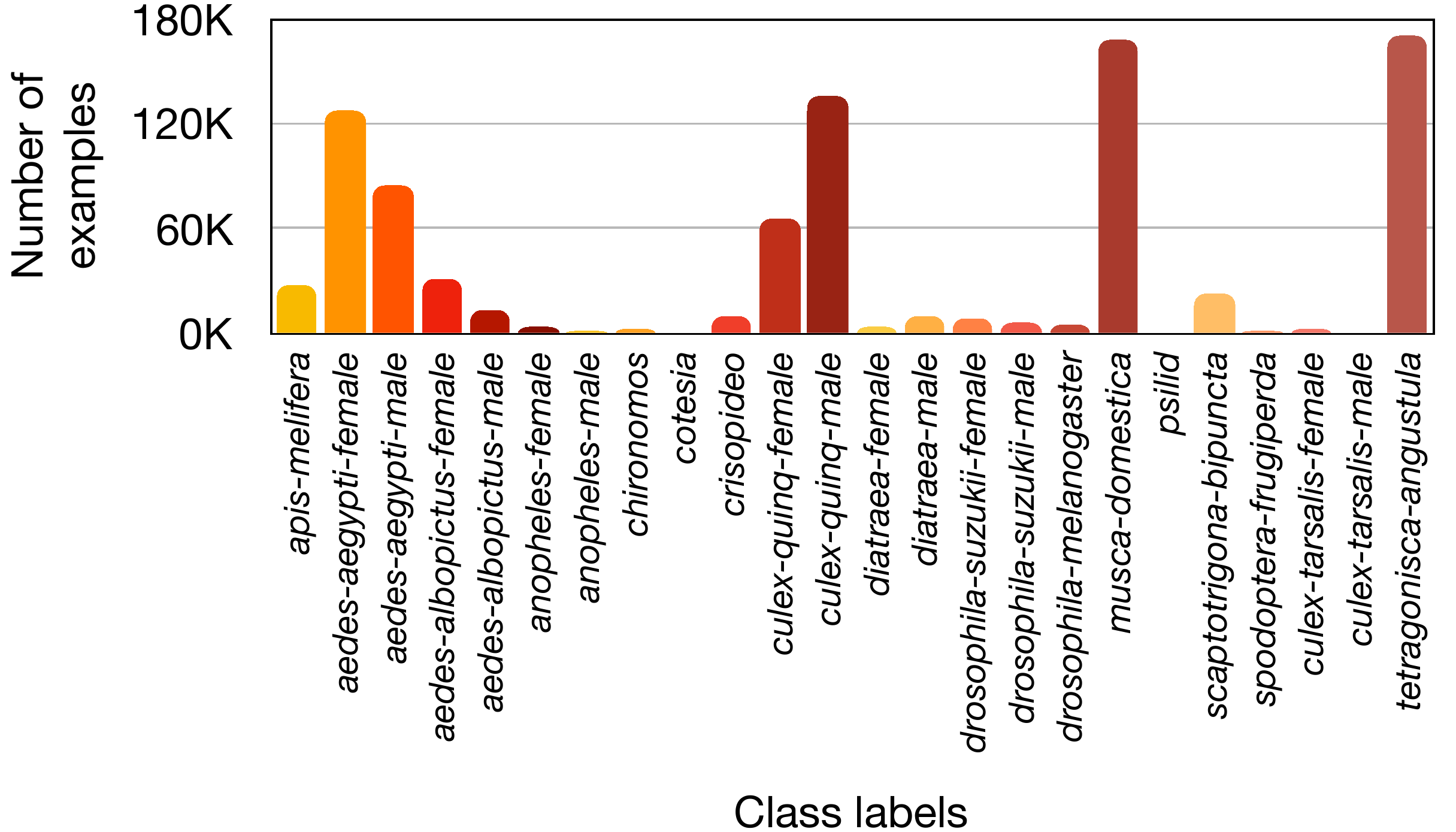}
    \caption{Class distribution of Out-of-control dataset.}
    \label{fig:out-of-control-dist}
\end{figure}

Fig.~\ref{fig:out-of-control-dist} shows the distribution over 24 class labels from the Out-of-control dataset. In this dataset, the \textit{tet-angustula} and \textit{musca} are the majority classes with 170,220 (18.81\%) and 168,819 (18.65\%) instances, respectively.  While classes such as \textit{psilid} and \textit{cx-tarsalis-male} are the minority classes with only 17 and 157 instances, respectively. Thus, this dataset has two main challenges: the lack of a pattern to distinguish the concepts and overcome the temporal overlap and imbalanced distribution. One additional note is that the proportions of the classes in the dataset are subject to data collection bias and do not represent the real proportion of the species in nature. Furthermore, we expect such proportions to vary according to time, region, and ambient conditions.

\begin{figure}[htb]
\centering
   \subfigure[Balanced]{
     \includegraphics[scale=0.31]{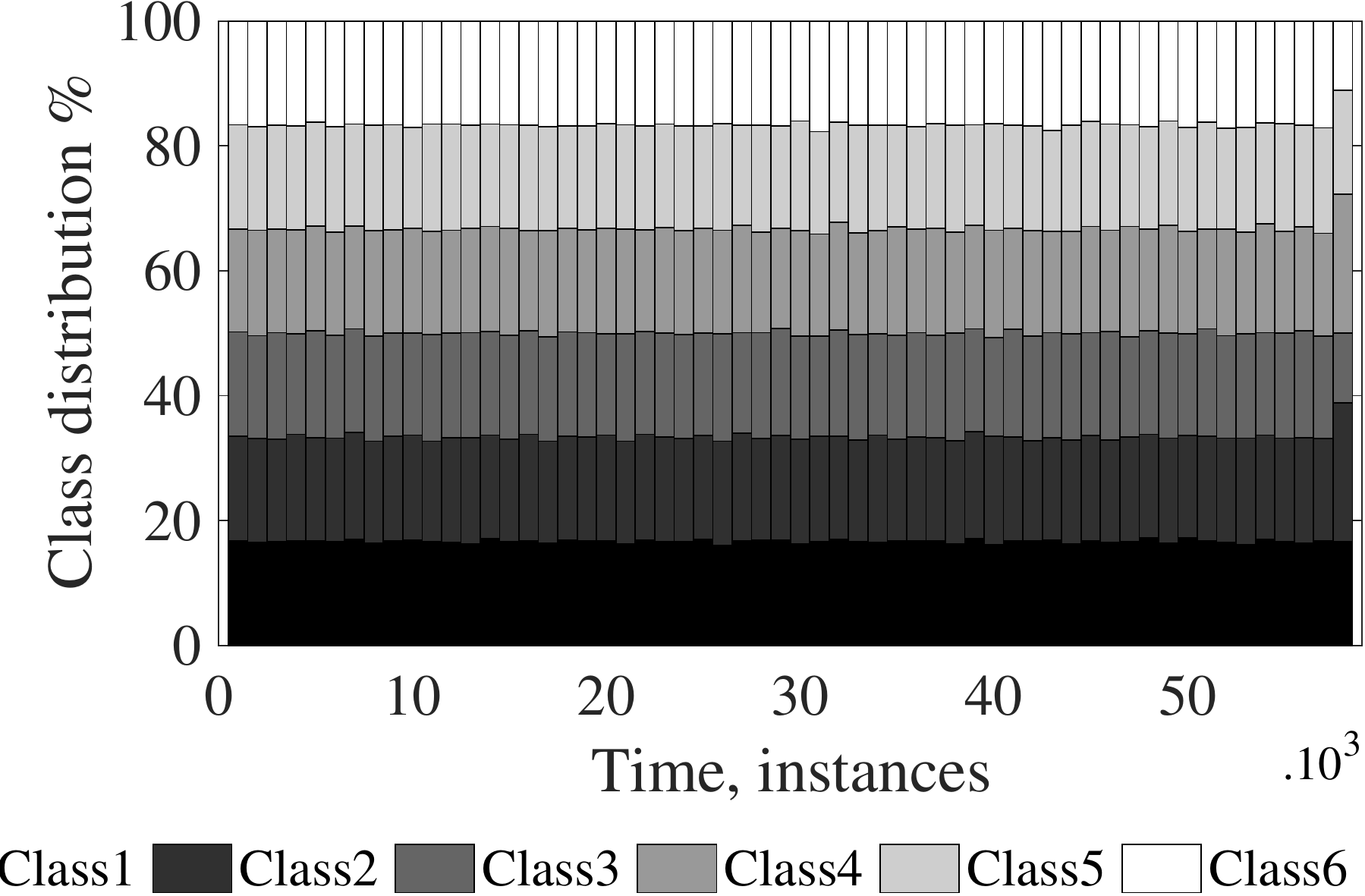}
   }
   \hspace{-0.28cm}
   \subfigure[Imbalanced]{
     \includegraphics[scale=0.31]{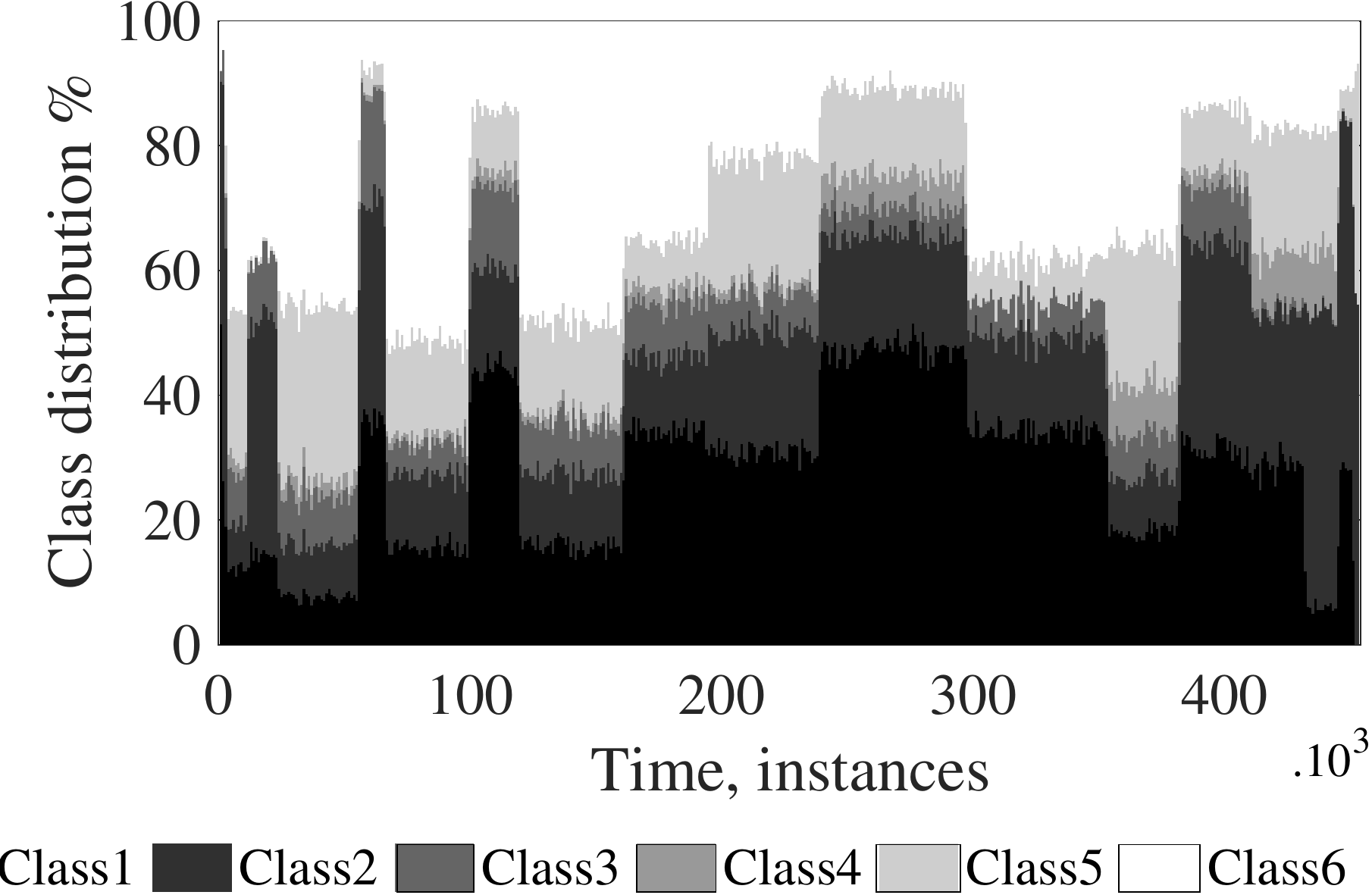}
   }
         \caption{Changes in class proportion for the Incremental dataset considering the balanced and imbalanced class data versions. Each bar in the plots represents the class proportions into a window with 1,000 consecutive examples in the stream.}
   \label{fig:bal_imbal}
\end{figure}

Fig.~\ref{fig:bal_imbal} illustrates the changes in the classes proportion over time for the two versions of the Incremental dataset. However, we note that not all balanced data versions are as well behaved as seen in Fig.~\ref{fig:bal_imbal}-(a).

All datasets have 33 features, as previously discussed in Section~\ref{subsec:data_collection}. Except for the Out-of-control dataset that has 24 class labels, all other datasets have 6 class labels related to the species \textit{Aedes aegypti} (female and male), \textit{Aedes albopictus} (female and male), and \textit{Culex quinquefasciatus} (female and male). The 24 class labels from the Out-of-control dataset can be seen in Fig.~\ref{fig:out-of-control-dist}. Besides the higher number of class labels of this data, another interesting characteristic is the emergence of new classes over time, which allows their use in the evaluation of approaches for novelty detection~\citep{masud2009integrating}. As is also often the disappearance of certain classes over time, this dataset can be useful for assessing solutions dealing with significant changes in $P(Y)$.

In Table~\ref{tab:benchmark_description}, we show a description of the datasets as the number of instances and the position of the change points. Their names can identify the patterns of changes for each dataset.

\begin{table}[htb]
    \centering
    \scriptsize
    \renewcommand\tabcolsep{2pt}   
    \caption{Description of the Insect Stream Datasets.}
    \label{tab:benchmark_description}
    \begin{tabular}{lrcc}
         \textbf{Dataset} & \textbf{Instances}   & \textbf{Change point(s)}  \\ \hline
         Incremental (bal.) & 57,018  & Throughout all the stream \\ 
         Incremental (imbal.) & 452,044 &  Throughout all the stream \\ 
         Abrupt (bal.) & 52,848 & 14352; 19500; 33240; 38682; 39510 \\ 
         Abrupt (imbal.) & 355,275 & 83859; 128651; 182320; 242883; 268380\\ 
         Incremental-gradual (bal.) & 24,150 &  14028\\ 
         Incremental-gradual (imbal.) & 143,323 & 58159 \\ 
         Incremental-abrupt-reoccurring (bal.) & 79,986 & 26568; 53364\\ 
         Incremental-abrupt-reoccurring (imbal.) & 452,044 & 150683; 301365 \\ 
         Incremental-reoccurring (bal.) & 79,986 & 26568; 53364 \\ 
         Incremental-reoccurring (imbal.) & 452,044 & 150683; 301365\\ 
         Out-of-control & 905,145 & Throughout all the stream \\ 
         \hline
    \end{tabular}

\end{table}

\subsection{Temporal Overlap}\label{subsec:insects:temporal:overlap}

Interesting datasets for streaming problems include different aspects discussed in this article: changes in the proportions of the classes over time, changes in the distribution of the features within each class over time, and temporal overlap, \textit{i.e.}, the dynamism of the overlap depending on which concept is responsible for the current examples.

We showed in the previous sections that the data we are providing have changes in the distribution of the features as we vary the temperature, and also a great deal of overlap between classes for at least the wing-beat frequency attribute. We consider our concept to relate to the temperature directly, and, for all but one version of the dataset, the temperature is the hidden variable that evolves while being stable within windows in the stream. For that reason, temperature overlap coincides with temporal overlap, and the former is the source of the latter.

A relevant question is whether there is a smaller overlap between the classes when we consider data for each temperature value than the data with all temperatures together. If that is not the case, we may need not worry about forgetting mechanisms to discard old data, since new concepts are likely to occupy empty regions in the feature space, as the temperature varies and we aggregate more data over time. However, if class overlap varies, a classification system can potentially benefit from identifying boundaries between different concepts and using models specifically trained for each one of them.

To illustrate this idea, consider a subset of the data that contains only female \textit{Aedes aegypti} and female \textit{Culex quinquefasciatus}. Each temperature was measured in Celsius degrees and rounded to the nearest integer. We sampled $2,500$ examples from each species for each one of the following temperatures (in Celsius): 24, 26, 28, 30, 32, and 34.

\begin{figure}[htbp]
    \centering
    \includegraphics[scale=0.35]{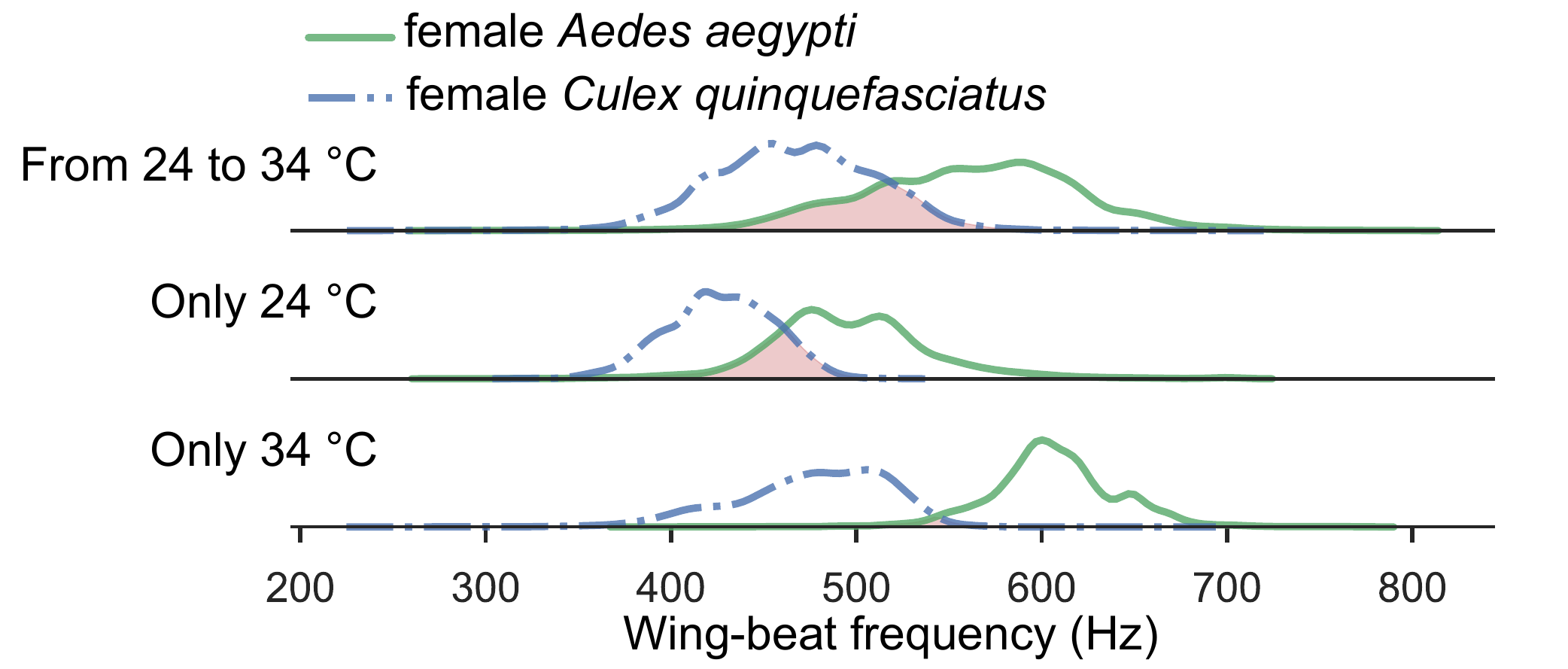}
     \caption[Case of temporal overlap]{Illustration of a case of temporal overlap. When we can discriminate the data according to the current temperature, we have smaller class overlap in the wing-beat frequency.}
    \label{fig:temporal:overlap:insects}
\end{figure}

In Fig.~\ref{fig:temporal:overlap:insects}, we visually illustrate the difference in class overlap for wing-beat frequency. This illustration is complemented by Table~\ref{tab:overlap:wbf}, which presents the numerical overlap between the two classes for each temperature. The overlap when all temperatures are considered together is 36\%, while the average overlap when each temperature is isolated is 23\%. The overlaps were estimated by taking the minimum between histograms with 100 bins.

\begin{table}[htbp]
\scriptsize
\centering
\caption{Values for a case of temporal overlap. When we can discriminate the data according to the current temperature, we have smaller class overlap in the wing-beat frequency.}
\label{tab:overlap:wbf}
\begin{tabular}{rcccccc}
\hline
Temperature ({\degree}C) & 24 & 26 & 28 & 30 & 32 & 34 \\
Overlap (\%) & 29 & 32 & 28 & 23 & 19 & 5 \\
\hline
\end{tabular}
\end{table}

Finally, to not limit ourselves to only one feature (wing-beat frequency), we indirectly measured the effect of the difference of overlaps by evaluating a classification task. We compared the use of individual classifiers for each temperature against a single classifier trained with data from all temperatures. The accuracy rates were obtained via 10-fold cross-validation and a Random Forest classifier with 200 trees. We used all 33 features from the insect dataset. Table~\ref{tab:overlap:acc:wbf} presents the accuracy rates obtained. The single classifier achieves 84\% accuracy for the whole data, while individual classifiers average 90\%. We note that greater differences can be observed depending on which temperature is individually assessed: some temperatures apparently suffer a greater deal with temporal overlap than other ones. One example is 24\degree C. It is the most difficult case even with an individual classifier, and is also the most harmed by the use of a conjoint classifier, with a 20\% difference in their accuracy.

\begin{table}[htbp]
\scriptsize
\centering
\caption{Indirect effect of temporal overlap. When we can discriminate the data according to the current temperature, we have higher accuracy for the insect data.}
\label{tab:overlap:acc:wbf}
\begin{tabular}{rcccccc}
Temperature ({\degree}C) & 24 & 26 & 28 & 30 & 32 & 34 \\
\hline
Individual classifiers & 86 & 87 & 88 & 89 & 92 & 98 \\
Single classifier      & 66 & 81 & 88 & 87 & 89 & 93 \\
\hline
\end{tabular}
\end{table}

\section{USP Data Stream Repository}\label{sec:repository}

Aiming to mitigate possible flaws in the experimental evaluation of future proposals on stream learning due to the lack of real-world data, we provide to the machine learning community a new public repository called USP Data Stream  Repository\footnote{Available online at \url{https://sites.google.com/view/uspdsrepository}}. In this repository, we make available 27 datasets from different real problems composed by 16 data previously evaluated by other works from literature and 11 new datasets obtained by the optical sensor for automatic insect recognition\footnote{The datasets are encrypted under the following password: DMKD2018}. It is important to note that we want to feed this repository regularly with new data from collaborative contributions.

We suggest that stream classifiers and drift detection algorithms should be tested on a wide range of datasets, mainly the real ones to avoid biased conclusions. It is a usual practice in more consolidated areas, such as machine learning in general  \citep{Dua:2017} and time-series \citep{UCRArchive}, which contributes to the research advancement and maturity of the data stream area. At the same time, comparisons against baseline methods such as those proposed by \citet{bifet2013pitfalls} are also essential for the better performance analysis of new proposals. In this direction, we include in the repository the results achieved by two simple baselines methods for all datasets.

\section{Evaluation and Discussion}\label{sec:evaluation}

Besides to provide benchmark datasets to evaluate classifiers and drift detection methods, we also report the results achieved by state-of-the-art methods in our proposed data. The availability of benchmark data accompanied by the results achieved by methods from literature aims to make experiments from different researchers from the data stream community easily comparable and reproducible.

We run all experiments of stream classification and drift detection using the MOA framework software~\citep{bifet2010moa}, which contains implementations of several state-of-the-art methods. 

\subsection{Classification}

In the experimental evaluation of the classification task, we consider two naive baseline classifiers~\citep{bifet2013pitfalls}: $i)$ No-Change and $ii)$ Majority-Class. Both approaches do not use any input attributes and classify only using past label information. The No-Change classifier ever predicts the next class label as the same as last seen class label. The Majority-Class made their prediction based on the majority class of a moving window over the stream with 1,000 instances. 

In addition to the baseline classifiers, we also evaluate the following stream algorithms: $i)$ incremental Naive Bayes (NB), $ii)$ Very Fast Decision Trees~\citep{hulten2001mining} with Naive Bayes classifiers at the leaves, $iii)$ Leveraging Bagging with 10 VFDT in the ensemble~\citep{bifet2010leveraging}, and $iv)$ Adaptive Random Forest~\citep{gomes2017adaptive}. We based our choices on the efficiency and popularity of the methods available for evaluation.

We consider prequential evaluation~\citep{gama2013evaluating} over a sliding window of 1,000 instances to evaluate the classification performance of the algorithms. In Table~\ref{tab:classification_results}, we show the results achieved by the methods.

\begin{table}[htb]
    \centering
    \scriptsize    
    \renewcommand\tabcolsep{2pt}   
    \caption{Prequential accuracy achieved by state-of-the-art methods in the Insect Stream Data.}
    \label{tab:classification_results}
    \begin{tabular}{lcccccc}
         \textbf{Dataset} & \textbf{No-Change}   & \textbf{Maj.Class} & \textbf{NB} & \textbf{VFDT} & \textbf{Lev.Bag.} & \textbf{ARF}  \\ \hline
         Inc (bal.)              & 16.04 & 11.51 & 47.37 & 45.65 & 61.42 & 64.29 \\
         Inc (imbal.)            & 28.23 & 29.76 & 49.30 & 44.92 & 75.13 & 78.94 \\
         Abrupt (bal.)           & 28.98 & 16.07 & 50.77 & 49.85 & 68.39 & 74.34 \\
         Abrupt (imbal.)         & 29.15 & 28.49 & 52.18 & 48.46 & 72.28 & 80.02 \\
         Inc-gradual (bal.)      & 38.43 & 15.76 & 52.32 & 51.85 & 72.51 & 77.92 \\
         Inc-gradual (imbal.)    & 30.16 & 29.52 & 57.46 & 53.36 & 73.21 & 79.35 \\
         Inc-abrupt-reoc (bal.)   & 42.39 & 16.65 & 58.55 & 58.39 & 70.91 & 74.95 \\
         Inc-abrupt-reoc (imbal.) & 28.16 & 29.76 & 52.34 & 51.03 & 69.13 & 77.60 \\
         Inc-reoc (bal.)          & 40.46 & 16.66 & 48.77 & 47.83 & 72.30 & 77.13 \\
         Inc-reoc (imbal.)        & 28.21 & 29.76 & 52.58 & 55.22 & 69.56 & 77.62 \\
         Out-of-control           & 13.06 & 18.80 & 45.99 & 44.70 & 53.58 & 70.45 \\
         \hline
    \end{tabular}
\end{table}

For all datasets, we can note that the Adaptive Random Forest (ARF) presented the best overall results, followed by the Leveraging Bagging (Lev.Bag.). For these methods, the overall results are around 70-80\% for different patterns of drifts, which are slightly inferior when compared with our previous evaluations on static data with a similar feature set in the problem of insect species recognition~\citep{de2013classification, silva2015exploring, qi2015effective}. As expected, both incremental algorithms (VFDT and Naive Bayes), which do not consider a strategy to deal with concept drifts explicitly, were outperformed by more powerful data stream classifiers. The poor performance of baseline classifiers gives us empirical evidence that undesirable characteristics such as temporal dependence and the prevalence of majority classes are underrepresented in our data.

Table~\ref{tab:classification_results} results provide a general view of the classification performance of the algorithms. However, in data streams, we are frequently interested in seeing these performances over time. Besides, the performance over time of some approaches, such as the baseline classifiers can help to understand the changes in the data. In this direction, we present below the individual evaluation for each dataset from our benchmark.

\begin{figure}[htb]
\centering
   \subfigure[Balanced]{
     \includegraphics[scale=0.32]{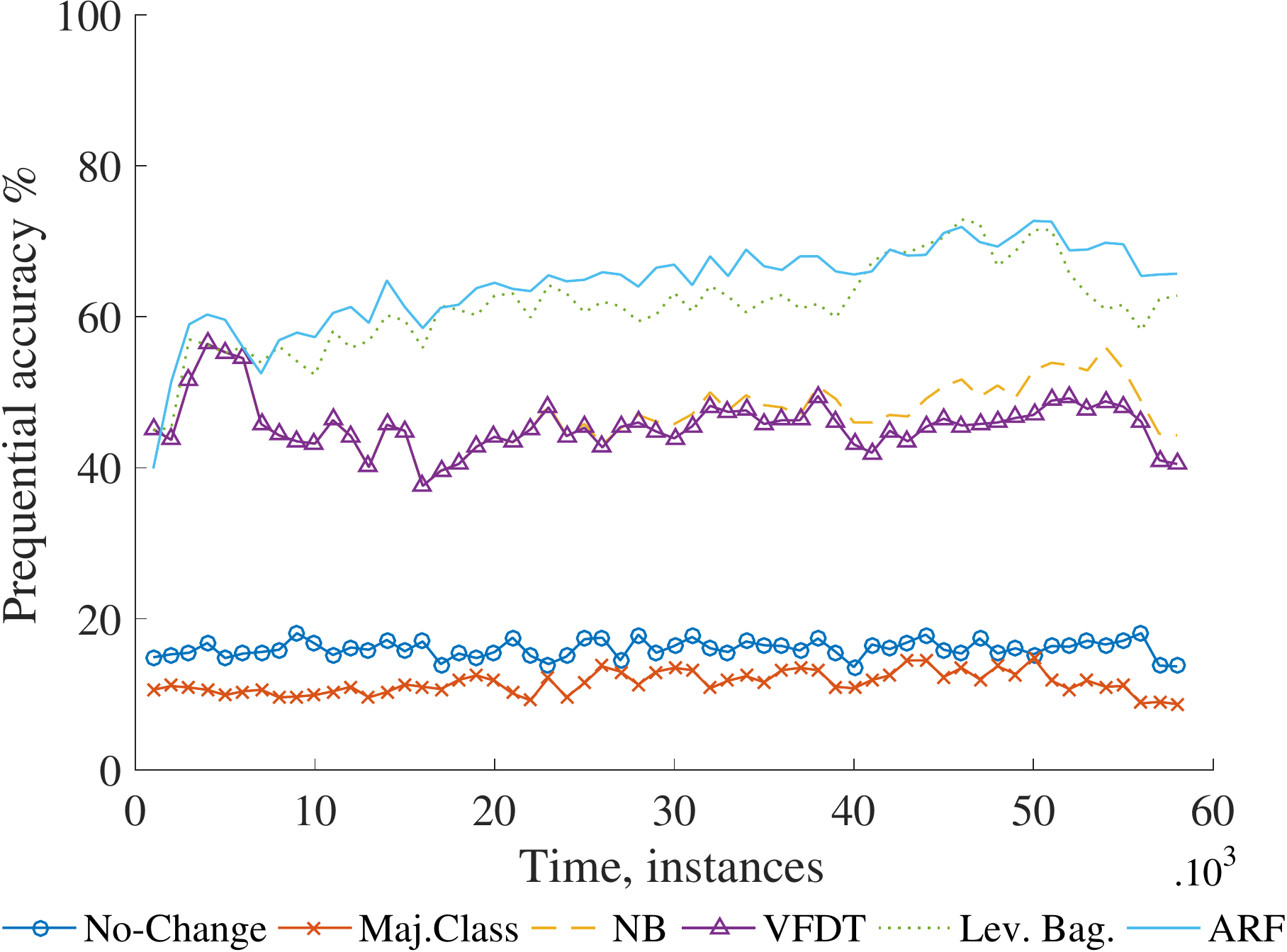}
   }
   \hspace{-0.29cm}
   \subfigure[Imbalanced]{
     \includegraphics[scale=0.32]{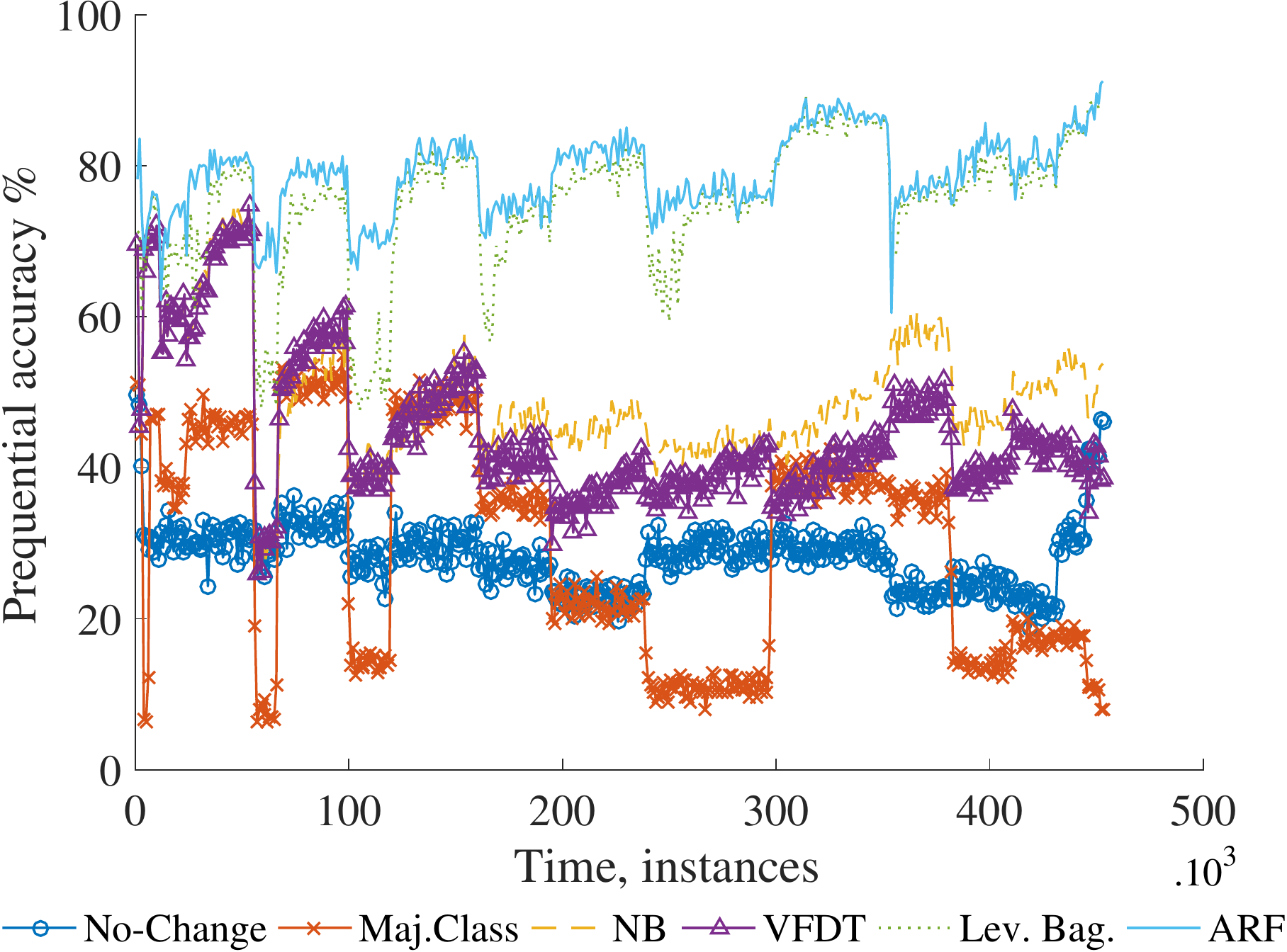}
   }
         \caption{Prequential accuracy on the  Incremental data.}
   \label{fig:incremental_results}
\end{figure}

Fig.~\ref{fig:incremental_results} shows the prequential accuracy results achieved over time by the compared methods for the balanced and imbalanced versions of Incremental data. Given the slow speed of the incremental changes, the algorithms tend to present more stable performances without significant accuracy increase or decrease. In the imbalanced version, the algorithms show instabilities in well-defined points, probably due to the $P(Y)$ changes.

\begin{figure}[htb]
\centering
   \subfigure[Balanced]{
     \includegraphics[scale=0.32]{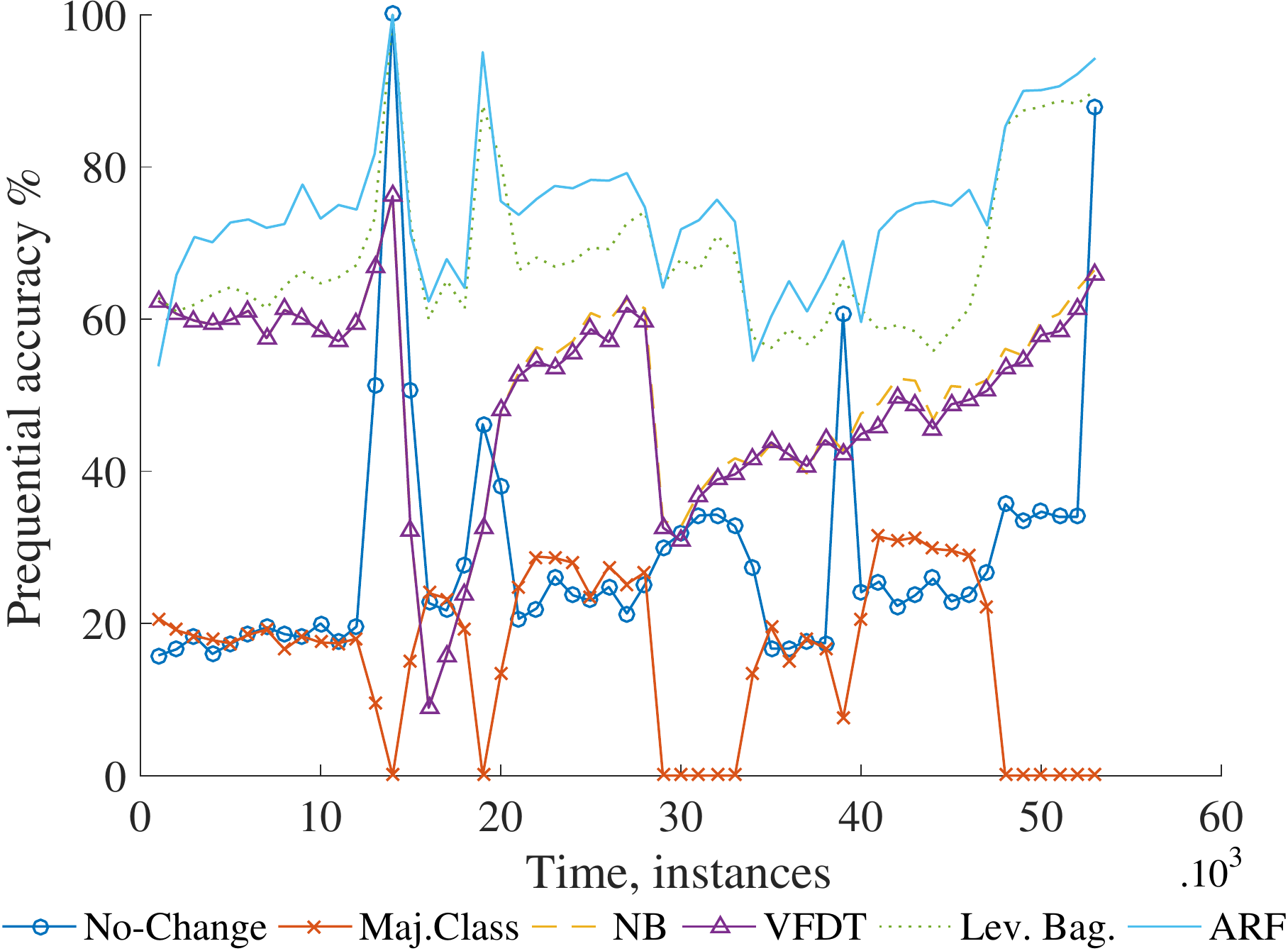}
   }
   \hspace{-0.29cm}
   \subfigure[Imbalanced]{
     \includegraphics[scale=0.32]{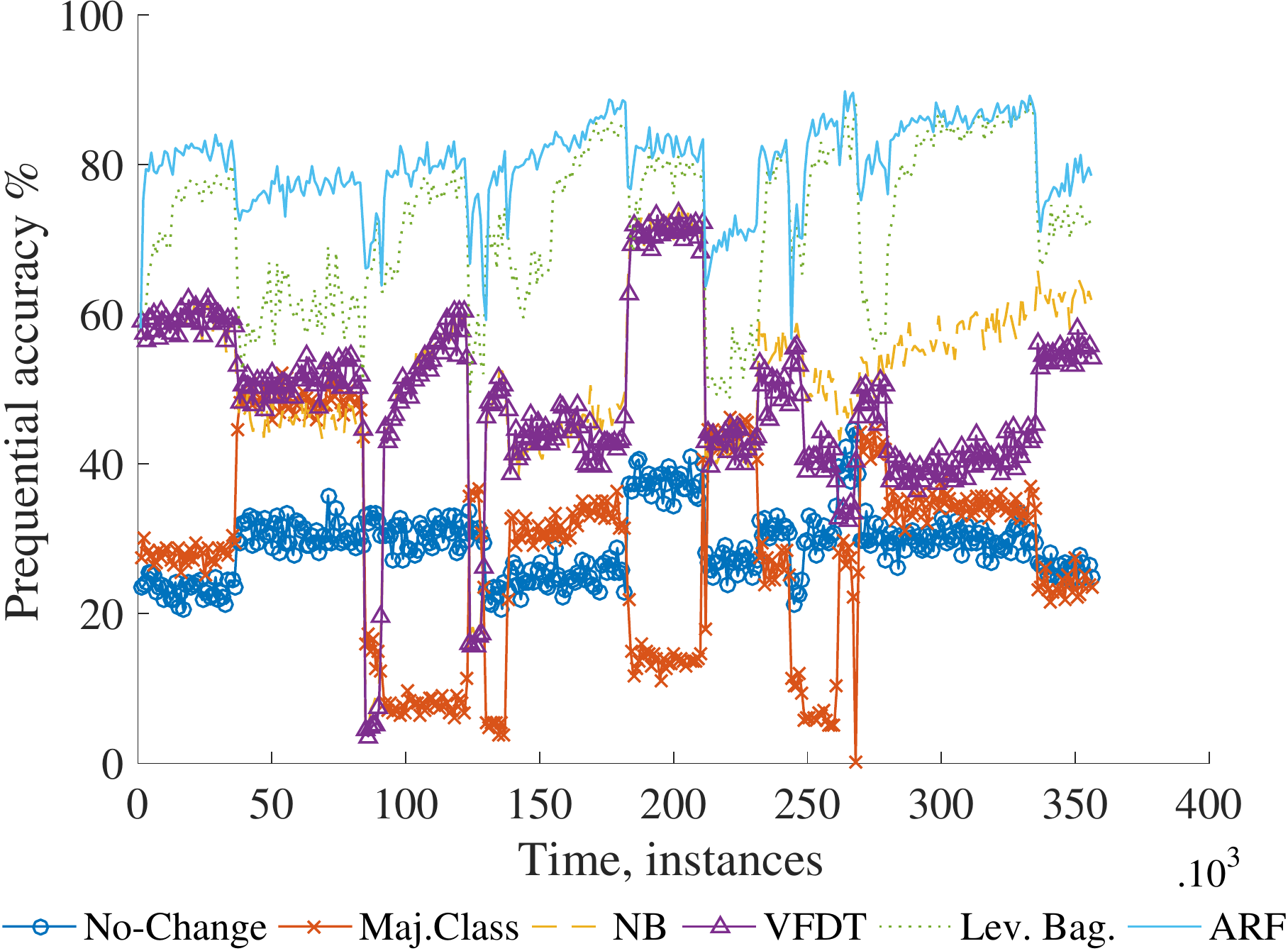}
   }
         \caption{Prequential accuracy on the Abrupt  data.}
   \label{fig:abrupt_results}
\end{figure}

Fig.~\ref{fig:abrupt_results} shows the results over time for the two versions of Abrupt data. In the balanced version, we can note the presence of temporal dependence in four different points of the stream (close to the times 14,000; 19,000; 40,000; and 52,000). However, in all cases, they are rapidly dissolved as we can note by the poor performance of No-Change classifier over time. 

\begin{figure}[htb]
\centering
   \subfigure[Balanced]{
     \includegraphics[scale=0.32]{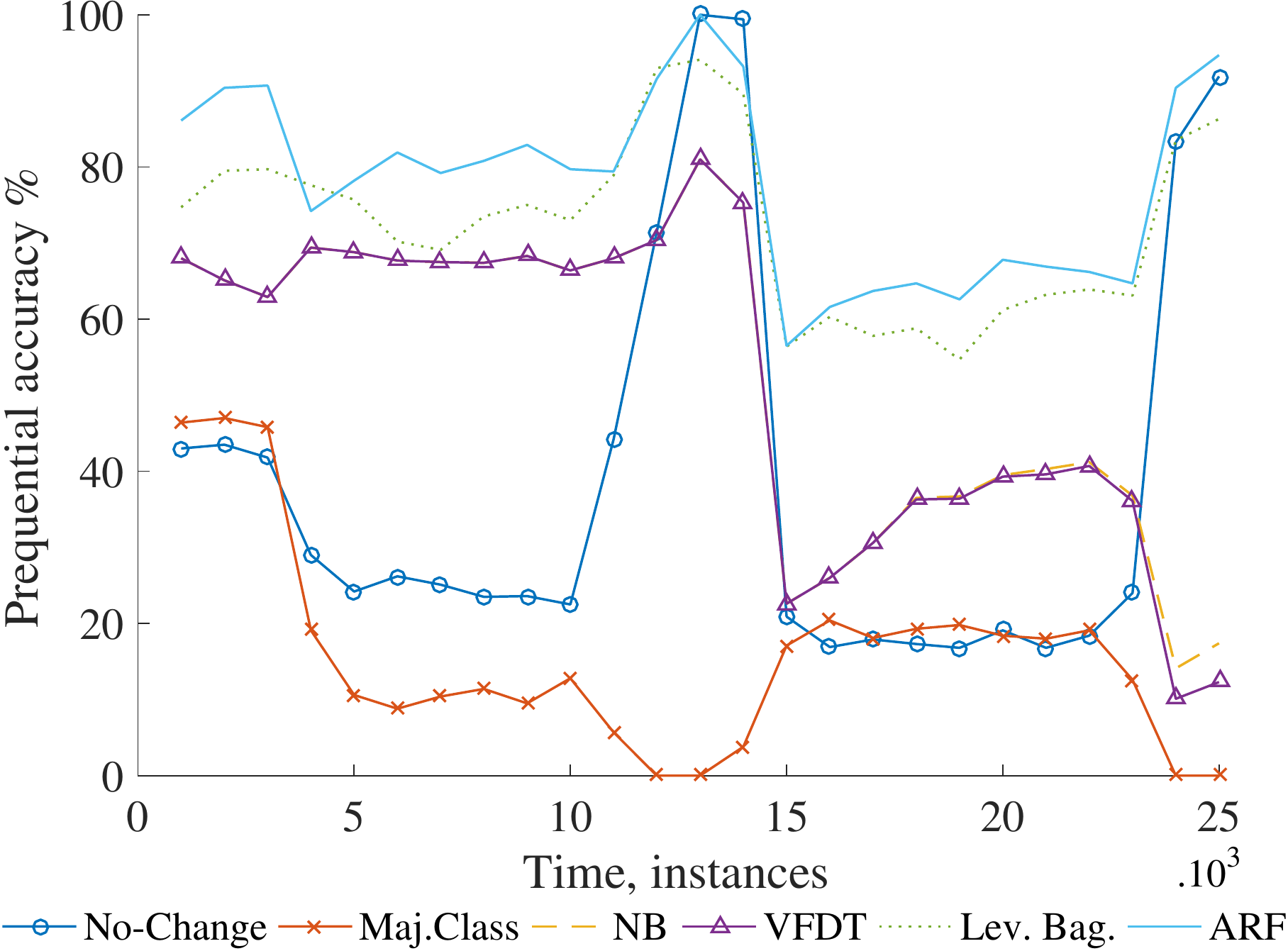}
   }
   \hspace{-0.29cm}
   \subfigure[Imbalanced]{
     \includegraphics[scale=0.32]{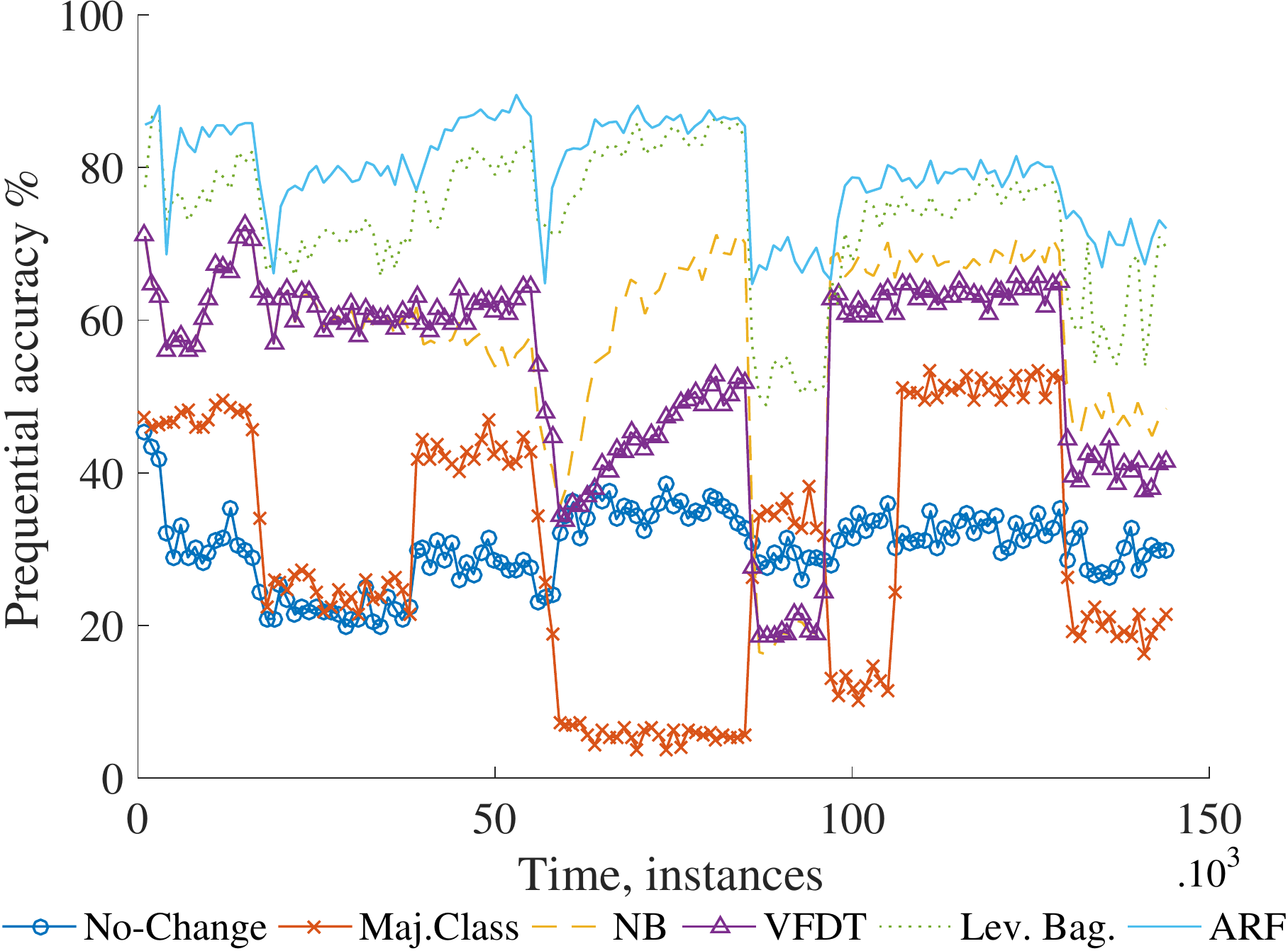}
   }
         \caption{Prequential accuracy on the  Incremental-gradual data.}
   \label{fig:incremental-gradual_results}
\end{figure}

Fig.~\ref{fig:incremental-gradual_results} shows the results over time for the Incremental-gradual data. For the balanced version, we see a drastic fall in the performances of the classifiers immediately before 15,000 instances from the stream, which is related to the occurrence of gradual drift. At this period, the stream presents instances from two different concepts at the same time until the change for a new concept is complete. 

\begin{figure}[htb]
\centering
   \subfigure[Balanced]{
     \includegraphics[scale=0.32]{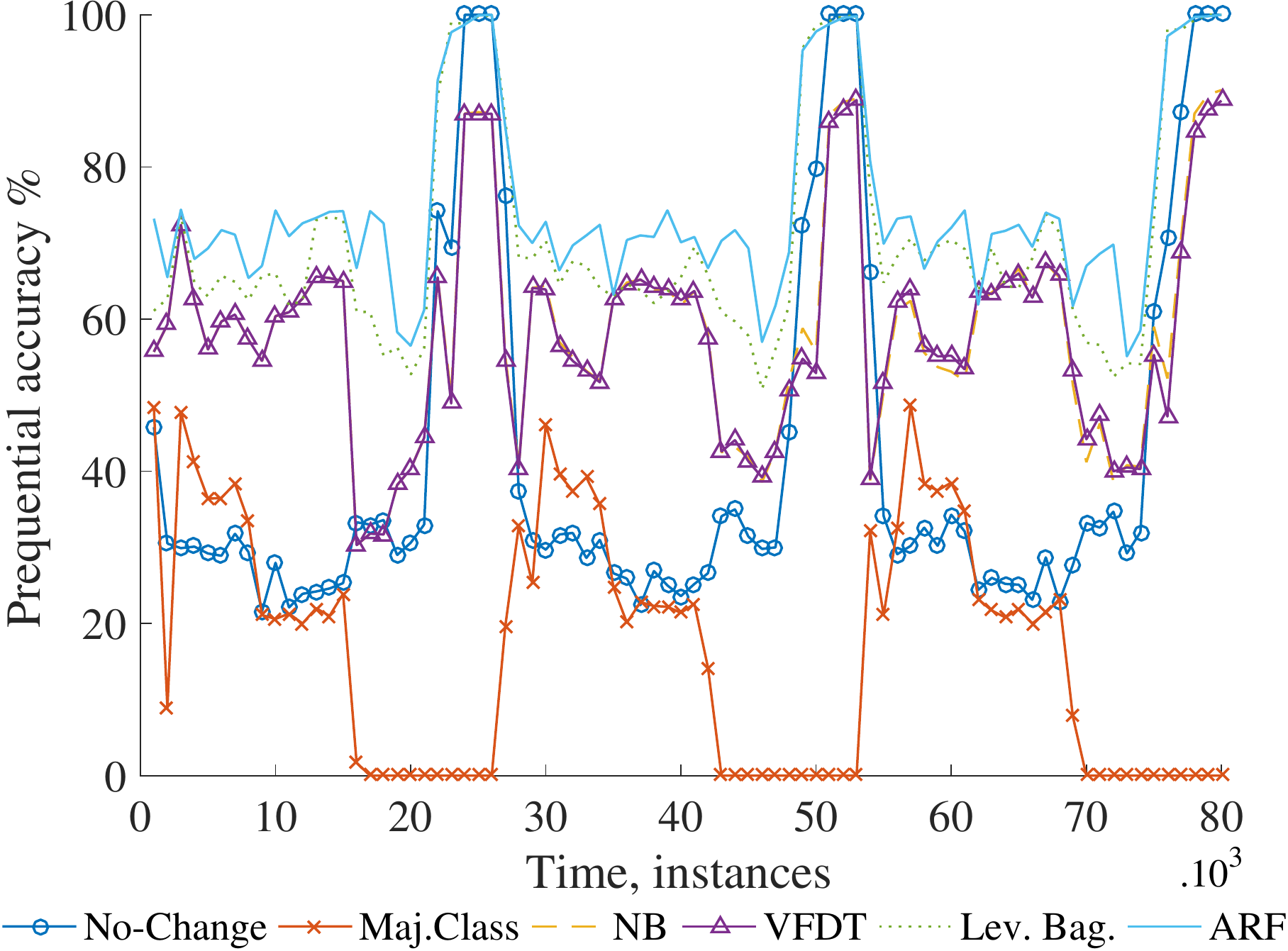}
   }
   \hspace{-0.29cm}
   \subfigure[Imbalanced]{
     \includegraphics[scale=0.32]{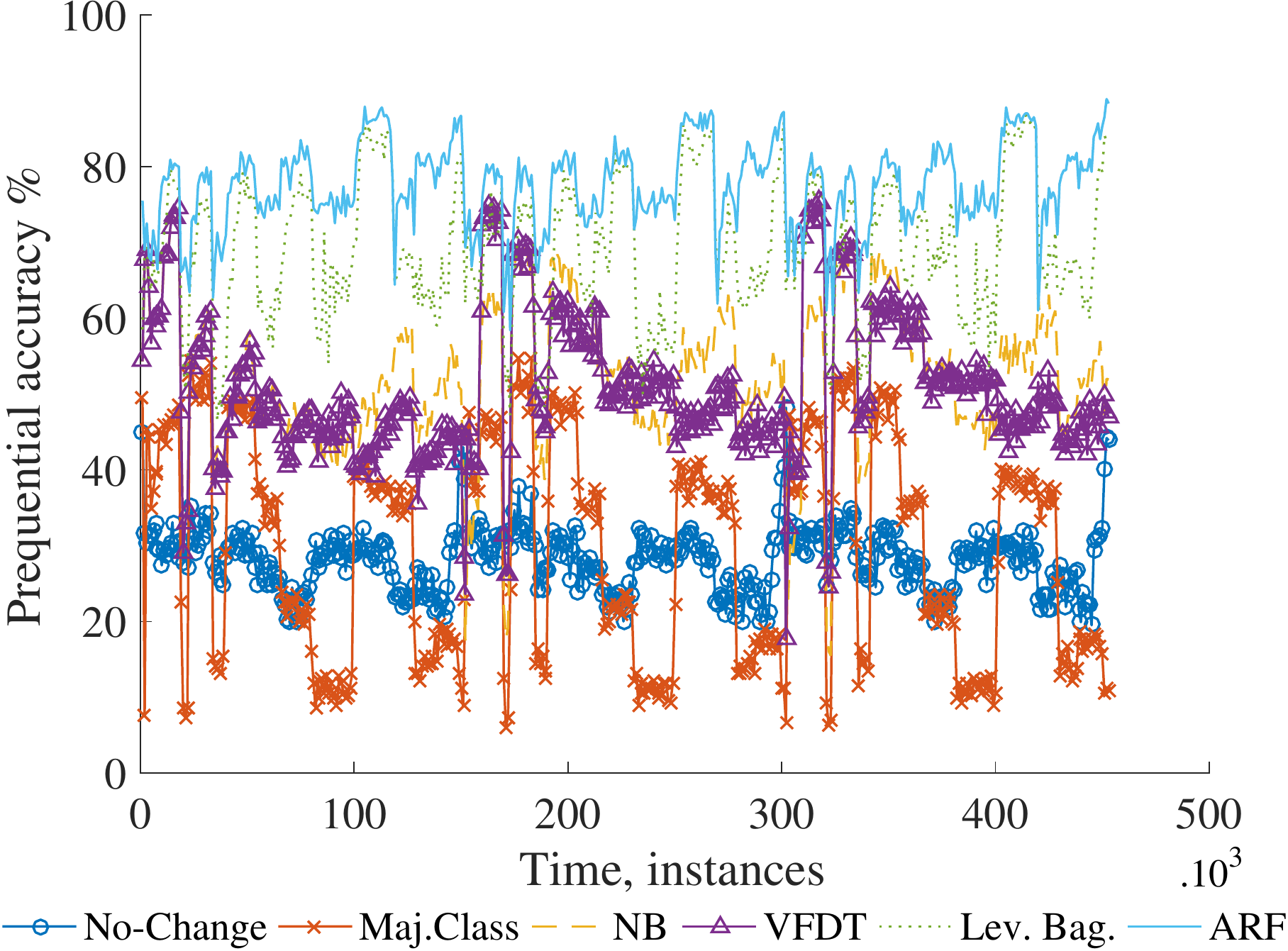}
   }
         \caption{Prequential accuracy on the  Incremental-abrupt-reoccurring data.}
   \label{fig:incremental-abrupt-reoccurring_results}
\end{figure}

Fig.~\ref{fig:incremental-abrupt-reoccurring_results} shows the results for the Incremental-abrupt-reoccurring data. In the balanced version, we can note three different periods where the classifiers achieve accuracy peaks in their performances, followed by a significant fall in the first two cases. These periods correspond to the end of a cycle of incremental changes and the start of an abrupt change. In the imbalanced data version, the analysis of VFDT results can help to understand the data better. In this case, we can note in three different periods, an incremental fall of the classifier performance with values between 80\% to 40\%.

\begin{figure}[htb]
\centering
   \subfigure[Balanced]{
     \includegraphics[scale=0.32]{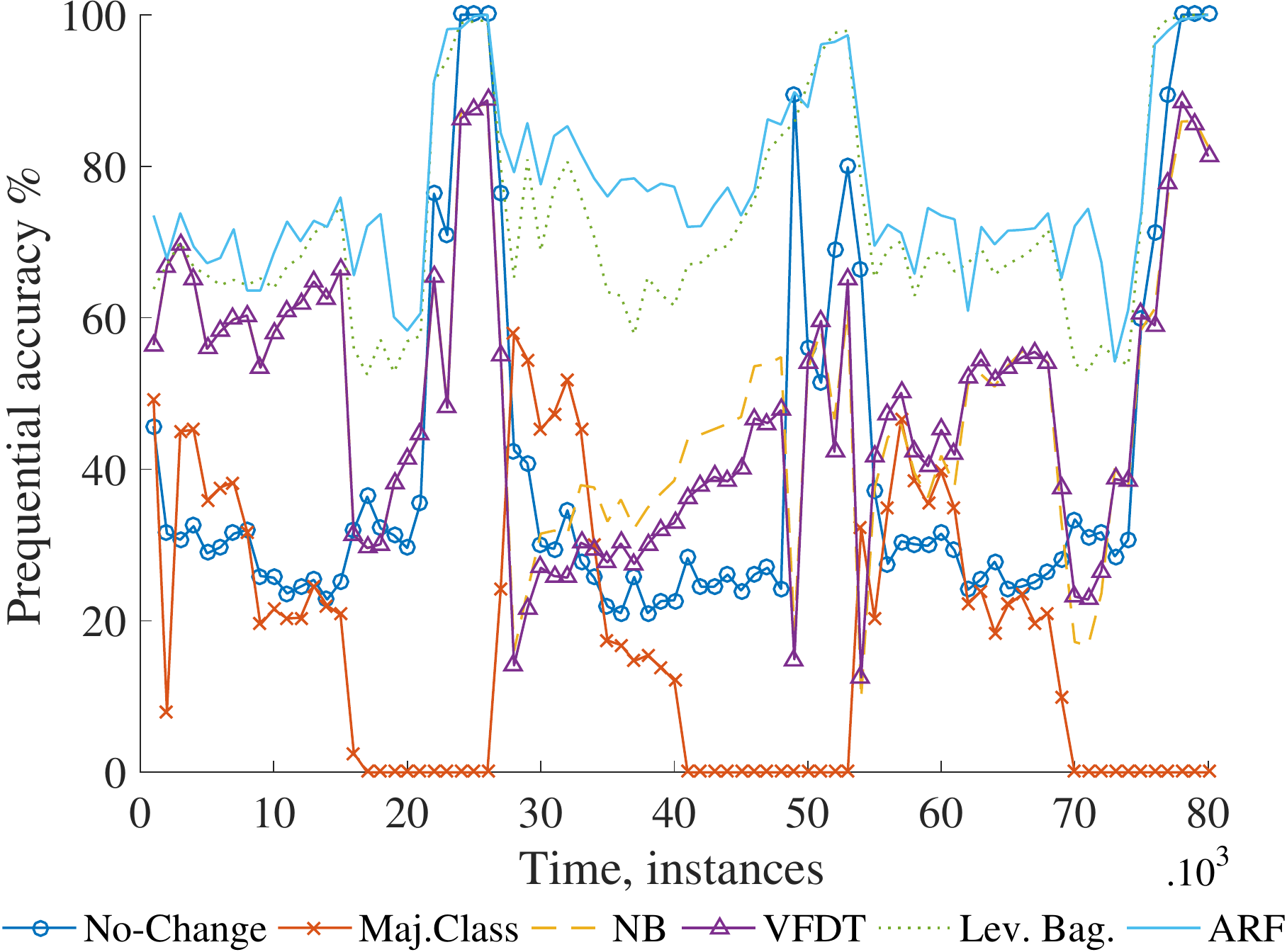}
   }
   \hspace{-0.29cm}
   \subfigure[Imbalanced]{
     \includegraphics[scale=0.32]{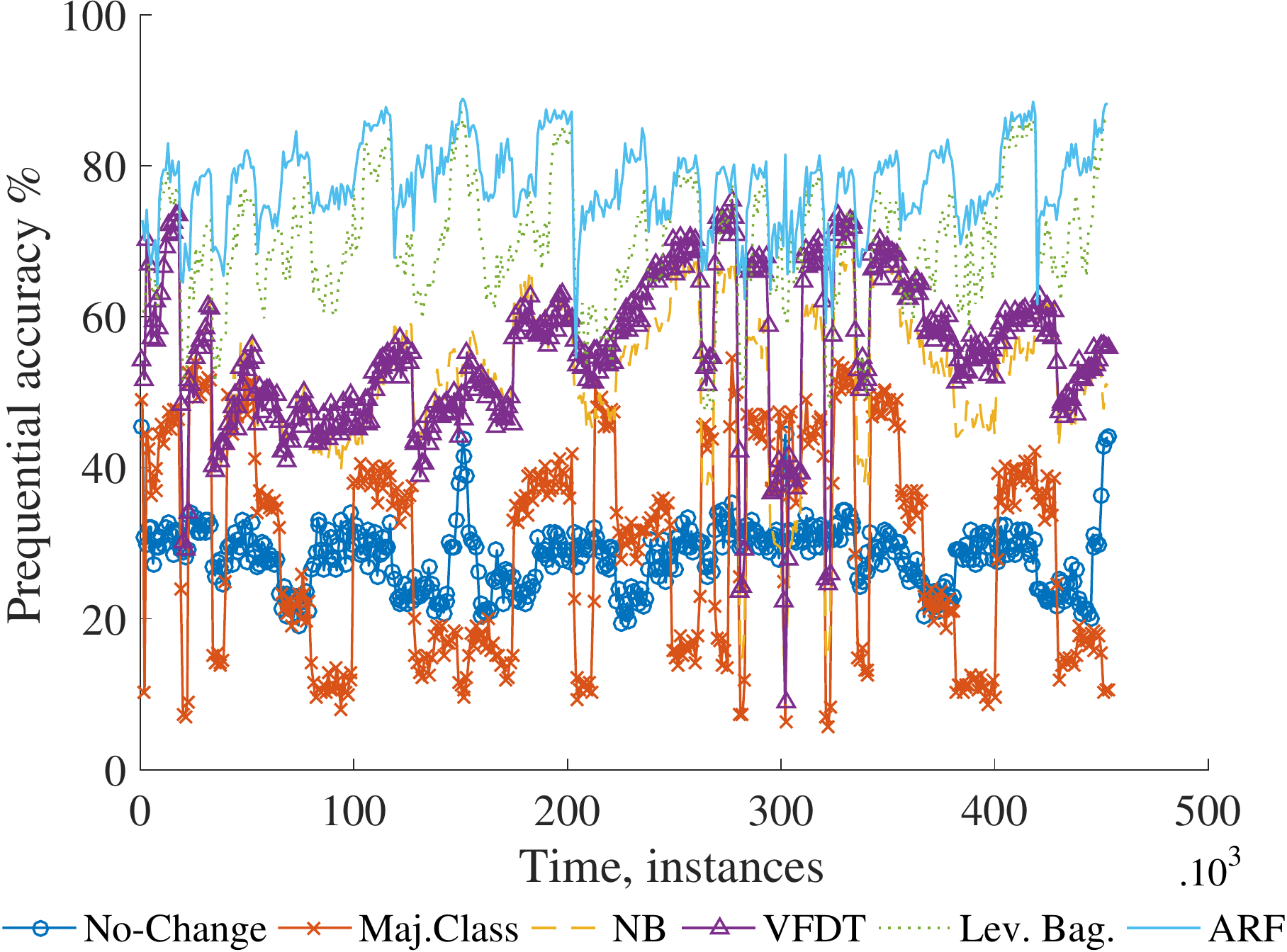}
   }
         \caption{Prequential accuracy on the  Incremental-reoccurring data.}
   \label{fig:incremental-reoccurring_results}
\end{figure}

In Fig.~\ref{fig:incremental-reoccurring_results}, we show the results for balanced and imbalanced versions of Incre\-men\-tal-reoccurring data. Although the main difference of this dataset with Incremental-abrupt-reoccurring data is the presence of abrupt changes at two different times, the results are very similar to those previously shown in Fig.~\ref{fig:incremental-abrupt-reoccurring_results}. It can mean that abrupt changes are not responsible for significant impacts in the performances of the algorithms, mainly when we observe recurring concepts in the stream. In general, the algorithms have more difficult to adapt to incremental changes.

\begin{figure}[htb]
\centering
     \includegraphics[scale=0.4]{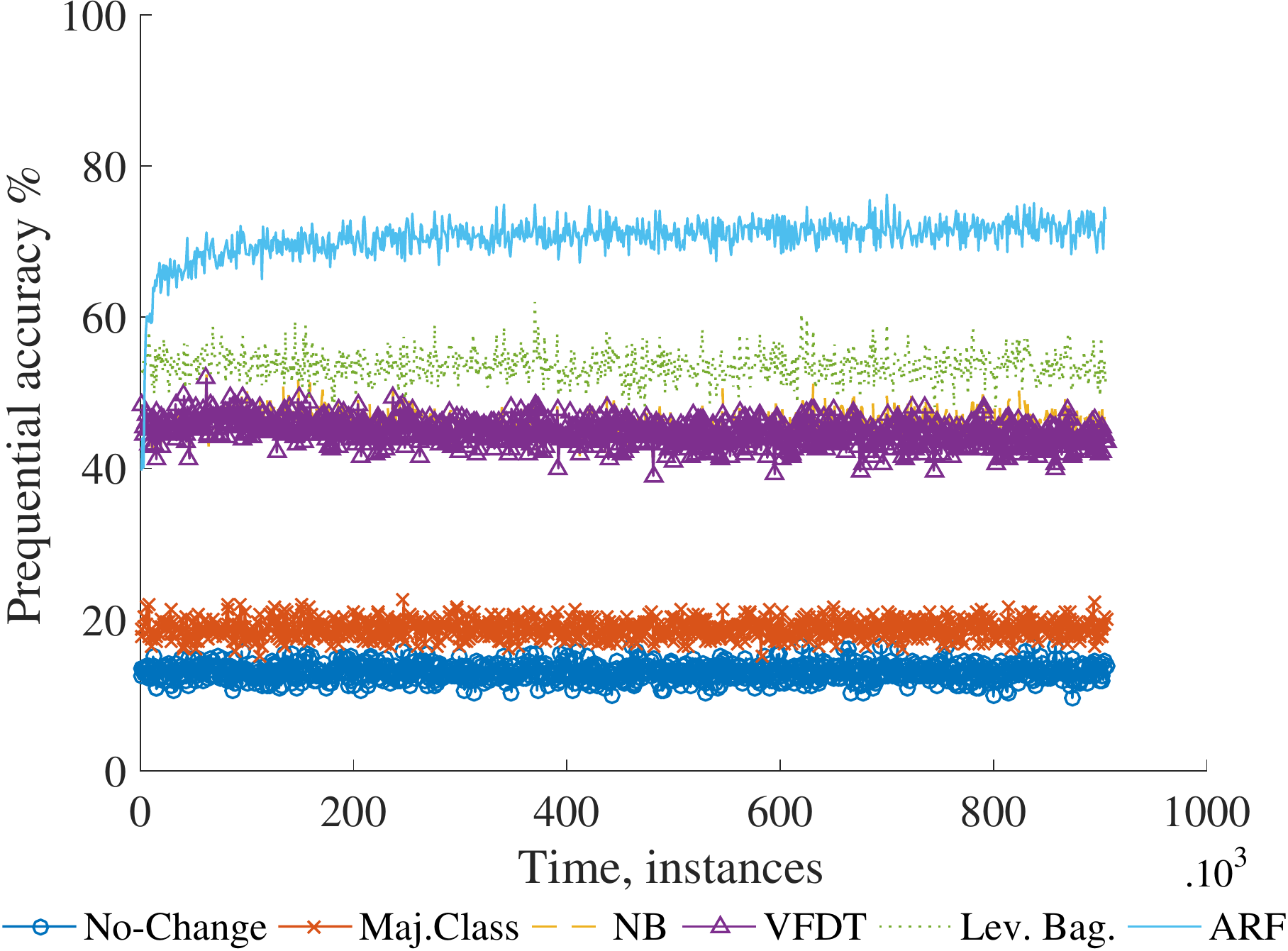}

         \caption{Prequential accuracy on the Out-of-control data.}
   \label{fig:out-of-control_results}
\end{figure}

Fig.~\ref{fig:out-of-control_results} shows the results for the Out-of-control data. It is interesting to note that although this dataset has a large number of class labels and undefined changes in type and number, the classifiers show more stable performances over time when compared with other datasets. However, the results are limited. For example, the best classifier (ARF), shows an overall prequential accuracy around 70\%.

\subsection{Drift Detection}

We choose representative methods for different drift detection approaches to evaluate the performance of detectors considering our benchmark data. Specifically, we consider the following methods:

\begin{itemize}
    \item Sequential analysis: Page-Hinkley Test (PHT) and CUSUM \citep{page1954continuous};
    \item Statistical process control: Drift Detection Method (DDM)~\citep{gama2004learning} and Exponentially Weighted Moving Average (EWMA) \citep{ross2012exponentially};
    \item Comparison of data distributions: Adaptive Windowing (ADWIN)~\citep{bifet2007learning}, SEED~\citep{huang2014detecting}, and Statistical Test of Equal Proportions (STEPD)~\citep{nishida2007detecting}.
\end{itemize}

Regarding the parameters of the detectors, we consider a window size with 1,000 examples and a minimum of 100 examples to detect a drift. For a fair comparison, the remaining parameters follow the default values suggested by MOA. As all evaluated methods require a base classifier, we consider the Naive Bayes for all approaches to standardize the experimental evaluation. Thus, we can evaluate the performance of drift detection methods based on the prequential accuracy without the influence of the classification algorithm. 

In Table~\ref{tab:cd_results}, we show the overall prequential accuracy (Acc.) and the total of changes detected (C.D.) by the different drift detectors evaluated considering the Insect Stream Data. For each dataset, we highlighted the best accuracy in \textbf{bold}. In general, the best results are achieved by the methods ADWIN and STEPD.

\begin{table}[htb]
    \centering
    \scriptsize
    \renewcommand\tabcolsep{1.2pt}   
    \caption{Overall prequential accuracy (Acc.) and total of changes detected (CD) by different drift detectors.}
    \label{tab:cd_results}
    \begin{tabular}{lcc|cc|cc|cc|cc|cc|cc}
        \multirow{2}{*}{\textbf{Dataset}} & \multicolumn{2}{c|}{\textbf{ADWIN}} & \multicolumn{2}{c|}{\textbf{PHT}} & \multicolumn{2}{c|}{\textbf{CUSUM}} & \multicolumn{2}{c|}{\textbf{DDM}} & \multicolumn{2}{c|}{\textbf{EWMA}} & \multicolumn{2}{c|}{\textbf{SEED}} & \multicolumn{2}{c}{\textbf{STEPD}} \\ 
        & \textbf{Acc.} & \textbf{CD} & \textbf{Acc.} & \textbf{CD} & \textbf{Acc.} & \textbf{CD} & \textbf{Acc.} & \textbf{CD} & \textbf{Acc.} & \textbf{CD} & \textbf{Acc.} & \textbf{CD} & \textbf{Acc.} & \textbf{CD} \\
        \hline
Inc (bal.) & 52.68 & 3 & 54.17 & 1 & \textbf{56.63} & 5 & 52.72 & 1 & 47.37 & 0 & 54.96 & 9 & 56.55 & 30 \\
Inc (imbal.) & \textbf{61.02} & 136 & 58.79 & 42 & 59.97 & 99 & 49.32 & 9 & 50.37 & 104 & 58.50 & 86 & 59.96 & 225 \\
Abrupt (bal.) & 62.48 & 7 & 62.14 & 6 & 64.63 & 8 & 60.36 & 5 & 65.40 & 90 & 65.73 & 21 & \textbf{66.02} & 28 \\
Abrupt (imbal.) & 58.72 & 94 & 59.89 & 37 & 60.70 & 71 & 56.78 & 9 & 52.23 & 85 & 60.31 & 68 & \textbf{61.52} & 185 \\
Inc-gradual (bal.) & \textbf{72.26} & 6 & 68.30 & 7 & 69.20 & 9 & 65.40 & 6 & 71.39 & 39 & 70.38 & 14 & 71.51 & 25 \\
Inc-gradual (imbal.) & \textbf{67.70} & 36 & 62.57 & 20 & 63.53 & 41 & 55.25 & 15 & 58.90 & 182 & 62.30 & 64 & 62.32 & 64 \\
Inc-abrt-reoc (bal.) & 63.80 & 22 & 63.25 & 17 & 65.12 & 25 & 61.35 & 16 & 66.23 & 114 & 67.90 & 60 & \textbf{68.77} & 61 \\
Inc-abrt-reoc (imbal.) & 59.98 & 157 & 58.51 & 90 & 59.15 & 120 & 53.13 & 31 & 51.41 & 76 & 58.79 & 199 & \textbf{60.22} & 297 \\
Inc-reoc (bal.) & 65.93 & 26 & 64.59 & 16 & 65.87 & 21 & 63.96 & 21 & 66.45 & 108 & \textbf{69.82} & 47 & 69.45 & 59 \\
Inc-reoc (imbal.) & \textbf{60.39} & 152 & 58.16 & 67 & 59.65 & 122 & 55.13 & 34 & 51.91 & 96 & 59.00 & 163 & 59.68 & 242 \\
Out-of-control & \textbf{49.92} & 237 & 47.22 & 52 & 48.40 & 155 & 45.75 & 3 & 45.99 & 0 & 46.86 & 98 & 48.83 & 444 \\
\hline
    \end{tabular}
\end{table}

We chose three different cases to better analyze the drift detection task over time. Specifically, we present the results of STEPD, a method based on the comparison of distributions composed by data of two accuracies achieved by the classifier in two times: the recent one and the overall one. The balanced data versions of Abrupt, Incremental-abrupt-reoccurring, and Incremental-reoccurring datasets were analyzed. 

\begin{figure}[htb]
    \centering
    \includegraphics[scale=0.45]{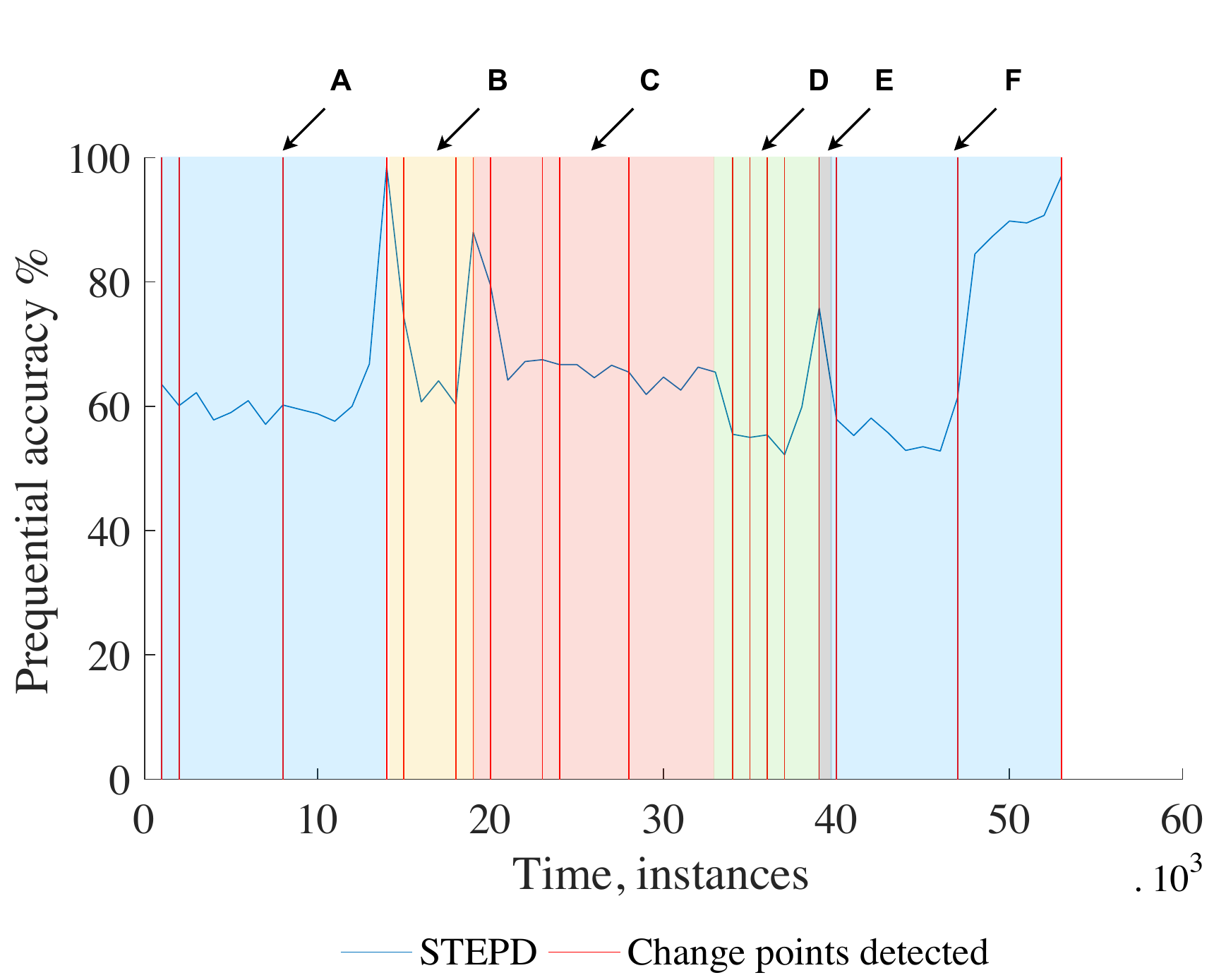}
    \caption{Prequential accuracy of Naive Bayes classifier with STEPD drift detection method and the change points detected considering the Abrupt data (balanced). Different background colors represent the six concepts (A-F) of this data.}
    \label{fig:cd_abrupt}
\end{figure}

In Fig.~\ref{fig:cd_abrupt}, we show the prequential accuracy achieved by the Naive Bayes classifier using the STEPD drift detector method on Abrupt data. In this figure, we can see the 28 change points detected in the vertical red lines. In this data, we have six different concepts that occur after five abrupt changes. Different background colors in the figure represent the six concepts (A-F) of this data. We can note in Fig.~\ref{fig:cd_abrupt} that even during the arrival of instances from a stable concept, the method incorrectly detects different change points. In most cases, the model adaptations in these points do not lead to better accuracy, except in the last changes identified into the concepts B, D, and F. We also can note that all abrupt changes were correctly identified.

\begin{figure}[htb]
    \centering
    \includegraphics[scale=0.45]{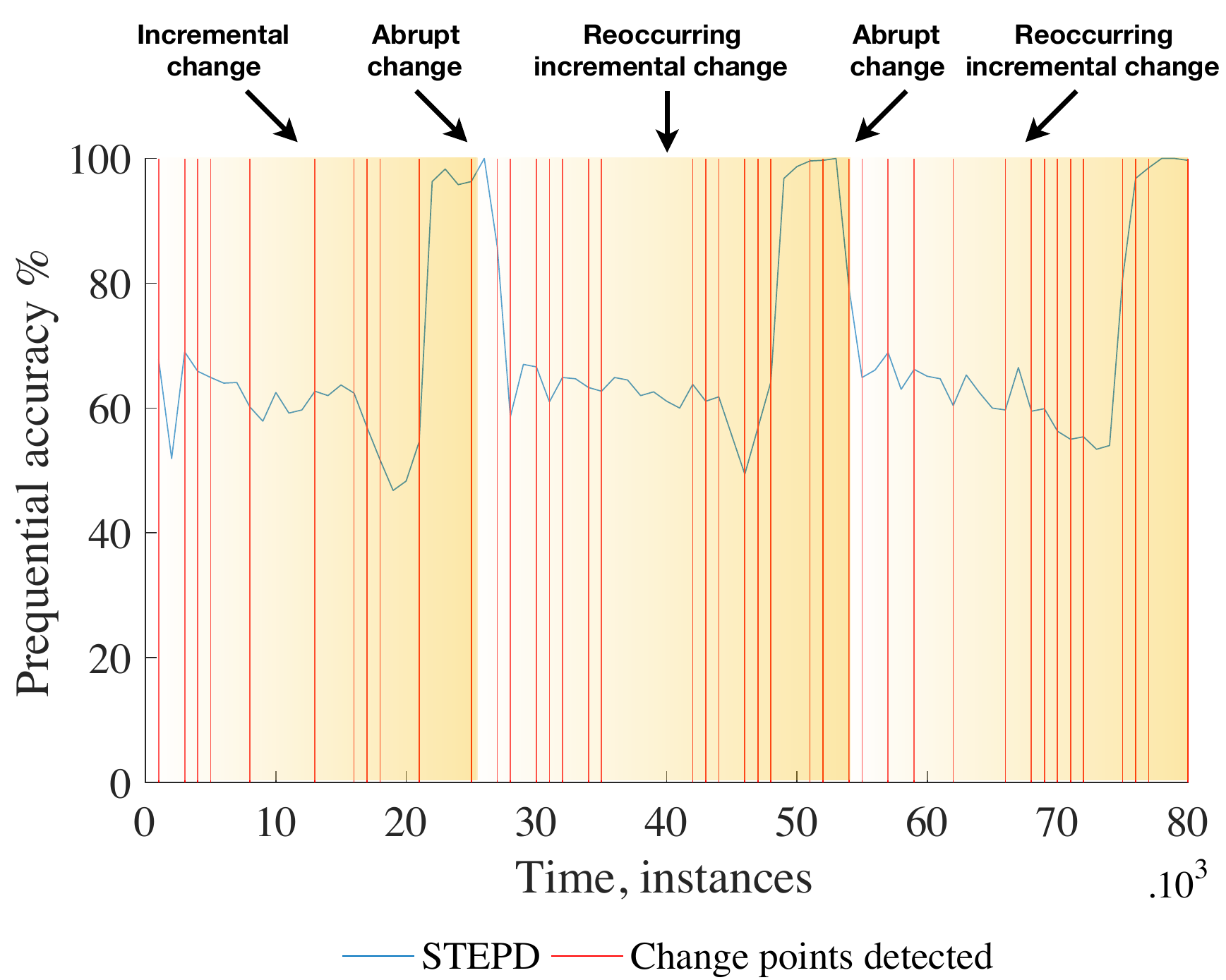}
    \caption{Prequential accuracy of Naive Bayes classifier with STEPD drift detection method and the change points detected considering the Incremental-abrupt-reoccurring data (balanced). The gradient in the background color represents the incremental changes. The abrupt changes occur between two consecutive incremental changes. All the changes are indicated in the top view of the figure.}
    \label{fig:cd_abrupt_reoc}
\end{figure}

Similarly, in Fig.~\ref{fig:cd_abrupt_reoc} we present the results on Incremental-abrupt-reoccurring data. In this data, we have two different points with well defined abrupt changes. However, it also occurs minor incremental changes between the abrupt changes. The gradient in the background color of the figure represents the incremental changes. All the changes are indicated in the top view of the figure. Given the constant occurrence of incremental changes in these data over all the stream, we can note a high number of change points identified by the method in Fig.~\ref{fig:cd_abrupt_reoc}. Specifically, STEPD identified 61 change points.

\begin{figure}[htb]
    \centering
    \includegraphics[scale=0.45]{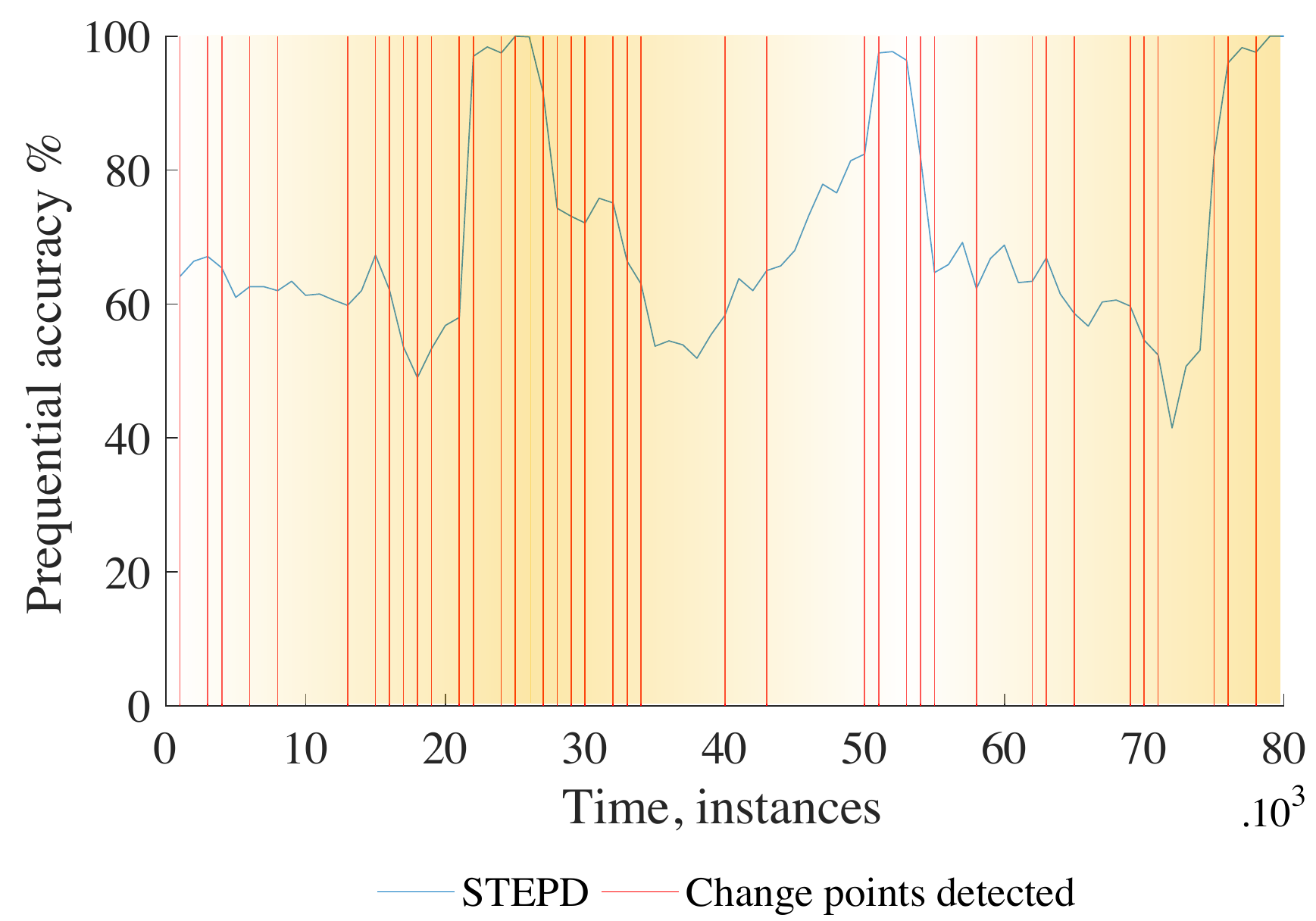}
    \caption{Prequential accuracy of Naive Bayes classifier with STEPD drift detection method and the change points detected considering the Incremental-reoccurring data (balanced). The gradient in the background color represents the incremental changes.}
    \label{fig:cd_inc_reoc}
\end{figure}

In Fig.~\ref{fig:cd_inc_reoc}, we show the results on Incremental-reoccurring data. As this data only present incremental changes over time, it is more difficult to precisely indicate the change points in the stream. However, we show a general view of these changes by the gradient in the background color of the figure. In this dataset, STEPD identified 59 change points. 

\section{Conclusions}\label{sec:conclusions}
In this paper, we discuss the challenges faced by the stream learning community concerning the reduced number of real-world data and the lack of a benchmark to evaluate adaptive classifiers and drift detectors. This gap leads to the use of synthetic data accompanied by a small number of real data in the evaluation of new proposals. The main problem of this practice is the possibility of data bias, which can lead to incorrect conclusions about stream algorithms behavior. We also present a review regarding the main real datasets evaluated in the literature and discuss some weaknesses in such data as the lack of knowledge about the type/pattern of change and when it occurs in the stream.

To mitigate some of the problems identified in the evaluation of stream methods concerning the lack of real data, we propose the use of 11 new datasets collected by an optical sensor that measures the flying behavior of insects. This data is used in a relevant application of public health related to the use of a Smart Trap to attract and capture target species such as the vector of diseases. In this application, non-stationary data are generated over time in streaming fashion due to changes in the environment, which impacts the insects' behavior. Our proposed data has interesting characteristics to be explored by researchers, such as different patterns of changes (incremental, abrupt, gradual, and reoccurring), indicators of the presence of each change and when they occur, the presence of complex changes in the class distribution, a significant number of instances, among others.

Although the proposed benchmark constitutes an essential contribution to the stream mining community, it is also important to note that such data also have some limitations. We highlighted two of them. First, to precisely indicate the drift points and the types of drift, we performed a manual manipulation in the original arrival order of the examples. Also, to avoid problems such as temporal dependence, we performed a shuffling procedure into a window with similar examples. In practice, such procedures do not affect the meaning of the application, which can experience the simulated changes in real environments. However, such manipulation could be interpreted as responsible for generating data \textit{semi-real} or not entirely real. The second limitation, which most of the datasets from literature also presents, is the lack of time-stamps. Such limitation poses some restrictions to evaluate issues where the time is an additional constraint factor in the learning task. For example, with the time-stamps, it is possible to verify if the classification model is updated at the available time between the examples' arrival. Also, the algorithms can take this time into consideration to perform other updates in idle periods of the classifier.

We also provide to the machine learning community a new public repository called USP Data Stream Repository, where we make available 27 datasets from different real problems composed by 16 data previously evaluated by other works from literature and 11 new datasets obtained by the optical sensor for automatic insect recognition. In this repository, we also present the results achieved by two baseline methods for all datasets. This repository will be regularly fed with new data from our future works and donation.

\begin{acknowledgements}
The authors would like to thank Prof. Juliano J. Corbi and their laboratory staff, as well as Edi Samuel B. Mendon\c{c}a and PETE Company by the support in the data collection. This study was financed in part by S\~ao Paulo Research Foundation (FAPESP) in the grant numbers \#16/04986-6, \#17/22896-7, and \#18/05859-3, the Brazilian National Council for Scientific and Technological Development (CNPq) in the grant number 306631/2016-4, Coordena\c{c}\~ao de Aperfei\c{c}oamento de Pessoal de N\'ivel Superior - Brasil (CAPES) - Finance Code PROEX-6909543/D, and the United States Agency for International Development (USAID, grant AID-OAA-F-16-00072).
\end{acknowledgements}

\bibliographystyle{spbasic}      
\bibliography{refs}   

\end{document}